\documentclass[10pt]{article}
\input{macros}
\usepackage{mathrsfs, soul}
\usepackage{todonotes}
\usepackage[dvipsnames]{xcolor}
\setstcolor{red}
\allowdisplaybreaks
\newcommand{\TV}{\mathrm{TV}}
\newcommand{\Ham}{\mathrm{Ham}}

\newcommand{\Yi}[1]{\textcolor{Bittersweet}{(Yi: #1)}}
\newcommand{\Elly}[1]{\textcolor{Green}{(Elly: #1)}}

\begin{document}

\title{Differentially Private Score Estimation with Applications to Score-Based Generative Models
}

\author{
    Mengxiao Gao\thanks{School of Operations Research and Information Engineering,  Cornell University; } \and
    Elly K. H. Hung\thanks{Department of Statistics, University of Warwick. } \and
	Yuchen Wu${}^{\ast}$ \and 
    Yi Yu${}^{\dagger}$
} 
 
\date{\today}

\maketitle

\begin{abstract}

Estimating the score function of a distribution plays a central role in statistics and machine learning, with applications in distribution estimation, sampling, and score-based generative modeling. 
However, learning scores from sensitive data raises privacy concerns. 
These concerns are particularly pronounced in score-based generative modeling, where trained models may memorize individual training examples and expose sensitive information through membership inference and data extraction attacks.

Although non-private score estimation has been extensively studied, the statistical limits of its private counterpart remain less understood. 
In this work, we study score estimation under $(\alpha,\beta)$-differential privacy for $d$-dimensional sub-Gaussian distributions with locally $\gamma$-H\"older-continuous scores.
We develop a sampling-based private score estimator using quadratic B-splines and the exponential mechanism, and we implement our proposal through projected Langevin Monte Carlo with runtime polynomial in the sample size $n$.
The proposed estimator satisfies $(\alpha,\beta)$-differential privacy, and achieves expected density-weighted squared error $\mbox{polylog}(n) \max(n^{-2\gamma/(d+2\gamma+2)},\,(n\alpha)^{-2\gamma/(d+\gamma+1)})$.
We also establish a nearly matching private minimax lower bound, thereby characterizing the privacy-accuracy trade-off for score estimation.
As an application, we construct a differentially private diffusion model whose privacy guarantee is independent of the number of denoising steps.
We show that, with an appropriate discretization scheme, the resulting diffusion model attains nearly minimax-optimal expected total variation error among all private distribution estimators. 
We validate our theoretical findings through experiments on both synthetic and real-world datasets. 

\end{abstract}

\tableofcontents

\section{Introduction}

\subsection{Score estimation and its applications}

The \emph{score function} of a distribution with positive, differentiable density $\rho_0$ is defined as $\nabla \log\rho_0(x)$, where the gradient is taken with respect to the input $x$. 
This function captures the local variation of the log-density and uniquely determines the underlying distribution. 
In practice, the score function is typically unknown and must be estimated from data.
The resulting estimates play an important role in a range of statistical and computational tasks, including those discussed below.

\paragraph{Score-based generative models. }

Score-based generative models (SGMs) are a powerful class of machine learning models that learn to sample from an unknown distribution by estimating the score functions of its noise-perturbed versions \citep{sohldickstein2015deep,song2019generative,ho2020denoising,song2021denoising,song2021scorebased}. 
\emph{Diffusion models} represent one widely used class of SGMs.   
Specifically, a diffusion model is built on a sequence of intermediate distributions $\{\rho_k\}_{k = 0}^{N + 1}$ obtained from a forward noising process. 
Starting from the target distribution $\rho_0$, each step of the forward process introduces a small amount of noise to transform $\rho_k$ into $\rho_{k+1}$, until $\rho_{N + 1}$ approximates an easy-to-sample reference distribution (typically standard Gaussian).
To generate samples from $\rho_0$, a diffusion model initializes samples from the reference distribution and applies a sequence of denoising steps that approximately reverses the forward noising process. 
In particular, for $k \in \{1, \ldots, N\}$, the $k$-th denoising step approximately transforms $\rho_{N - k + 2}$ into $\rho_{N - k + 1}$, and is guided by an estimate of the intermediate score $\nabla \log \rho_{N - k + 2}$.
Note that for numerical stability, the reverse process in diffusion models typically stops at $\rho_1$ rather than continuing to $\rho_0$. 
We illustrate sample generation with a diffusion model using the following schema: 
\begin{align*}
    \rho_{N + 1} \xrightarrow{\mathrm{sample}} Y_0 \xrightarrow{{\hat s_0\, \approx \,\nabla \log \rho_{N + 1} }} Y_{1} \xrightarrow{\hat s_1\, \approx \, \nabla \log \rho_{N} } Y_2 \xrightarrow{\hat s_2\, \approx \, \nabla \log \rho_{N - 1} } \ldots  \xrightarrow{\hat s_{N - 1} \, \approx \, \nabla \log \rho_{2} } Y_{N} \,\overset{d}{\approx}\, \rho_1. 
\end{align*}

\paragraph{Distribution estimation via score matching. }

Score matching is a machine learning technique that estimates an unknown distribution $\rho_0$ by fitting a candidate density whose score approximates that of the target distribution \citep{hyvarinen2005estimation}. 
Specifically, given a family of densities $\cF$ and samples from the target distribution $\rho_0$, score matching selects a candidate density by minimizing an empirical counterpart of the following population discrepancy between the model and target scores: 
\begin{align*}
	\hat\rho \in \argmin_{\rho \in \cF} \, \E_{\rho_0}\left[ \|\nabla \log \rho(X) - \nabla \log \rho_0(X) \|_2^2 \right]. 
\end{align*} 
An advantage of score matching over likelihood-based estimation is that it avoids evaluating potentially intractable normalizing constants for densities in $\cF$. 
The above objective involves the unknown target score $\nabla \log \rho_0$. Under suitable regularity conditions, integration by parts yields an objective that can be estimated directly from observations and differs from the original objective only by an additive constant.
The fitted model then defines an estimate of the target distribution. 
Score matching was first developed for parametric models, and was later extended to non-parametric density estimation through kernel exponential families, for which consistency and convergence guarantees have been established \citep{sriperumbudur2017density,sutherland2018efficient,zhou2020nonparametric}. 

\paragraph{Sampling with Langevin dynamics. }
Many Markov chain Monte Carlo methods rely on Langevin dynamics to sample from a target distribution using its score function \citep{0b5e028f-1a33-3fed-ae08-229fd5443c3f,dalalyan2017theoretical}. 
Specifically, for a target density $\rho_0$ on $\mathbb{R}^d$, the corresponding overdamped Langevin diffusion is defined by the following stochastic differential equation: 
\begin{align*}
	\mathrm{d}X_t=\nabla\log\rho_0(X_t)\,\mathrm{d}t+\sqrt{2}\dd B_t, 
\end{align*}
where $(B_t)_{t\geq0}$ is a standard Brownian motion in $\mathbb{R}^d$.
Under suitable regularity and ergodicity conditions, this diffusion has invariant density $\rho_0$, and the law of $X_t$ converges to the target distribution as $t\to\infty$.
Discretizing this diffusion yields practical sampling algorithms that require only score evaluations, avoiding the need to evaluate the potentially intractable normalizing constant of $\rho_0$.
When the score of $\rho_0$ is unknown, substituting an estimate learned from data provides a data-driven approach to approximate sampling \citep{song2019generative,block2020generative,cao2026on}.

\subsection{Privacy considerations}

\begin{figure}[hbt!]
    \centering
    \includegraphics[width=0.3\linewidth]{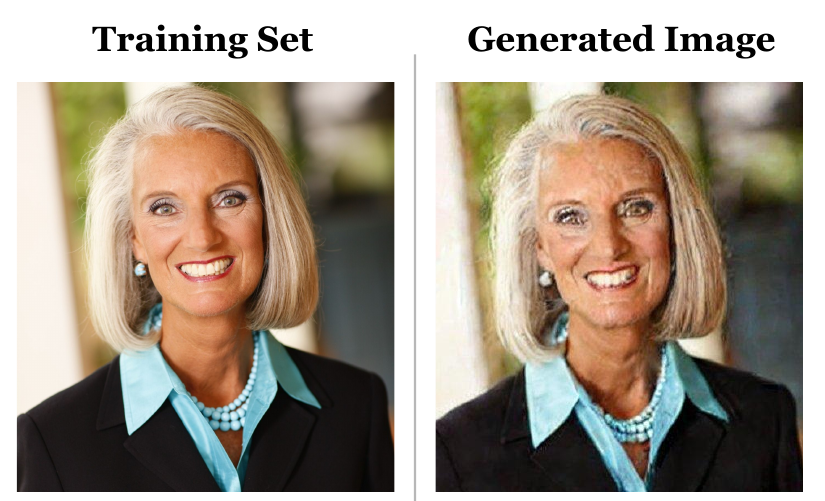}
    \caption{Memorization in a diffusion model. A training photograph (left) and a near-identical image generated by Stable Diffusion (right). Source: \cite{291199}, Figure 1.}
    \label{fig:memorization}
\end{figure}

Many applications involve data that contains sensitive information, such as medical records and financial transactions.
Releasing score estimates learned from such data may disclose information about individual training examples, even when the underlying dataset is not released. 
This concern is particularly pronounced in the context of modern machine learning. 
For example, prior work on membership inference attacks has shown that access to a trained model can reveal whether a particular observation was included in its training dataset \citep{shokri2017membership,yeom2018privacy,carlini2022membership}.
The development of score-based generative models has also raised concerns about \emph{memorization}, where the model may reproduce samples from the training dataset in response to model queries \citep{291199,somepalli2023understanding,wen2024detecting}. 
For example, Figure \ref{fig:memorization} from \cite{291199} demonstrates that prompting Stable Diffusion with a person's name can yield a near-identical reproduction of a photograph from its training dataset. 
Beyond mitigating memorization, private score estimation also enables the release of reusable statistical models learned from sensitive data. 
Such models can support downstream analysis, for example through synthetic data generation, without requiring access to original observations \citep{ghalebikesabi2023differentially}.
These concerns motivate us to develop score estimation methods that achieve high accuracy while providing rigorous privacy guarantees for individual training observations.  

\emph{Differential privacy} (DP) provides a rigorous mathematical framework for quantifying privacy guarantees \citep{dwork2006calibrating,dwork2014algorithmic}.
Informally, a randomized procedure is differentially private if changing a single observation in its input dataset has only a limited effect on the distribution of its output. 
This stability limits the information that the output can reveal about an individual observation while allowing the procedure to learn population-level patterns.
DP has become a widely recognized standard for providing rigorous privacy guarantees in statistical data analysis. 
Developing score estimation methods with rigorous accuracy and DP guarantees is therefore particularly important in applications involving sensitive data, where reliable statistical inference must be balanced with the protection of individual information.

Prior theoretical work on non-private score estimation has developed estimation methods, characterized their statistical limits, and analyzed their implications for score-based generative models \citep{block2020generative,oko2023diffusion,chen2023score,han2024neural,wibisono2024optimal,zhang2024minimax,dou2024optimal,pmlr-v291-gatmiry25b,pmlr-v291-chen25e,cai2025minimax,pmlr-v336-chewi26a,yu2026diffusion}. 
Recent work has also studied DP score estimation in the context of score-based generative modeling.
In particular, \citet{dockhorn2023differentially} train diffusion models using differentially private stochastic gradient descent \citep[DP-SGD,][]{abadi2016deep}, and introduce noise multiplicity to improve training without incurring additional privacy cost.
In their framework, privacy loss accumulates across training iterations, so a fixed privacy budget constrains the number of iterations when the noise scale and sampling rate are held fixed.
\cite{ghalebikesabi2023differentially} take a different perspective and show that privately fine-tuned diffusion models can generate synthetic images that support accurate downstream classification.
Subsequent work improves training efficiency through latent representations and selective fine-tuning \citep{liu2024differentially}, as well as low-rank adaptation of pretrained diffusion models \citep{tsai2025differentially}. 
Although these works demonstrate the feasibility and empirical effectiveness of private diffusion models, theoretical understanding of their privacy-utility trade-offs remains limited.
\citet{huang2025diffusion}  
take a step toward establishing a theoretical foundation for private score learning by analyzing the convergence rate of DP-SGD.
Their analysis takes an optimization perspective, but 
does not analyze the statistical limits of DP score estimation.

Motivated by these developments, in this work we address the following problems: 

\begin{center}
	\emph{What are the statistical limits of differentially private score estimation?  Can we develop efficient algorithms that attain these limits? } 
\end{center}

Characterizing the statistical limits of private score estimation clarifies the cost of privacy and helps practitioners make informed decisions about the trade-off between privacy and accuracy.

\subsection{Our contributions}

We summarize our main contributions below.

\paragraph{Minimax optimal sampling-based DP score estimator.}
We develop a score estimator based on quadratic basis splines. 
To ensure privacy, we use the exponential mechanism \citep[e.g.,][]{mcsherry2007mechanism}
to define a sampling distribution over the spline coefficients, and propose to approximately sample from this distribution using projected Langevin Monte Carlo \citep{bubeck2018sampling}. 
The resulting algorithm satisfies $(\alpha,\beta)$-DP and runs in time polynomial in sample size $n$. 
For sub-Gaussian target distributions with locally $\gamma$-H\"older-continuous scores, we further show that the expected density-weighted squared estimation error is at most a constant times $\log^2(n) \cdot\max( n^{-2\gamma / (d + 2\gamma +2)}, \, (n \alpha)^{-2\gamma / (d + \gamma +1)} )$.  
We also establish a minimax lower bound for the same class of distributions that matches our upper bound up to polylogarithmic factors in its dependence on the sample size and privacy parameter.

\paragraph{Application to diffusion models.}
We apply our results on DP score estimation to build a private diffusion model.
Specifically, at each step of the reverse diffusion process, we combine our DP score estimator with a DP histogram to estimate the score of the corresponding noise-perturbed distribution.
These intermediate score estimates are obtained by post-processing the same jointly private releases, so the overall privacy guarantee does not deteriorate as the number of denoising steps increases. 
We further show that when applied to Denoising Diffusion Probabilistic Models (DDPMs) \citep{ho2020denoising} and equipped with an appropriately chosen discretization scheme, 
these score estimates lead to a diffusion model whose output distribution attains the minimax rate up to polylogarithmic factors in its dependence on the sample size and privacy parameter, as measured by the total variation distance between its output distribution and the target distribution.

\paragraph{Simulation studies.}
We evaluate the proposed methods on both synthetic and real-world datasets, covering both clean score estimation and diffusion model applications. 
For the distributions considered, our method's observed risk decays at the predicted minimax rate or faster. 
We also provide partial empirical support for our method's privacy protection by showing that its output is less sensitive to changes in a single data point than that of the non-private baseline.

\subsection{Other related works}

We review in this section literature on DP distribution estimation and non-private score estimation. 
We also briefly discuss other theoretical advances for diffusion models. 

\paragraph{Differentially private distribution estimation.}
Several works have studied differentially private density estimation as an approach to private distribution estimation. 
In particular, \cite{wasserman2010statistical} show that, for Lipschitz densities supported on $[0,1]^d$, a privatized histogram attains an expected squared $L^2$ error of order $n^{-2/(d+2)}$ when the privacy parameters are fixed. 
Allowing the privacy parameters to vary with the sample size, \cite{barber2014privacy} establish the minimax rate $n^{-2/(d+2)}+(n\alpha)^{-2/(d+1)}$ under pure $\alpha$-DP for the same distribution class. Their rate is also attained by a private histogram estimator.
 \cite{lalanne2023cost} study one-dimensional periodic Sobolev densities and establish matching minimax lower and upper bounds under both pure DP and zero-concentrated DP. 
\cite{lalanne2024privately} later extend this analysis to periodic Sobolev classes supported on $[0, 1]^d$. 
Density estimation has also been studied under local DP, where each observation is privatized before being communicated to the analyst \citep{duchi2018minimax,butucea2020local}. 
On a separate note, complementary works develop DP mechanisms for approximating empirical query functions \citep{wagner2023fast,liu2024kernel}.  

Density estimation error alone does not generally control score estimation error, see Section 1.1 of \cite{sriperumbudur2017density} for a discussion on this point. 
Moreover, computationally efficient density estimation does not automatically yield efficient sampling \citep{talwar2019computational},  
whereas our DP diffusion model provides an efficient sampling algorithm.

\paragraph{Non-private score estimation.}
Non-private score estimation has been studied through both parametric and nonparametric approaches \citep{hyvarinen2005estimation,sriperumbudur2017density,sutherland2018efficient,zhou2020nonparametric}.
Recent theoretical work has investigated the statistical limits of score estimation.
In particular, \cite{wibisono2024optimal} establish the minimax rate $n^{-2/(d+4)}$ (up to polylogarithmic factors) for sub-Gaussian distributions with Lipschitz-continuous scores under density-weighted squared error. 
\cite{lewis2025learning} establish minimax score estimation rates for one-dimensional log-concave distributions, and show that combining shape constraints with smoothness can yield faster rates than smoothness alone.
A complementary line of work studies score estimation for Gaussian-perturbed distributions in the context of diffusion models.
For example, \cite{zhang2024minimax} derive estimation error upper bounds for the scores of Gaussian convolved distributions for sub-Gaussian targets. 
\cite{dou2024optimal} establish sharp minimax rates across Gaussian noise levels for compactly supported H\"older densities bounded above and away from zero.
These analyses also connect score estimation to distribution estimation through diffusion models under suitable density regularity assumptions.
More recently, \citet{yu2026diffusion} extend the analysis to target distributions with exponential and polynomial tails, deriving tail-dependent score estimation and sampling guarantees.

\paragraph{Other theoretical advances for diffusion models. }
Various other results have been established to explain
the overall success of diffusion models, covering topics including convergence rates \citep{lee2023convergence,chen2023sampling,benton2024nearly,li2024accelerating,wu2024stochastic,gupta2025faster,li2025unified,li2025odt,huang2025convergence,huang2025fast,li2025faster}, conditional sampling \citep{chidambaram2024does,wu2024theoretical,bradley2024classifier,li2025provable,jiao2025towards,guo2026conditional,cao2026robustness}, and applications in statistical or optimization settings \citep{montanari2023posterior,chung2023diffusion,li2024diffusion,chen2025diffusion,mei2025deep,jiao2026provable}. 
We skip a detailed discussion here for compactness, and refer interested readers to \cite{chen2024overview} for a summary of recent theoretical developments.

\subsection{Organization}

The remainder of this paper is organized as follows.
In Section \ref{sec:preliminaries}, we review the relevant background on diffusion models, differential privacy, and basis splines.
In Section \ref{sec:DP-score-estimation} we present our sampling-based DP score estimator. 
We apply our results to construct a private diffusion model in Section \ref{sec:DM-application}, and present numerical experiments that support our theoretical findings in Section \ref{sec:experiments}. 
In Section \ref{sec:discussion}, we discuss our results and outline directions for future research.

\subsection{Notation}

For $n \in \N_+$, we define the set $[n] = \{1, 2, \ldots, n\}$. 
For $x \in \R$, we denote its positive part by $x_+ = \max\{x, 0\}$.  
For $\mu \in \R^d$ and $r > 0$, we define $\cB(\mu, r) = \{x \in \R^d: \|x - \mu\|_2 \leq r\}$. 
For a matrix $A$, we denote by $\lambda_{\max}(A)$ its largest eigenvalue, and $\lambda_{\min}(A)$ its smallest eigenvalue. 
For a convex set $K$ and a vector $x$, we denote by $\Proj_K(x)$ the projection of $x$ onto $K$. 
For two positive sequences $\{u_n\}_{n \geq 1}$ and $\{v_n\}_{n \geq 1}$,
the notation $v_n=\widetilde O(u_n)$ (respectively $\widetilde\Omega(u_n)$) means that there exist numerical constants $c\in\mathbb R$ and $C>0$ such that, for all sufficiently large $n$, $v_n\leq C u_n(\log u_n)^c$ (respectively $v_n\geq C u_n(\log u_n)^c$).  
We also say $v_n = \widetilde{\Theta}(u_n)$ if $v_n = \widetilde{O}(u_n)$ and $v_n = \widetilde{\Omega}(u_n)$. 
For a function $f$ and a probability distribution $\rho$, we denote by $\|f\|_{\rho} =( \int \|f(x)\|_2^2 \,\rho(x)\,\dd x)^{1/2}$ the
weighted $L^2$ norm. 
For two $k$-dimensional vectors $u$ and $v$, we denote by $\Ham(u,v)=\sum_{j=1}^k \mathbbm{1}_{u_j\neq v_j}$ their Hamming distance.

\section{Preliminaries}
\label{sec:preliminaries}

In this section, we review the relevant background on diffusion models, differential privacy, and quadratic basis splines. 

\subsection{Diffusion models}
\label{sec:diffusion-models}

In this section, we introduce the basics of diffusion models \citep{ho2020denoising,song2021denoising}. 
A diffusion model is a generative model trained on samples from a target distribution $\rho_0$ to generate new samples from that distribution. 
At a high level, a diffusion model consists of a \emph{forward process} and a \emph{reverse process}, which we separately discuss below.

\paragraph{Forward process.}
In the forward process of a diffusion model, noise is progressively added to the target distribution, transforming it into a noise distribution that is easy to sample from. 
Specifically, the forward process consists of a sequence of $d$-dimensional random vectors $\bar Y_0 \to \bar Y_1 \to \ldots \to \bar Y_{N + 1}$ defined as follows:
\begin{align*}
	& \bar Y_{0} \sim \rho_0, \\
	& \bar Y_{k + 1} = e^{-\bar \Delta_k} \bar Y_k + \sqrt{1 - e^{-2 \bar\Delta_k}}\, G_k, \qquad k = 0, 1, \ldots, N,  
\end{align*}
where $N \in \N_+$ denotes the number of diffusion model steps. 
In the above display, $G_0, G_1, \ldots, G_{N} \sim_{\text{i.i.d.}} \normal (0, I_d)$ are independent of $\bar Y_0$, and $(\bar\Delta_k)_{k = 0}^{N}$ stands for the positive step sizes of the forward process. 
A simple induction argument yields 
\begin{align}
\label{eq:forward}
\begin{split}
	& \bar Y_k = e^{-\bar  t_k} \bar Y_0 + \sqrt{1 - e^{-2 \bar t_k}}\, G_k', \qquad k = 0, 1, \ldots, N + 1\\
	& \bar t_k = \sum_{j = 0}^{k - 1} \bar \Delta_j, \qquad (\bar Y_0, G_k') \sim \rho_0 \otimes \normal (0, I_d). 
\end{split}
\end{align}
Let $\rho_k$ denote the marginal distribution of $\bar Y_k$. By definition, at initialization $\bar Y_0$ follows the target distribution $\rho_0$.  
Let $T=\bar  t_{N + 1}$ denote the total time horizon of the forward process.
By \cref{eq:forward}, we see that the law of $\bar Y_{N+1}$ converges in distribution to a standard Gaussian distribution as $T \to\infty$.

\paragraph{Reverse process.}
The reverse process, on the other hand, approximately transforms noise into samples from the target distribution. 
Specifically, it is a Markov chain $Y_0 \to Y_1 \to \ldots \to Y_N$ whose marginal distributions approximately match those of the forward process in reverse order, with $Y_k \overset{d}{\approx} \bar Y_{N+1-k}$ for $k=0,1,\ldots,N$. 
To improve numerical stability, the reverse process usually ends at $Y_N \overset{d}{\approx} \bar Y_1 \sim \rho_1$, rather than continuing to the target distribution $\rho_0$.
As for the initialization of the reverse process, when $T =\bar t_{N + 1}$ is sufficiently large, recall $\bar Y_{N + 1} \overset{d}{\approx} \normal(0,I_d)$. 
We therefore initialize the reverse process by sampling from the standard Gaussian distribution.

The reverse process of a diffusion model can be formulated in several different ways. 
In the following, we present a specific construction related to DDPMs \citep{ho2020denoising}: 
\begin{align}
\label{eq:DDPM}
\begin{split}
	& Y_0 \sim \normal (0, I_d), \\
	& Y_{k + 1} = e^{\Delta_{k}} Y_k + 2 (e^{\Delta_{k}} - 1) \hat s_k (Y_k) + \sqrt{e^{2\Delta_{k}} - 1} \, W_k, \qquad k = 0, 1, \ldots, N - 1.
\end{split}
\end{align}
In the above display, $W_0, W_1, \ldots, W_{N - 1} \sim_{\text{i.i.d.}} \normal (0, I_d)$ are independent of $Y_0$, and $\Delta_k = \bar \Delta_{N - k}$. 
For $k=0, 1, \ldots,N-1$, the function $\hat s_k: \R^d \to \R^d$ is an estimator of $\nabla\log\rho_{N+1-k}$, the true score function of $\rho_{N + 1 -k}$.
We define $t_k = \sum_{j = 0}^{k - 1} \Delta_j$ with $t_0 = 0$. 
Observe that for $Y \sim \rho_{N + 1 - k}$, it holds that $Y \overset{d}{=} e^{-T + t_k} \,\bar Y_0 + \sqrt{1 - e^{-2T + 2t_k}}\, W$ for $(\bar Y_0, W) \sim \rho_0 \otimes \normal (0, I_d)$.  
In summary, to generate new samples with DDPM, we first sample $Y_0$ from the standard Gaussian distribution, and then sequentially pass it through $N$ denoising steps according to \cref{eq:DDPM}. 
Note that only the reverse process is implemented during sample generation.
The updating coefficients in \cref{eq:DDPM} arise from the exponential integrator solver \citep{zhang2022fast} and have been widely adopted for diffusion model sampling. 
We expect our results to extend to most reasonable numerical solvers, but we focus on the exponential integrator here for simplicity of exposition.
We also expect our results to extend to other diffusion model reverse processes beyond DDPM.

Prior work has shown that diffusion models converge to the target distribution $\rho_0$ at polynomial rates. We next present a representative result for DDPMs.

\begin{lem}[Theorem 2 in Appendix C of \citealt{benton2024nearly}]
\label{thm:benton}
	Suppose $T \geq 1$, $\E_{\rho_0}[\|X\|_2^2] < \infty$, and there exists $\kappa > 0$, such that $\Delta_k \leq \kappa \min \{1, T - t_{k + 1}\}$ for all $k = 0, 1, \ldots, N - 1$. 
    Denote by $\hat \rho_1$ the marginal distribution of $Y_N$ from  \cref{eq:DDPM}, then there exists a positive numerical constant $C_0$, such that 
	\begin{align*}
		\KL ( \rho_1 \parallel \hat \rho_1 ) \leq C_0 \left( \varepsilon^2_{\mathrm{score}} + \kappa^2 dN + \kappa d T + \kappa \E_{\rho_0}[\|X\|_2^2] + (d + \E_{\rho_0}[\|X\|_2^2]) e^{-2T} \right), 
    \end{align*} 
where 
	\begin{align*}
    \varepsilon^2_{\mathrm{score}} = \sum_{k = 0}^{N - 1} \Delta_k \,\E_{\rho_{N + 1 - k}}\left[ \| \nabla \log \rho_{N + 1 - k}(X) - \hat s_k(X)\|_2^2 \right]. 
    \end{align*}
\end{lem}

\subsection{Differential privacy}

We briefly review DP in this subsection.
Given $n$ observations $X_1,\ldots,X_n$, we denote the corresponding dataset by $\cD$.  
Two datasets $\cD$ and $\cD'$ are called \emph{neighboring}, denoted by $\cD \sim \cD'$, if one can be obtained from the other by replacing a single observation.  
We next define $(\alpha, \beta)$-DP.

\begin{defn}[$(\alpha, \beta)$-DP]
\label{def:DP} 
    For $\alpha > 0$ and $\beta \in [0, 1]$, 
    we say a randomized mechanism $M$ satisfies $(\alpha,\beta)$-DP if for every pair of neighboring datasets $\cD \sim \cD'$ and every measurable subset $S$ of its output space,
    \begin{align*}
        \P\big(M(\cD) \in S \big) \leq e^{\alpha} \P\big(M(\cD') \in S\big) + \beta, 
    \end{align*}
    where the probabilities are taken over the randomness of $M$. 
\end{defn}

The case $\beta=0$ corresponds to pure DP, and $\beta>0$ permits an additive relaxation and corresponds to approximate DP. 
Smaller values of $\alpha$ and $\beta$ impose stronger privacy requirements.
The following lemma lists several standard properties of DP.

\begin{lem}[Properties of DP, \citealt{dwork2006calibrating}]
\label{lem:DP-property}
\
\begin{itemize}
	\item {(Post-processing)} If $M$ satisfies $(\alpha,\beta)$-DP, then $F\circ M$ satisfies the same privacy guarantee for any measurable, possibly randomized procedure $F$ that accesses the dataset $\cD$ only through $M(\cD)$.
	\item {(Composition)} 
    Let $\mathcal{M}_i$ be an $(\alpha_i, \beta_i)$-DP mechanism for $i \in [k]$, where $k \in \mathbb{N}_+$. 
    If the mechanisms $\{\cM_i: i \in [k]\}$ are run using independent internal randomness, then the joint mechanism $\cM(\cD) = (\cM_1 (\cD), \cM_2 (\cD), \ldots \cM_k(\cD))$ is $(\sum_{i = 1}^k\alpha_i, \sum_{i = 1}^k\beta_i)$-DP. 
\end{itemize}
	
\end{lem}

We next define the DP minimax risk, which is the smallest worst-case risk achievable by estimators satisfying a specified DP guarantee.

\begin{defn}[DP minimax risk]
Let $\mathcal P$ be a class of distributions over $\cX$, let $\theta:\mathcal P\to\Theta$ denote the population quantity of interest, and let $L:\mathcal A\times\Theta\to[0,\infty]$ be a measurable loss function.
For privacy parameters $\alpha>0$ and $\beta\in[0,1]$, let $\mathcal M_{\alpha,\beta}$ denote the class of estimators that map datasets of $n$ observations in $\cX$ to $\mathcal A$ and satisfy $(\alpha,\beta)$-DP. 
The $(\alpha,\beta)$-DP minimax risk for this estimation problem is then defined as
\begin{align*}
\inf_{\widehat\theta\in\mathcal M_{\alpha,\beta}}
\sup_{P\in\mathcal P}
\mathbb E\!\left[
L\bigl(\widehat\theta(\cD),\theta(P)\bigr)
\right],
\end{align*}
where $\cD$ consists of $n$ i.i.d.~observations drawn from $P$, and the expectation is taken over both the data and the internal randomness of $\hat\theta$.
\end{defn}

\subsection{Quadratic B-splines}
\label{sec:splines}

We construct our estimators using tensor-product quadratic B-splines on $\mathbb{R}^d$ \citep{de1978practical}. We begin with the univariate cardinal quadratic B-spline
%
%
%
\begin{align}
\label{eq:B2}
    B_2(x) = \frac{1}{2} (x_+)^2 - \frac{3}{2}(x - 1)_+^2 + \frac{3}{2}(x - 2)_+^2 - \frac{1}{2} (x - 3)_+^2, \qquad x \in \R. 
\end{align}
Note that $B_2(x) \in [0, 3/4]$ for all $x \in \R$, is supported on $[0, 3]$, belongs to $C^1(\R)$, and satisfies the partition-of-unity identity, meaning that $\sum_{j \in \Z} B_2(x - j) = 1$ for all $x \in \R$. 
For a mesh width $b_n > 0$ and each $j \in \Z$, define the univariate quadratic B-spline $\psi_j(x) = B_2(b_n^{-1}x-j)$.
Note that $\mathrm{supp}(\psi_{j}) = [b_nj, b_n(j + 3)]$, and the partition-of-unity property of $B_2$ yields $\sum_{j \in \Z} \psi_j(x) = 1$ for all $x \in \R$.  
For each multi-index $\nu = (\nu_1,\ldots,\nu_d) \in \Z^d$, define $\psi_\nu(x) = \prod_{r=1}^d \psi_{\nu_r}(x_r)$ for $x = (x_1,\ldots,x_d) \in \R^d$. 
The univariate partition-of-unity property then yields $\sum_{\nu \in \Z^d} \psi_\nu(x) = 1$ for all $x \in \R^d$.

We now derive bounds on the derivatives of a univariate spline. Differentiating \cref{eq:B2} yields 
\begin{align}
\label{eq:B2p}
	B_2'(x) = (x)_+ - 3(x-1)_+ + 3(x-2)_+ - (x-3)_+.
\end{align}
For any sequence of coefficients $(c_j)_{j \in \Z}$, the spline $g_c(x) = \sum_{j \in \Z} c_j \psi_j(x)$ then satisfies
\begin{align*}
    g_c'(x) = \frac{1}{b_n} \sum_{j \in \Z}
    (c_j-c_{j-1}) B_1(x/b_n-j),
\end{align*}
where $B_1(x)=x_+-2(x-1)_++(x-2)_+$. 
Since $B_1$ is nonnegative and satisfies $\sum_{j \in \Z} B_1(x - j) = 1$ for all $x \in \R$, it follows that
\begin{align}
\label{eq:gcp-upper}
	\|g_c' \|_{\infty} \leq \frac{1}{b_n} \sup_{j \in \Z} |c_{j + 1} - c_j|.
\end{align}

\section{Differentially private score estimation}
\label{sec:DP-score-estimation}

In this section, we present our main results on DP score estimation. We formulate the problem in \cref{sec:setup} and introduce our estimator in \cref{sec:DP-score-estimator}. 
In \cref{sec:risk}, we show that our estimator attains the DP minimax rate up to polylogarithmic factors in its dependence on the sample size and privacy parameter, by deriving a risk upper bound and a nearly matching lower bound.

\subsection{Problem setup}
\label{sec:setup}

Our goal is to estimate the score function of a distribution $\rho_0$ on $\mathbb{R}^d$ from $X_1,\ldots,X_n\sim_{\text{i.i.d.}}\rho_0$ using an $(\alpha, \beta)$-DP algorithm.
Throughout, we assume that $\rho_0$ belongs to a distribution class $\cP_{\xi,L,\gamma,M}$ defined below.

\begin{defn}[Distribution class $\cP_{\xi,L,\gamma,M}$]
For $\xi, L, M > 0$ and $\gamma \in (0,1]$, a distribution on $\R^d$ with density~$\rho$ belongs to $\cP_{\xi,L,\gamma,M}$ if and only if it satisfies the following conditions: 
\begin{enumerate}
    \item $\rho$ is positive and continuously differentiable;
    \item  $\|\E_\rho[X]\|_2\le M$; 
    \item $\rho$ is $\xi$-sub-Gaussian, i.e.,
	\begin{align*}
    \E_{X \sim \rho}\left[\exp\left(\langle v,X-\E_{\rho}[X]\rangle\right)\right]\le
    \exp\left(
    \frac{\xi^2\|v\|_2^2}{2}\right),
    \qquad \mbox{for all }v\in\mathbb R^d;
	\end{align*}
    \item the score $\nabla\log\rho$ satisfies
	\begin{align*}
    \|\nabla\log\rho(x)-\nabla\log\rho(y)\|_2
    \le
    L\|x-y\|_2^\gamma,
    \qquad
    \mbox{for all }x, y \in \R^d \mbox{ with }\|x-y\|_2\le1.
	\end{align*}
\end{enumerate}
\end{defn}

Throughout this paper, we assume $\xi^2 L>2$, where the lower bound $2$ can be replaced by any numerical constant greater than $1$.
This condition ensures that $\cP_{\xi,L,\gamma,M}$ is non-empty (see \cref{lemma:cP-non-empty} in the appendix for a rigorous justification). 
When $\gamma=1$, distributions in $\cP_{\xi,L,\gamma,M}$ have Lipschitz continuous score functions, and in this setting the non-private minimax risk for score estimation is already characterized by \cite{wibisono2024optimal}.

\begin{rem}
We explain here why we consider a local H\"older smooth condition instead of a global one, and why $\gamma$ has to be no larger than 1.
When $\gamma=1$, the local and global Lipschitz conditions are equivalent with the same Lipschitz constant.  
For $\gamma \in (0, 1)$, a global H\"{o}lder condition on the score is incompatible with sub-Gaussianity.  To see this, suppose that $\|\nabla \log\rho(x)-\nabla \log\rho(y)\|_2 \le L\|x-y\|_2^\gamma$ for all $x, y \in \R^d$. Taking $y = 0$ gives $\|\nabla \log\rho(x)\|_2 \leq \|\nabla \log\rho(0)\|_2 + L\|x\|_2^\gamma$ for all $x \in \R^d$.  Integrating $\nabla\log\rho$ along a ray yields
	\begin{align*}
    \log\rho(x) - \log \rho(0) \geq 
    -\left(\|\nabla \log\rho(0)\|_2 \cdot\|x\|_2 + L\|x\|_2^{1+\gamma}\right).
	\end{align*}
Since $1 + \gamma < 2$, this lower bound implies that $\int_{\mathbb{R}^d} \exp(a\|x\|_2^2)\rho(x)\,\mathrm{d}x = \infty$ for every $a>0$, whereas sub-Gaussianity requires this integral to be finite for some $a>0$.

Lastly, if $\gamma > 1$, even the local H\"{o}lder condition forces the score to be constant. This is because, dividing the segment from $x$ to $y$ into $m$ equal subsegments and summing the local bounds gives
\[
    \|s(x)-s(y)\|_2
    \leq Lm^{1-\gamma}\|x-y\|_2^\gamma,
\]
for all sufficiently large $m$.  Taking $m \to \infty$ yields $s(x) = s(y)$ for all $x, y \in \mathbb{R}^d$.
\end{rem}
For an estimator $\hat s$ of $\nabla \log \rho_0$, we quantify the score estimation error using the loss
\begin{align}
\label{eq:R-loss}
    R(\hat s, \rho_0) = \int_{\mathbb{R}^d} \|\hat s(x) - \nabla \log \rho_0(x)\|_2^2 \,\rho_0(x) \,\mathrm{d} x. 
\end{align}
Loss \eqref{eq:R-loss} is widely used in the score estimation literature, and we refer readers to Section 1 in \cite{wibisono2024optimal} for a discussion of this choice.

\subsection{DP score estimator based on the exponential mechanism}
\label{sec:DP-score-estimator}

To ensure DP guarantees, we adopt the exponential mechanism \citep{mcsherry2007mechanism}.
Specifically, given a collection of candidate outputs and a data-dependent loss, an exponential mechanism samples candidates with weights that decrease exponentially with their loss to ensure privacy. 
At a high level, the exponential mechanism ensures DP while favoring outputs with lower values of the adopted loss, thereby balancing privacy and accuracy.

To construct our DP score estimator, we combine the standard score matching loss \citep{hyvarinen2005estimation} with the exponential mechanism to define a distribution over spline coefficients. 
We show that the induced distribution is log-concave and supported on a polytope, hence enabling approximate sampling in polynomial time via projected Langevin Monte Carlo \citep[LMC,][]{bubeck2018sampling}.

In the remainder of this section, we describe the construction of our DP score estimator. 
We begin by truncating the data and constructing the spline basis for estimation. We then derive the distribution induced by the exponential mechanism and describe how to use projected LMC to approximately sample from it.

\paragraph{Step I: Data preprocessing via truncation.} 
Recall that we observe $X_1, \ldots, X_n \sim_{\mathrm{i.i.d.}}\rho_0$. 
Let $\Omega_{R_n} = [-R_n, R_n]^d$ denote the $d$-dimensional hypercube centered at $\vec{0}_d$ with side length $2R_n$, where $R_n$ is defined in \cref{eq:cT-R} below.
To preprocess data, we define a truncation operator $\cT_{R_n}: \R^d \to \R^d$ that maps each observation into $\Omega_{R_n}$: 
\begin{align}
\label{eq:cT-R}
    \cT_{R_n}(x) =  x\mathbbm{1}\big\{ |x_i| \leq  R_n  \mbox{ for all }i \in [d]\big\}, \qquad \mbox{with } R_n = M + C \xi(\sqrt{d} + \sqrt{10\log n / c}),
\end{align}
where $c,C>0$ are numerical constants in the sub-Gaussian tail bound in \cref{lem:subgaussian-concentration}. 
By \cref{lem:subgaussian-concentration}, for every $i\in[n]$ we have
\begin{align}
\label{eq:X-neq-TX}
    \P_{\rho_0}\big(X_i \neq \cT_{R_n}(X_i ) \big) \leq n^{-10}. 
\end{align}
The remaining steps of our algorithm use the truncated observations $Z_i = \cT_{R_n}(X_i) \in \Omega_{R_n}$ for $i \in [n]$. 

\paragraph{Step II: Setting up the spline basis.}
Recall the quadratic B-splines introduced in Section \ref{sec:splines}. 
We next consider the collection of B-spline functions whose supports overlap with a high-probability set under $\rho_0$. 
We begin by defining a smooth cutoff function $\chi_{R_n}: \R^d \to [0, 1]$ with the following properties: 
\begin{align}
\label{eq:chiR}
	\chi_{R_n}(x) = 1 \mbox{ for }x \in \Omega_{R_n}, \qquad \chi_{R_n}(x) = 0 \mbox{ for }x \notin \Omega_{2R_n} \qquad \mbox{and} \qquad \|\nabla \chi_{R_n}\|_{\infty} \leq \frac{C_{\chi}}{R_n}, 
\end{align}
where $C_{\chi} > 0$ is a positive numerical constant. We refer the readers to \cref{lemma:chiR} in the appendix for a construction of such $\chi_{R_n}$. 
For each $a \in [d]$ and $\nu \in \Z^d$, we define the vector-field basis function $\Phi_{a, \nu}(x) = \chi_{R_n}(x) \psi_{\nu}(x) e_a$, 
where $e_a$ denotes the $a$-th standard basis vector in $\R^d$, and define the effective index set by
\begin{align*}
	\cJ = \Big\{(a, \nu) \in [d] \times \Z^d: \int \|\Phi_{a, \nu}(x)\|_2^2  \,\d x \neq  0 \Big\}. 
\end{align*}
Since $\chi_{R_n}$ is supported on $\Omega_{2R_n}$, the effective index set satisfies $|\cJ| \leq d\, ( 4R_n b_n^{-1} + 4 )^d$. 
For a coefficient vector $\theta = (\theta_{a, \nu})_{(a, \nu) \in \cJ} \in \R^{|\cJ|}$, letting $\theta_{a,\nu}=0$ for $(a,\nu)\notin \cJ$, we define %
\begin{align}
\label{eq:f-theta}
	f_{\theta}(x) = \sum_{a = 1}^d \sum_{\nu \in \Z^d} \theta_{a, \nu} \Phi_{a, \nu}(x). 
\end{align}
For $a \in [d]$, we denote by $f_{\theta, a}$ the $a$-th coordinate of $f_{\theta}$. Observe that 
\begin{align}
\label{eq:f-theta-g-theta}
	f_{\theta, a}(x) = \chi_{R_n}(x) g_{\theta, a}(x) \qquad \mbox{and} \qquad g_{\theta, a}(x) = \sum_{\nu \in \Z^d} \theta_{a, \nu} \psi_\nu(x). 
\end{align}
In the remainder of this section, we construct a score estimator using the basis functions indexed by $\cJ$ and the truncated data $Z_1,\ldots,Z_n$.

\paragraph{Step III: Distribution induced by the exponential mechanism.}
We next introduce the distribution over the spline coefficients induced by the exponential mechanism. 
We adopt the score matching loss \citep{hyvarinen2005estimation} to construct such a distribution. 
Specifically, for $\theta \in \R^{|\cJ|}$, the score matching loss for a single observation $z$ is given by
\begin{align}
\label{eq:score-matching-loss}
	\ell_{\theta}(z) = \frac{1}{2} \|f_{\theta}(z)\|_2^2 + \diver f_{\theta}(z). 
\end{align}
We consider a distribution over the spline coefficients supported on a polytope $\Theta_{n, \gamma}$:
\begin{align}
\label{eq:Thetan}
    \Theta_{n,\gamma} = \left\{\theta \in \mathbb{R}^{\vert{}\mathcal{J}\vert{}} : \max_{(a,\nu) \in \cJ} \vert{}\theta_{a,\nu}\vert{} \le \cR_n, \,\vert{}\theta_{a,\nu} - \theta_{a,\nu+e_q}\vert{} \le Lb_n^\gamma \text{ for all } q \in [d] \mbox{ and } (a,\nu),(a,\nu+e_q)\in\mathcal J \right\}, 
\end{align}
where recall that $b_n > 0$ is the mesh width of the quadratic B-splines, and $\cR_n$ is defined in \cref{eq:piD} below. 
We show in Appendix \ref{sec:proof-sketch} that this polytope encodes smoothness constraints on the score function.

\begin{figure}[hbt!]
	\centering \includegraphics[width=0.45\textwidth]{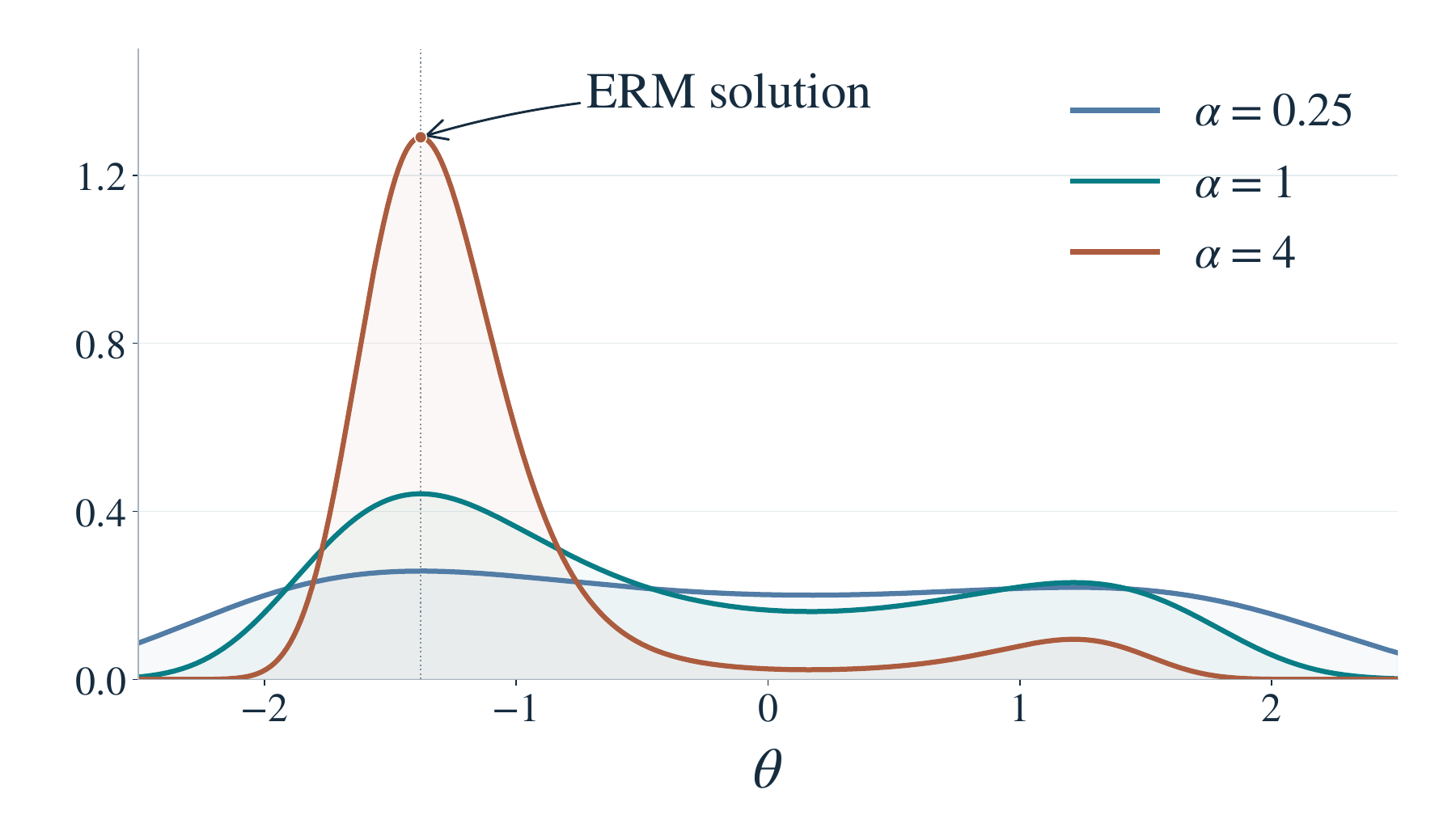}
	\caption{Illustration of distributions induced by the exponential mechanism. }
	\label{fig:exponential-mechanism}
\end{figure}

We next state the distribution induced by the exponential mechanism. 
Specifically, for a dataset $\cD$ consisting of samples $X_1, \ldots, X_n$, we define the following distribution over $\Theta_{n, \gamma}$: 
\begin{align}
\label{eq:piD}
\begin{split}
	& \pi_\cD(\theta) \propto \mathbbm{1}\{\theta \in \Theta_{n, \gamma}\} \exp \Big( -\frac{\alpha}{4Q_{n, \gamma}} \sum_{i = 1}^n \ell_{\theta}(Z_i) \Big), \\
	& Q_{n, \gamma} =d \cR_n^2 + d  Lb_n^{\gamma -1} \qquad \mbox{and} \qquad \cR_n = C_0 L\bigl(1+\xi\sqrt d+3\sqrt d\,R_n + M\bigr), 
\end{split}
\end{align}
where $Z_i = \cT_{R_n}(X_i)$ for $i \in [n]$ and $C_0>0$ is a numerical constant from \cref{lem:score-upper-bound}. 
The exponential mechanism \eqref{eq:piD} defines a distribution from which we can sample to obtain a private output. 
The privacy parameter $\alpha$ controls the shape of this distribution and the resulting trade-off between privacy and accuracy. 
In particular, larger values of $\alpha$ yield distributions that are more concentrated around the empirical risk minimizer, preserving accuracy at the expense of privacy. 
Conversely, smaller values of $\alpha$ produce flatter distributions, hence strengthening privacy at the expense of accuracy. 
See Figure \ref{fig:exponential-mechanism} for an illustration of distributions induced by the exponential mechanism. 

As shown in \cref{lem:exponential-mechanism} in the appendix, releasing a single exact draw from $\pi_\cD$ satisfies $(\alpha,0)$-DP. 
In practice, however, exact sampling is rarely feasible, so we rely on approximate sampling methods such as Markov chain Monte Carlo.
To improve the numerical stability of approximate sampling, we include a ridge regularization term in the log density of $\pi_{\cD}$ and propose to sample from the resulting distribution: 
\begin{align}
\label{eq:pi-tilde}
	\tilde\pi_{\cD} (\theta) \propto \mathbbm{1}\{\theta \in \Theta_{n, \gamma}\} \exp \Big( -\frac{\alpha}{4Q_{n, \gamma}} \sum_{i = 1}^n \ell_{\theta}(Z_i) - \frac{\min (r_n, 2\cR_n^{-2}\log(1 + \Upsilon_{\alpha, \beta} / 2) )}{2|\cJ|} \|\theta\|_2^2 \Big). 
\end{align}
In the above display, $r_n = \max( n^{-2\gamma / (d + 2\gamma +2)}, \, (n \alpha)^{-2\gamma / (d + \gamma +1)})$ and $\Upsilon_{\alpha, \beta} = \beta / (1 + e^{\alpha})$, where recall $\alpha$ and $\beta$ denote the privacy parameters.

\paragraph{Step IV: Approximate sampling using projected LMC.}
We next show that $\tilde\pi_{\cD}$ defined in \cref{eq:pi-tilde} is strongly log-concave, and present a projected LMC algorithm to approximately sample from this distribution.
To see log-concavity, we define the following $d \times |\cJ|$ matrix-valued function: 
\begin{align*}
	\Phi(z) = \Big[ \Phi_1(z) \mid \Phi_2(z) \mid \, \ldots\, \mid  \Phi_{|\cJ|}(z)  \Big], \qquad z \in \R^d,
\end{align*} 
where $\{\Phi_i: i \in [|\cJ|]\}$ is a permutation of $\{\Phi_{a, \nu}: (a, \nu) \in \cJ\}$
so that $f_{\theta}(z) = \Phi(z)\theta$ for all $z \in \R^d$. 
We also define $h_k(z) = \diver \Phi_k (z)$ for $k \in [|\cJ|]$ and $h(z) = (h_1(z), \,h_2(z),\, \ldots, \, h_{|\cJ|}(z)  )^{\top} \in \R^{|\cJ|}$.
Note that $\ell_{\theta}(z) =  \theta^{\top} \Phi(z)^{\top} \Phi(z) \theta / 2 + h(z)^{\top} \theta$, hence $\tilde\pi_{\cD}$  admits the following representation: 
\begin{align*}
	& \tilde \pi_\cD(\theta) \propto \mathbbm{1}\{\theta \in \Theta_{n, \gamma}\} \exp \Big( -\frac{1}{2} \theta^{\top} \Sigma_\cD \theta - \omega_\cD^{\top} \theta \Big), \\
	& \Sigma_\cD = \frac{\alpha}{4Q_{n, \gamma}} A_\cD + \frac{\min (r_n, 2\cR_n^{-2}\log(1 + \Upsilon_{\alpha, \beta} / 2) )}{|\cJ|} I_{|\cJ|} \qquad \mbox{and} \qquad \omega_\cD = \frac{\alpha}{4Q_{n, \gamma}} h_\cD,  
\end{align*}
where $A_\cD = \sum_{i = 1}^n \Phi(Z_i)^{\top} \Phi(Z_i) \succeq 0$ and $h_\cD = \sum_{i = 1}^n h(Z_i)$. 
Since $\Sigma_\cD$ is positive definite, we conclude that the distribution $\tilde \pi_\cD$ is strongly log-concave on $\Theta_{n, \gamma}$. 

We next present the projected LMC algorithm to approximately sample from $\tilde \pi_{\cD}$ in \cref{alg:projected_LMC}, where $\Proj_{\Theta_{n,\gamma}}$ denotes projection onto the convex set $\Theta_{n,\gamma}$.  

\begin{algorithm}
\caption{Projected LMC}
\label{alg:projected_LMC}
\begin{algorithmic}[1]
\State \textbf{Input:} Number of iterations $\cL$, step size $\delta$; 
\State Initialize $\theta_0 = \vec{0}_{|\cJ|}$; 
\For{$t = 1, 2, \ldots, \cL$}
\State Sample $g_t \sim \mathcal{N}(0, I_{|\cJ|})$ that is independent of the past iterates; 
\State Update $\theta_{t} = \Proj_{\Theta_{n, \gamma}} (\theta_{t - 1} - \frac{\delta}{2} (\Sigma_\cD \theta_{t - 1} + \omega_\cD) + \sqrt{\delta} g_t)$; 
\EndFor \\
\Return $\theta_\cL$. 
\end{algorithmic}	
\end{algorithm}

Let $\hat\theta_{\mathrm{LMC}} = \theta_{\cL}$ denote the output of \cref{alg:projected_LMC}. We use these spline coefficients to define the score estimator $f_{\hat\theta_{\mathrm{LMC}}}$ (recall this is from \cref{eq:f-theta}).
In the next section, we show that $f_{\hat\theta_{\mathrm{LMC}}}$ satisfies $(\alpha,\beta)$-DP and achieves the DP minimax risk up to polylogarithmic factors.

\subsection{DP minimax risk}
\label{sec:risk}

In this section, we upper bound the risk of our DP score estimator $f_{\hat\theta_{\mathrm{LMC}}}$ from \cref{sec:DP-score-estimator}.
We also establish a nearly matching minimax lower bound in its dependence on the sample size and privacy parameter. 
We first present the risk upper bound in \cref{thm:DP-score-upper-bound} below. 
In this theorem, we use the bound in \cite{bubeck2018sampling} to set the number of projected LMC iterations.

\begin{thm}
\label{thm:DP-score-upper-bound}	

Fix $\alpha > 0$ and $\beta \in (0,1]$.
For $\rho_0 \in \cP_{\xi, L, \gamma, M}$ and $X_1, \ldots, X_n \sim_{\text{i.i.d.}} \rho_0$, denote by $\hat\theta_{\mathrm{LMC}}$ the output of \cref{alg:projected_LMC} with the following parameter choices: 
\begin{align*}
	& \delta = \tilde \Theta\left( \,\frac{|\cJ|\cR_n^2}{\cL}\, \right ), \\
	& \cL = \tilde \Omega \Big( \,\frac{|\cJ|^3 \cR_n^6 \max \big( |\cJ|, \,\bar \lambda\sqrt{|\cJ| } \cR_n L b_n^{\gamma}, \, \bar \lambda{|\cJ|} \cR_n^2 + \sqrt{|\cJ|} \cR_n n \alpha \sqrt{2d} Q_{n, \gamma}^{-1} b_n^{-1} \big)^{12}}{\min(r_n,\Upsilon_{\alpha,\beta}/2)^{12} L^6 b_n^{6\gamma}}  \Big), \\
	& \bar \lambda = \frac{\min (r_n, 2\cR_n^{-2}\log(1 + \Upsilon_{\alpha, \beta} / 2) )}{|\cJ|} + \frac{nd\alpha}{4Q_{n, \gamma}} \qquad \mbox{and} \\
    & b_n = \max(n^{-1 / (d + 2\gamma +2)}, \, (n \alpha)^{-1 / (d + \gamma +1)}) \cdot \sqrt{\log n}.
\end{align*}
Then, for all sufficiently large $n$ we have
\begin{align*}
\E\big[ R\big( f_{\hat\theta_{\mathrm{LMC}}}, \,\rho_0 \big) \big] \leq C_{L, \xi, d, \gamma, M}  \max\big( n^{-2\gamma / (d + 2\gamma +2)}, \, (n \alpha)^{-2\gamma / (d + \gamma +1)} \big) \cdot (\log n)^2. 
\end{align*}
In the above display, $C_{L, \xi, d, \gamma, M}$ is a positive constant that depends only on $(L, \xi, d, \gamma, M)$. 
In addition, the score estimate $\cD \mapsto f_{\hat\theta_{\mathrm{LMC}}}$ is $(\alpha, \beta)$-DP. 

\end{thm}

\begin{proof}[Proof of \cref{thm:DP-score-upper-bound}]
	We give a proof sketch of \cref{thm:DP-score-upper-bound} in Appendix \ref{sec:proof-sketch} and present a full proof in Appendix \ref{proof:thm:DP-score-upper-bound}.
\end{proof}

\begin{rem}
\label{rem:LMC}

The privacy parameter $\beta$ does not enter the risk upper bound in \cref{thm:DP-score-upper-bound}, but it affects the bound on computational cost. 
As is common in MCMC complexity analyses, the LMC iteration bound in \cite{bubeck2018sampling} involves unspecified numerical constants.
Our numerical experiments in Section \ref{sec:experiments} therefore use practical tuning methods and convergence diagnostics to select the number of LMC iterations, without formally certifying the prespecified privacy level.  
In fact, sufficient mixing guarantees privacy.  
Specifically, let $\hat{\pi}_{\mathcal{D}}$ denote the conditional distribution of the LMC output given dataset $\mathcal{D}$. 
By \Cref{lem:privacy-approximate-sampling}, the condition $\sup_{\mathcal{D}} \mathrm{TV}(\hat{\pi}_{\mathcal{D}}, \pi_{\mathcal{D}}) \leq (1 + e^{\alpha})^{-1}\beta$
ensures that $\cD \mapsto f_{\hat\theta_{\mathrm{LMC}}}$ is $(\alpha, \beta)$-DP.  Lemmas~\ref{lemma:ridge-regularized-distribution} and \ref{thm:sampling} together establish this condition for the parameter choices in \Cref{thm:DP-score-upper-bound}, while the iteration upper bound involves unspecified numerical constants from \cite{bubeck2018sampling}.

\end{rem}


We next establish a minimax lower bound for private score estimation that nearly matches the upper bound in its dependence on the sample size and privacy parameter.
For $\alpha, \beta > 0$, we let $\mathcal{M}_{\alpha,\beta}$ denote the class of $(\alpha,\beta)$-DP estimators.

\begin{thm}
\label{thm:score-holder-lower}
Fix $\alpha>0$ and $0<\beta\leq \min (\alpha, 1)$.
Assume $L \xi^2 > 2$ and $M \geq \sqrt{d} \xi$. 
Recall $R$ is defined in \cref{eq:R-loss}. 
Then for all sufficiently large~$n$, it holds that
\begin{align*}
    \inf_{\widehat s \in \mathcal{M}_{\alpha, \beta}} \sup_{\rho\in\mathcal P_{\xi, L, \gamma, M}}
    \mathbb E_{\rho^{\otimes n}}
    \left[
        R(\hat s, \rho)
    \right] \geq  C_{L, \xi, d, \gamma}
    \max\big( n^{-2\gamma / (d + 2\gamma +2)}, \, (n \alpha)^{-2\gamma / (d + \gamma +1)} \big),
\end{align*}
where $\hat s$ is any $(\alpha, \beta)$-DP function of  $X_1, \ldots, X_n \sim_{\text{i.i.d.}} \rho$, and $C_{L, \xi, d, \gamma}$ is a positive constant that depends only on $(L, \xi, d, \gamma)$.  
\end{thm}

\begin{proof}[Proof of \cref{thm:score-holder-lower}]
	We prove \cref{thm:score-holder-lower} in Appendix \ref{proof:thm:score-holder-lower}. 
\end{proof}

Theorems~\ref{thm:DP-score-upper-bound} and \ref{thm:score-holder-lower} together establish that our private score estimator attains the minimax rate
\[
    \max\big( n^{-2\gamma / (d + 2\gamma +2)}, \, (n \alpha)^{-2\gamma / (d + \gamma +1)} \big)
\]
up to a polylogarithmic factor.  
The first term is the non-private minimax rate, which our proof of \cref{thm:DP-score-upper-bound} also suggests that empirical risk minimization (ERM) with loss \eqref{eq:score-matching-loss} attains this rate up to the same polylogarithmic factor.  
Note that this non-private ERM estimator can be computed efficiently by solving a convex optimization problem.  
This result complements that of \cite{wibisono2024optimal}, which establishes a risk upper bound for a kernel-based estimator under $\gamma$-H\"older continuity of the score, without a corresponding minimax lower bound for $\gamma \in (0, 1)$.

These two theorems also quantify the cost of privacy.  
When the privacy budget $\alpha = \tilde \Omega( n^{-(\gamma+1)/(d+2\gamma+2)})$, enforcing privacy preserves the non-private minimax rate up to polylogarithmic factors.  
Under stricter privacy constraints with $\alpha = \tilde O( n^{-(\gamma+1)/(d+2\gamma+2)})$, the privacy term dominates, and the minimax risk scales as $(n \alpha)^{-2\gamma / (d + \gamma +1)}$ up to polylogarithmic factors.



\section{Application to diffusion models}
\label{sec:DM-application}

In this section, we apply our score estimation results to construct a DP diffusion model, focusing on DDPMs \citep{ho2020denoising}.  
As discussed in Section \ref{sec:diffusion-models},
constructing a diffusion model requires estimating the score functions $\{\nabla\log\rho_{N + 1 - k}: k=0,1,\ldots,N-1\}$ of a sequence of intermediate distributions. 
A direct approach is to estimate each score function separately using our DP score estimator in \cref{sec:DP-score-estimator}, and then combine the resulting estimates to form a DP diffusion model. 
This approach, however, would incur a privacy cost that grows with the number of denoising steps $N$, which can be large in practice.  
To avoid accumulating privacy loss, we note that the score functions at different diffusion times are temporally correlated, as they correspond to different Gaussian-convolved versions of the same distribution.
We propose to exploit this temporal structure to construct score estimates without accumulating the privacy cost.

The rest of this section proceeds in two parts.
\Cref{subsec:private-histogram} presents the construction of our DP scores obtained from a private correction. 
\Cref{subsec:dpddpm} shows that the output distribution of the resulting DDPM estimates the target distribution at the minimax rate in total variation distance under privacy constraints, up to polylogarithmic factors.

\subsection{Score estimation with private correction}
\label{subsec:private-histogram}

We first discuss the temporal correlation of the target score functions at different diffusion times. 
Note that for $k = 0, 1, \ldots, N-1$,
the intermediate score function $\nabla \log \rho_{N + 1 - k}$ can be expressed as
\begin{align}
\label{eq:score-identity}
	\nabla \log \rho_{N + 1 - k}(x) = -\frac{x}{1 - e^{-2T + 2t_k}} +  \frac{e^{-T + t_k}}{1 - e^{-2T + 2t_k}}\, \frac{\int y \rho_0(y) \varphi_{k}(x - e^{-T + t_k} y) \, \dd y}{\int \rho_0(y) \varphi_{k}(x - e^{-T + t_k}  y) \, \dd y}, 
\end{align}
where $\varphi_{k}$ denotes the density of a Gaussian distribution with mean zero and covariance $(1 - e^{-2T + 2t_k})I_d$.  Replacing $\rho_0$ with a private density estimator therefore leads to intermediate score estimators without further access to the data.

We construct this density estimator by refining a DP histogram using our private estimate of $\nabla \log \rho_0$.    
Our construction exploits the smoothness of the log-density while preserving a tractable form. 
In particular, we can evaluate the score estimates at each diffusion time as finite sums involving one-dimensional Gaussian densities and cumulative distribution functions (CDFs), our DP score estimate, and the DP histogram.
We present the construction below in three steps.

\subsubsection*{Step I: DP histogram}

We begin by estimating $\rho_0$ using a DP histogram on a spatial grid similar to that used to construct the B-spline estimator from Section \ref{sec:DP-score-estimator}.
Specifically, for $\nu \in \Z^d$ we define 
\begin{align*}
	A_{\nu} = \prod_{r = 1}^d \left[\bar b_n \nu_r, \, \bar b_n (\nu_r + 1) \right) \cap \Omega_{R_n} \qquad \mbox{and} \qquad \cH = \left\{ \nu \in \Z^d: |A_{\nu}| > 0 \right\},  
\end{align*}
where $|A_{\nu}|$ denotes the Lebesgue volume of $A_{\nu}$, and $\bar b_n=\max\{n^{-1/(d+2\gamma+2)},\,(n\alpha / 2)^{-1/(d+\gamma+1)}\}\log^{1/2}(n)$ is chosen similarly as $b_n$ in \cref{thm:DP-score-upper-bound} but with $\alpha$ replaced by $\alpha / 2$. 
This choice ensures that the corresponding score estimate is $(\alpha/2, \beta)$-DP, and with a private histogram that satisfies $(\alpha/2, 0)$-DP, we can apply the composition lemma (\Cref{lem:DP-property}) to ensure that the overall procedure is $(\alpha, \beta)$-DP. 
If $R_n/\bar b_n$ is an integer, points on the positive outer faces of $\Omega_{R_n}$ may belong to no cell in $\cH$.
In this case, we assign these points to the adjacent cells inside $\Omega_{R_n}$.  
Observe that $|\cH|\leq ({2R_n}{\bar b_n^{-1}}+2)^d$,
where we recall $R_n$ is from \cref{eq:cT-R}.   

With $X_1, \ldots, X_n \sim_{\text{i.i.d.}} \rho_0$, $\cD$ being the dataset consisting of these $n$ samples and $Z_i=\cT_{R_n}(X_i)\in\Omega_{R_n}$, for each $\nu \in \cH$ we define the empirical probability of cell $A_{\nu}$ by $\bar p_{\cD,\nu}^\hist=n^{-1}\sum_{i=1}^n\mathbbm 1\{Z_i\in A_\nu\}$. 
To induce privacy, 
we let $\zeta = (\zeta_\nu)_{\nu \in \cH} \in \R^{|\cH|}$ be a vector of independent Laplace random variables with scale $4/(n\alpha)$, and define
\begin{align}
\label{eq:private-histogram-weights}
\hat p_\cD^\hist =\Proj_{\cS_{\cH}}(\bar p_\cD^\hist+\zeta) \qquad \mbox{and} \qquad \cS_{\cH}=\Big\{p\in[0,1]^{|\cH|}:\sum_{\nu \in \cH} p_{\nu} = 1\Big\}, 
\end{align}
where $\bar p_\cD^\hist = (\bar p_{\cD, \nu}^\hist)_{\nu \in \cH} \in \R^{|\cH|}$. 
We then define our DP histogram-based density estimator as
\begin{align}
\label{eq:private-histogram-density}
 \hat\rho_\cD^\hist(x)=\sum_{\nu \in \cH}\frac{\hat p^\hist_{\cD,\nu}}{|A_\nu|}\mathbbm 1\{x\in A_\nu\}, \qquad x \in \R^d.
\end{align}
The following lemma establishes a privacy guarantee for $\hat \rho_\cD^\hist$.  

\begin{lem}
\label{lem:private-histogram}
	The density estimate $\cD \mapsto \hat \rho_\cD^\hist$ is $(\alpha / 2, 0)$-DP. 
\end{lem}

\begin{proof}[Proof of \cref{lem:private-histogram}]
	We prove \cref{lem:private-histogram} in Appendix \ref{proof:lem:private-histogram}. 
\end{proof}

\subsubsection*{Step II: Score-based refinement}
We next introduce a refinement of \eqref{eq:private-histogram-density} using our DP score. 
The intuition behind this construction is that when $x \in A_\nu$ and $A_{\nu }$ is chosen to be a sufficiently small bin, $\log \rho_0(x) \approx \log \rho_0(c_\nu) + \langle x - c_\nu, \nabla \log \rho_0(c_\nu) \rangle$, where $c_{\nu}$ denotes the center of $A_{\nu}$. We therefore tilt the histogram within each bin by the exponential of the estimated linear term and normalize to preserve each bin's mass. 

Throughout the remainder of this section, let $\hat\theta_{\mathrm{LMC}}$ denote the output of \cref{alg:projected_LMC} run with parameters ensuring $(\alpha/2,\beta)$-DP. 
Using independent intrinsic randomness for $\hat\theta_{\mathrm{LMC}}$ and the private histogram \eqref{eq:private-histogram-density}, \cref{lem:private-histogram} and the composition lemma imply that the joint release of $(f_{\hat\theta_{\mathrm{LMC}}},\hat\rho_\cD^\hist)$ is $(\alpha,\beta)$-DP.
We then use $(f_{\hat\theta_{\mathrm{LMC}}},\hat\rho_\cD^\hist)$ to construct a score-refined density estimate.
Specifically, for each $\nu\in\cH$, define 
\begin{align}
\label{eq:sc-density}
\begin{split}
& v_{\cD,\nu} = f_{\hat\theta_{\mathrm{LMC}}}(c_\nu), \qquad Z_{\cD,\nu}=\int_{A_\nu}
       \exp\big(\langle v_{\cD,\nu},y-c_\nu\rangle\big)\,\dd y,\\
& q_{\cD,\nu}(y) = \frac{\mathbbm1\{y\in A_\nu\}
       \exp\big(\langle v_{\cD,\nu}, \,y - c_\nu\rangle\big)}{Z_{\cD,\nu}} \qquad \mbox{and} \qquad \hat\rho_\cD(y)=
       \sum_{\nu\in\cH}\hat p^\hist_{\cD,\nu}q_{\cD,\nu}(y).
\end{split}
\end{align}
Observe that in the interior of $A_\nu$ we have $\nabla\log q_{\cD,\nu}=v_{\cD,\nu}$.
The normalizers $Z_{\cD,\nu}$ are finite and positive, and one can verify that $\hat\rho_\cD$ forms a probability density function. 
Note that $Z_{\cD, \nu}$ can be evaluated explicitly. 
In particular, for $\nu \in \cH$, we write $\bar A_{\nu} = \prod_{q = 1}^d [\ell_{\nu, q}, u_{\nu, q}] $, where $\bar A_{\nu}$ denotes the closure of $A_{\nu}$.
For $q \in [d]$, denote by $v_{\cD, \nu, q}$ the $q$-th coordinate of $v_{\cD, \nu}$, $c_{\nu, q}$ the $q$-th coordinate of $c_{\nu}$, and define
\begin{align*}
 z_{\cD,\nu, q}= \begin{cases}
 \displaystyle
 \frac{e^{v_{\cD,\nu,q}(u_{\nu, q}-c_{\nu,q})}
       -e^{v_{\cD,\nu,q}(\ell_{\nu, q}-c_{\nu,q})}}{v_{\cD,\nu,q}},
        & v_{\cD,\nu, q} \neq 0,\\[1ex]
 u_{\nu, q}-\ell_{\nu, q}, & v_{\cD,\nu,q}=0.
 \end{cases}
\end{align*}
Then one can verify that $Z_{\cD,\nu}=\prod_{q = 1}^d z_{\cD,\nu, q}$.
Therefore, $\hat\rho_\cD$ can be explicitly evaluated. 

\subsubsection*{Step III: Score estimation along the reverse process}
Following \cref{eq:score-identity}, we can use $\widehat{\rho}_{\mathcal{D}}$ to define the following estimator of $\nabla\log\rho_{N+1-k}$:
\begin{align}
\label{eq:sk}
	\hat s_k(x) = -\frac{x}{1 - e^{-2T + 2t_k}} +  \frac{e^{-T + t_k}}{1 - e^{-2T + 2t_k}}\, \frac{\int y\, \hat \rho_\cD(y) \varphi_{k}(x - e^{-T + t_k} y) \, \dd y}{\int \hat\rho_\cD(y) \varphi_{k}(x - e^{-T + t_k}  y) \, \dd y}. 
\end{align}
By construction of $\hat{\rho}_\mathcal{D}$, the right-hand side of \cref{eq:sk} admits an exact finite-sum representation. Specifically, denote by $\mathsf\Phi$ the CDF of a one-dimensional standard Gaussian distribution, and let $\phi$ be its density. For $\nu \in \cH$, $k \in \{0\} \cup [N - 1]$ and $x = (x_1, \ldots, x_d)^{\top} \in \R^d$, define
\begin{align*}
	& G_{\cD, \nu, k}(x) = \frac{1}{Z_{\cD, \nu}} \prod_{q=1}^d \frac{\Lambda_{\nu, q}(x)\exp\left( v_{\cD,\nu,q}(e^{T - t_k} x_q-c_{\nu,q}) + (e^{2T - 2t_k} - 1)v_{\cD,\nu,q}^2/2 \right)}
        {e^{-T + t_k}} \quad \mbox{and} \\
	& M_{\cD, \nu, k, q}(x) = e^{T - t_k} x_q + (e^{2T - 2t_k} - 1) v_{\cD, \nu, q} + \frac{\sqrt{e^{2T - 2t_k} - 1}}{\Lambda_{\nu, q}(x)} \left( \phi(\iota_{\nu, q}^-(x))-\phi(\iota_{\nu, q}^+(x))   \right),
\end{align*}
where for $\nu \in \cH$, $q \in [d]$ and $x \in \R^d$, 
\begin{align*}
	& \Lambda_{\nu, q}(x) = \mathsf\Phi \left( \iota_{\nu, q}^+(x)\right) - \mathsf\Phi \left( \iota_{\nu, q}^-(x) \right), \\
	& \iota_{\nu, q}^-(x) =  \frac{\ell_{\nu, q} - e^{T - t_k} x_q - (e^{2T - 2t_k} - 1) v_{\cD, \nu, q}}{\sqrt{e^{2T - 2t_k} - 1}} \quad \mbox{and} \\
	& \iota_{\nu, q}^+(x) =  \frac{u_{\nu, q} - e^{T - t_k} x_q - (e^{2T - 2t_k} - 1) v_{\cD, \nu, q}}{\sqrt{e^{2T - 2t_k} - 1}}.
\end{align*}
For $\nu \in \cH$, $k \in \{0\} \cup [N - 1]$ and $x \in \R^d$, we let $M_{\mathcal D,\nu,k}(x) =(M_{\mathcal D,\nu,k,q}(x))_{q=1}^d \in \R^d$.
Note that both $G_{D, \nu, k}$ and $M_{D, \nu, k}$ can be computed using our DP score estimator evaluated at bin centers and one-dimensional Gaussian cumulative distribution functions and densities. 
With the above definitions, one can verify that
\begin{align}
\label{eq:DM-score}
	\hat s_k(x) = -\frac{x}{1 - e^{-2T + 2t_k}} + \frac{e^{-T + t_k}}{1 - e^{-2T + 2t_k}}\,\frac{\sum_{\nu\in\cH}\hat p_{\cD,\nu}^\hist G_{\cD,\nu,k}(x) M_{\cD, \nu, k}(x)}
      {\sum_{\nu\in\cH}\hat p_{\cD,\nu}^{\hist}G_{\cD,\nu,k}(x)}.
\end{align}
This requires $O(d|\cH|)$ scalar evaluations per score evaluation. 
Note that all score estimates $(\hat s_k)_{k = 0}^{N - 1}$ are functions of $(f_{\hat\theta_{\mathrm{LMC}}},\hat\rho_\cD^\hist)$, hence the privacy cost does not increase as we increase the number of denoising steps $N$.

\subsection{DP diffusion model}
\label{subsec:dpddpm}

Substituting the private score estimates from \cref{eq:DM-score} into \cref{eq:DDPM} yields a private diffusion model, and we denote its output distribution by $\hat\rho_1$. 
In this section, we show that $\hat\rho_1$ estimates the target distribution $\rho_0$ at a nearly DP minimax-optimal rate in total variation distance.
Recall from Section \ref{sec:diffusion-models} that the diffusion model reverse process is discretized on the time grid $0=t_0<t_1<\ldots<t_N<T$, with step sizes $\Delta_k=t_{k+1}-t_k$ for $k=0,1,\ldots,N-1$. 
The process stops early at $t_N=T-\tau$, where $\tau=\bar\Delta_0>0$. 
Following \cite{benton2024nearly}, we consider a discretization scheme such that $T>1$, and for some $\kappa \in (0, 1)$, we have $\Delta_k \leq \kappa\min\{1,T-t_{k+1}\}$ for $k = 0, \ldots, N-1$.
We give an explicit construction of such a grid in \cref{lem:discretize-scheme} in the appendix.
In \cref{thm:ddpm-upper} below, we provide an upper bound on the total variation distance between $\rho_0$ and $\hat\rho_1$ under the specific discretization scheme from \cref{lem:discretize-scheme}.
In \cref{thm:private-ddpm-tv}, we prove a more general version of the theorem that holds for any discretization scheme satisfying the conditions in \cite{benton2024nearly}.

\begin{thm}
\label{thm:ddpm-upper}
Fix $\alpha>0$ and $0<\beta\leq \min (\alpha, 1)$.
Suppose that $\rho_0 \in \mathcal{P}_{\xi, L, \gamma, M}$. 
We consider the discretization scheme stated in the proof of \Cref{lem:discretize-scheme},  with the following parameter choices: 
\begin{align}
\label{eq:private-ddpm-parameters}
\begin{split}
 {\tau} = \bar b_n^{2\gamma+2}(\log n)^{2-\gamma}, \qquad T=1+\log\frac{1+\sqrt d\,(R_n+1)}{\sqrt{\tau}}
 \qquad \mbox{and} \qquad
 \kappa=\frac{\tau}{d\big(T+R_n^2+\log(1/\tau)+1\big)}.
\end{split}
\end{align}
Then the number of DDPM steps satisfies
$$N\leq \frac{C_1d}{\tau}\big(T+R_n^2+\log(1/\tau)+1\big) \big(T+\log(1/\tau)+1\big)$$
for some positive numerical constant $C_1$. Moreover, with score estimates from \cref{eq:DM-score}, the resulting DDPM is $(\alpha, \beta)$-DP, and for $n$ large enough we have 
\begin{align}
\label{eq:private-ddpm-rate}
 \E\left[ \TV(\rho_0,\hat\rho_1) \right] \leq C_{L,\xi,d,\gamma,M}
       \max \big(n^{-(\gamma+1) / (d+2\gamma+2)},
       \,(n\alpha)^{-(\gamma+1) / (d+\gamma+1)}\big) (\log n)^{3 / 2},  
\end{align}
where $C_{L,\xi,d,\gamma,M}$ is a positive constant that depends only on $(L,\xi,d,\gamma,M)$. 
\end{thm}

\begin{proof}
    We prove \Cref{thm:ddpm-upper} in Appendix \ref{proof:thm:ddpm-upper}. 
\end{proof}

The proof of \Cref{thm:ddpm-upper} is based on separately controlling the error in the score estimates and the error from early stopping, initialization and time discretization in DDPM implementation. The choice of~$\tau$ controls the early stopping error to be of order $\sqrt{\tau}$, and we choose $\tau$ to match the minimax rate. 
The parameters $T$ and $\kappa$ are chosen so that the initialization and discretization errors of the reverse process are also bounded by that rate. 

Finally, we establish a minimax lower bound for distribution estimation under total variation distance. 

\begin{thm}
\label{thm:tv-holder-lower}
Fix $\alpha>0$ and $0<\beta\leq \min (\alpha, 1)$.
Assume $L\xi^2>2$ and $M\geq\sqrt d\,\xi$.
Then for all sufficiently large $n$,
\begin{equation}
\label{eq:tv-holder-minimax-lower}
\inf_{\widehat\rho\in\mathcal M_{\alpha,\beta}}
\sup_{\rho\in\cP_{\xi,L,\gamma,M}}
\E_{\rho^{\otimes n}}\left[\TV(\widehat\rho,\rho)\right]
\geq C_{L,\xi,d,\gamma}
\max\big(
n^{-(\gamma+1) / (d+2\gamma+2)},
(n\alpha)^{-(\gamma+1) / (d+\gamma+1)}
\big),
\end{equation}
where the infimum is over all $(\alpha,\beta)$-DP distribution estimators based on $n$ i.i.d.~observations from $\rho$, and
the constant $C_{L,\xi,d,\gamma}>0$ depends only on $(L,\xi,d,\gamma)$ and is independent of $M$.
\end{thm}

\begin{proof}[Proof of \cref{thm:tv-holder-lower}]
	We prove \cref{thm:tv-holder-lower} in Appendix \ref{proof:thm:tv-holder-lower}. 
\end{proof}

Theorems~\ref{thm:ddpm-upper} and \ref{thm:tv-holder-lower} together establish that the proposed DDPM is minimax optimal up to poly-logarithmic factors for private distribution estimation in total variation distance, and DDPM also provides an explicit algorithm for sampling from this distribution.  
This rate is determined by the larger of the non-private term $n^{-(\gamma+1) / (d+2\gamma+2)}$ and the privacy term $(n\alpha)^{-(\gamma+1) / (d+\gamma+1)}$.  
Enforcing privacy therefore leaves the minimax rate unchanged up to polylogarithmic factors when $\alpha = \tilde \Omega( n^{-(\gamma+1)/(d + 2 \gamma + 2)})$.  
On the other hand, the privacy term dominates when $\alpha = \tilde O( n^{-(\gamma+1)/(d + 2 \gamma + 2)})$. 
This dependence on the sample size, smoothness, and privacy budget parallels the results of \cite{lalanne2024privately} for private estimation of $(\gamma+1)$-Sobolev smooth densities supported on a hypercube.  
Specifically, up to logarithmic factors, our DP minimax rate in total variation distance has the same exponent of $n$ as the square root of their squared-$L_2$ minimax rate for private density estimation, after converting their zero-concentrated DP guarantee to $(\alpha, \beta)$-DP \cite[Proposition 1.3 in][]{bun2016concentrated}.  
Although the loss functions and distribution classes differ, this comparison connects our diffusion model sampling guarantees to existing minimax results for private nonparametric density estimation.

\section{Numerical experiments} 
\label{sec:experiments}
In this section, we present numerical experiments that support our theoretical results, using both synthetic data with known densities and real data. 

\subsection{Experiments with synthetic data} \label{subsec:1d-synthetic}

In our first experiment, we consider a Gaussian density in $\R$ with a Weierstrass perturbation of its log-density:  
\begin{align}
\label{eq:Weierstrass}
\rho_0(x)\propto \exp\left(-\frac{x^2}{2} -\frac14\sum_{k=0}^{\infty}2^{-k(1+\gamma)}\cos(2^kx)\right). 
\end{align}
For $0<\gamma<1$, the score of distribution \eqref{eq:Weierstrass} is locally $\gamma$-H\"older continuous, and this regularity is sharp: on any interval of positive length, it fails to be $\gamma'$-H\"older continuous for every $\gamma'\in(\gamma,1]$. In particular, for all $x, y \in \R$ with $|x - y| \leq 1$, 
\begin{align*}
	& |\nabla \log \rho_0(x) - \nabla \log \rho_0(y)| \leq L_{\gamma} |x - y|^{\gamma}, \\
	& L_{\gamma} =1 + \frac{1}{4}\left(\frac{2^{1-\gamma}}{2^{1-\gamma}-1}+\frac{2}{1-2^{-\gamma}}\right),\qquad 0<\gamma<1.
\end{align*}
For $\gamma = 1$, we use the following distribution with a truncated perturbation to ensure that its score function has a finite Lipschitz constant: 
\begin{align*}
	\rho_0(x)\propto \exp\left(-\frac{x^2}{2} -\frac14\sum_{k=0}^{40}2^{-2k}\cos(2^kx)\right). 
\end{align*}
For the privacy parameters, we consider $\alpha \in \{0.5, 1, 2, 4, 8\}$ and set $\beta = 0.01$. 
For the smoothness parameter~$\gamma$, when varying smoothness we take $\gamma\in\{0.2,0.4,0.6,0.8,1\}$, and when varying the privacy budget we fix $\gamma=0.5$. 
We vary the training sample size over $n \in \{1000, 2000, 4000, 8000, 16000\}$.
For each setting, we use 500,000 projected LMC iterations, and set $\delta = 0.2 / \lambda_{\max}(\Sigma_{\cD})$. 
As noted in Remark~\ref{rem:LMC}, we provide convergence diagnostics for the chosen number of LMC iterations in Appendix \ref{sec:diagnostics}.
Additional numerical experiments with a two-dimensional target distribution are presented in Appendix \ref{sec:two-dim-synthetic}.

\begin{figure}[hbt!]
	\centering
	\includegraphics[width=\textwidth]{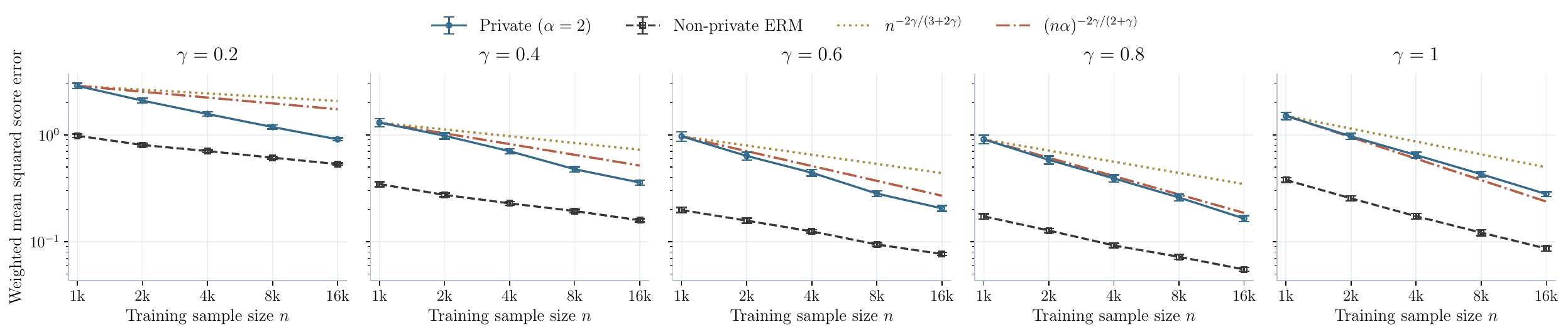}
	\caption{Clean score estimation error for distribution \eqref{eq:Weierstrass}. 
	Each panel fixes a smoothness level $\gamma$ and plots the mean density-weighted squared score error against the training sample size $n$ on logarithmic scales. 
	Blue curves average the risks over 100 independent experiments. 
	Error bars give 95\% confidence intervals. 
	The dotted and dash-dot lines represent the rates $n^{-2\gamma/(3+2\gamma)}$ and $(n\alpha)^{-2\gamma/(2+\gamma)}$, respectively. Their log-log intercepts are chosen separately so that both lines pass through the private error point at $n = 1000$ in each panel. These lines illustrate slopes for decay rates rather than actual lower bounds.
	}
	\label{fig:clean-score-risk1}
\end{figure}

\paragraph{Clean score estimation.}
We first consider clean score estimation. 
For each setting, we repeat the experiment independently 100 times and evaluate the density-weighted mean squared score error via Monte Carlo integration using 10,000 independent samples from the target density. 
We report the average error across these repetitions, with error bars indicating pointwise 95\% confidence intervals for the mean. 
Figure~\ref{fig:clean-score-risk1} plots the mean error against the training sample size on logarithmic scales for $\alpha=2$, with each panel corresponding to a different smoothness level $\gamma$. 
Figure \ref{fig:clean-score-risk1} also includes the non-private ERM error for comparison. 
We compare the observed decay with the theoretical reference rates $n^{-2\gamma/(3+2\gamma)}$ and $(n\alpha)^{-2\gamma/(2+\gamma)}$. 
The reference rate $(n\alpha)^{-2\gamma/(2+\gamma)}$ approximately matches the decay of our algorithm's risk for $\gamma \in \{0.8,1\}$. 
For $\gamma \in \{0.2,0.4,0.6\}$, our algorithm's risk decreases faster than the reference rates. This is consistent with the theory: our minimax rate $\mbox{polylog}(n)\max \{n^{-2\gamma/(3+2\gamma)}, \,(n\alpha)^{-2\gamma/(2+\gamma)}\}$ represents worst-case risk over a class of distributions and does not preclude faster convergence for particular target distributions. 

\begin{figure}[hbt!]
	\centering
	\includegraphics[width=\textwidth]{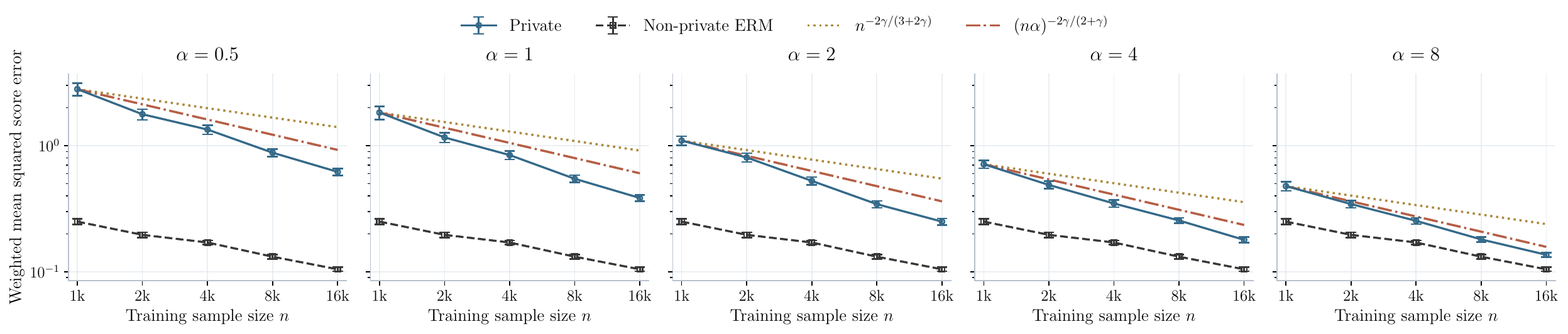}
	\caption{Clean-score estimation risk for distribution \eqref{eq:Weierstrass}. 
	We set $\gamma = 0.5$, and each panel fixes a privacy level $\alpha$ and plots the mean population-weighted squared score error against the training sample size $n$ on logarithmic scales. All other plotting conventions are as in Figure \ref{fig:clean-score-risk1}. 
	}
	\label{fig:clean-score-risk2}
\end{figure}

Figure \ref{fig:clean-score-risk2} plots the mean error against the training sample size on logarithmic scales for $\gamma=0.5$, with each panel corresponding to a different privacy level $\alpha$. 
The non-private ERM error is included for comparison. 
Over the sample sizes considered, the observed risk of our private estimator decays faster than the minimax reference rate. 
As $\alpha$ increases, the private risk approaches the non-private risk, which is consistent with the theoretical prediction.

\begin{figure}[hbt!]
\centering
\begin{subfigure}[t]{0.48\textwidth}
\centering
\includegraphics[width=\textwidth]{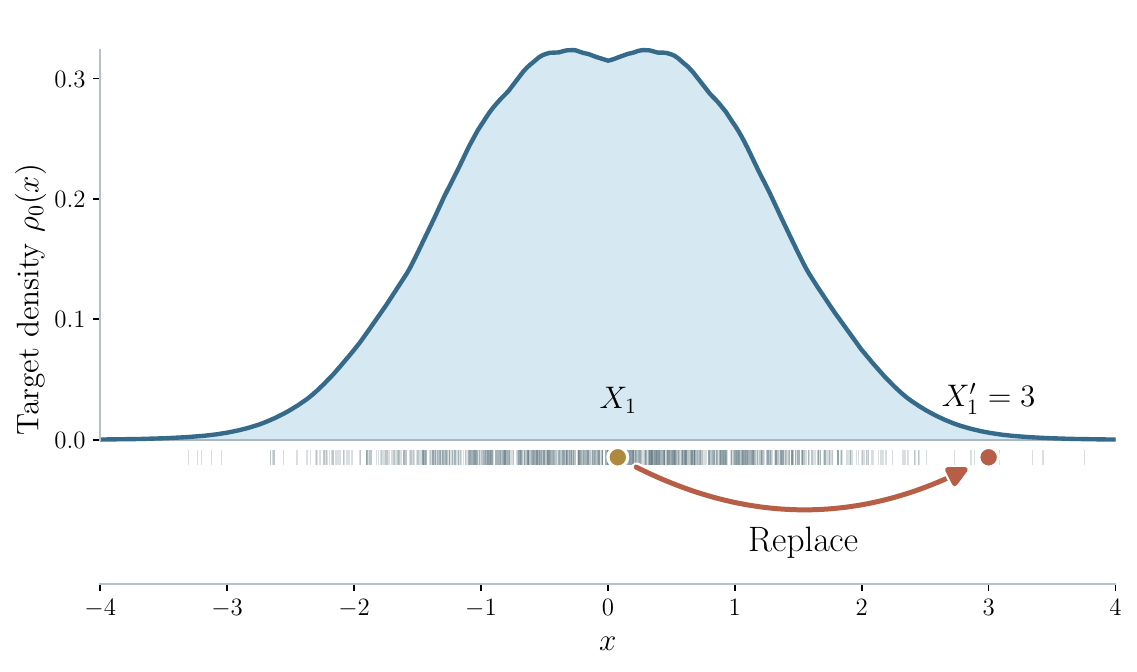}
\caption{The target distribution and sampled observations for $\gamma = 0.5$, illustrating the replacement of a single observation to $3$.}
\label{fig:perturb}
\end{subfigure}
\hfill
\begin{subfigure}[t]{0.48\textwidth}
\centering
	\includegraphics[width=\textwidth]{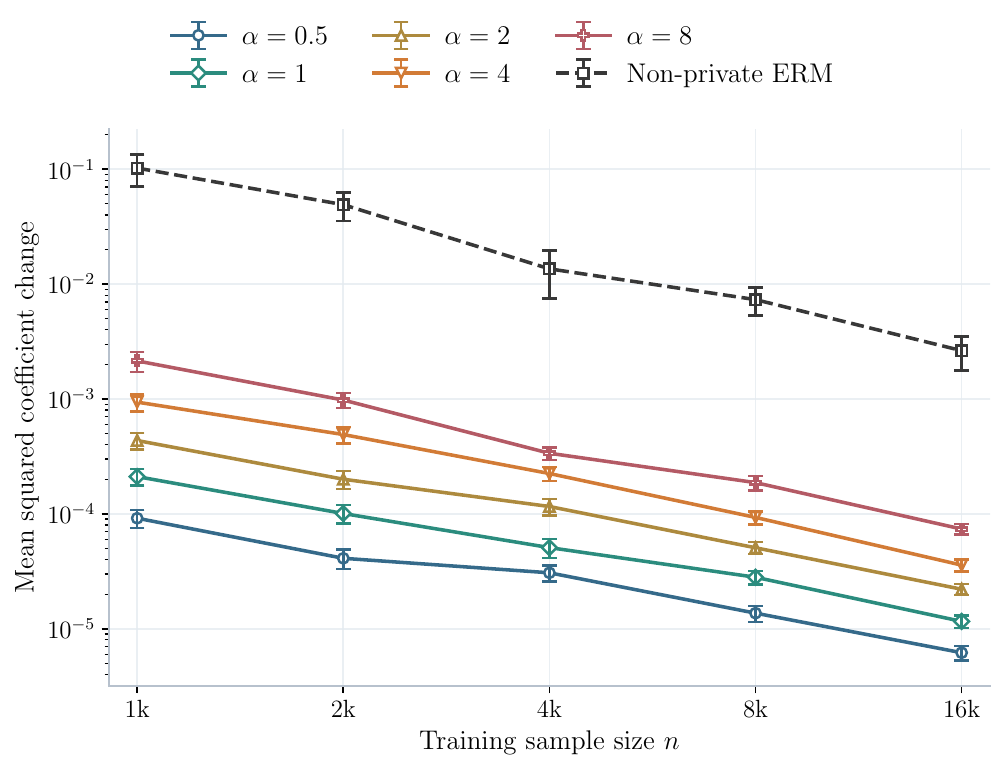}
	\caption{
	Coefficient changes induced by replacing one observation with $3$. 
	Colored solid curves show the normalized squared coefficient change $\|\theta_{\mathcal D'}-\theta_{\mathcal D}\|_2^2/|\mathcal J|$ for different $\alpha$, averaged over 100 independent repetitions. Black dashed curves shows the corresponding values for the non-private ERM estimator.
	}
	\label{fig:coefficient-change}
\end{subfigure}
\caption{Illustration of single observation perturbations.}
\end{figure}

We next assess privacy by examining the empirical stability of our DP score estimator under a single observation perturbation. 
Specifically, for each repetition, we construct two neighboring datasets $\cD$ and $\cD'$, where $\cD'$ is obtained by replacing the first training observation in $\cD$ with a fixed value $3$. 
The remaining $n-1$ shared observations are the same. 
We refer to Figure \ref{fig:perturb} for an illustration of such replacement. 

Let $\theta_{\mathcal D}$ and $\theta_{\mathcal D'}$ denote the spline coefficient estimates produced from $\mathcal D$ and $\mathcal D'$, respectively. 
We quantify the coefficient change by the normalized squared Euclidean distance $\|\theta_{\mathcal D'}-\theta_{\mathcal D}\|_2^2/|\mathcal J|$. 
We fix $\gamma=0.5$ and conduct 100 independent repetitions for each sample size $n$ and privacy level $\alpha$, generating a fresh dataset in each repetition. 
Within each pair of projected LMC chains, we use the same sequence of standard Gaussian vectors at corresponding iterations. 
We compare the resulting coefficient changes of our DP score estimator with those produced by the non-private ERM. 
Figure \ref{fig:coefficient-change} reports the mean normalized squared coefficient changes and pointwise 95\% confidence intervals for both estimators. 
The results show that our DP estimator exhibits substantially smaller coefficient changes under the single observation replacements considered here. 

\begin{figure}[hbt!]
	\centering

\end{figure}

\paragraph{Diffusion model application.}
We next utilize the learned DP score to build a diffusion model. 
Specifically, we combine each fitted clean score estimate with a private histogram, and we choose the histogram setting so that it is $(\alpha, 0)$-DP. 
Therefore, if the score estimator is $(\alpha,\beta)$-DP, the entire procedure is $(2\alpha,\beta)$-DP.
As before, we fix $\beta = 0.01$ and consider $\alpha \in \{0.5,1,2,4,8\}$. 
For each setting, we generate $20,000$ samples from the corresponding diffusion model and $50,000$ samples from the target distribution. 
We estimate the total variation (TV) error from the generated samples using $70$ equal-width bins on $[-8, 8]$ with additional bins for the two tails. 
As before, we independently repeat the experiments 100 times.

\begin{figure}[hbt!]
	\centering
	\includegraphics[width=\textwidth]{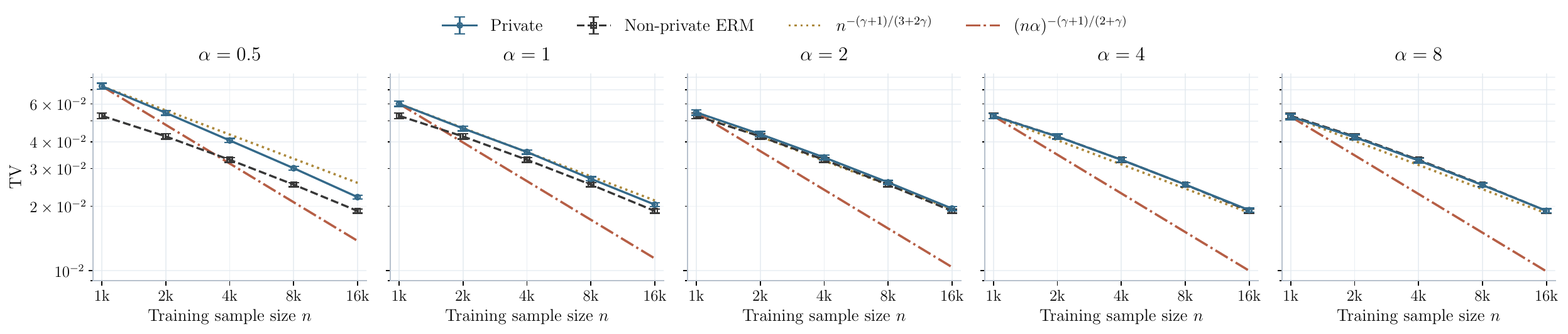}
	\caption{Mean TV error of the private diffusion model for $\gamma=0.5$, with each panel fixing a privacy level $\alpha$. Error bars indicate pointwise 95\% confidence intervals. Dotted and dash-dot lines show the reference rates, with their intercepts chosen separately to match the first private error point in each panel.}
	\label{fig:synthetic_DM1}
\end{figure}

Figure \ref{fig:synthetic_DM1} plots the mean TV error of the private diffusion model against the training sample size $n$ for $\gamma=0.5$, with each panel corresponding to a different privacy level $\alpha$. Error bars indicate pointwise 95\% confidence intervals for the mean. The dotted and dash-dot lines show the reference rates $n^{-3/8}$ and $(n\alpha)^{-3/5}$, respectively.
For comparison, we also report results for a diffusion model constructed using the non-private ERM score estimator and non-private histogram.
From the plots, we see that the TV error decreases steadily as the sample size increases. 
Over the sample sizes considered, the observed decay for our DP diffusion model more closely follows the rate $n^{-3/8}$ than $(n\alpha)^{-3/5}$. 
As $\alpha$ increases, the performance gap between the private and non-private diffusion models becomes smaller.

\begin{figure}[hbt!]
	\centering
	\includegraphics[width=\textwidth]{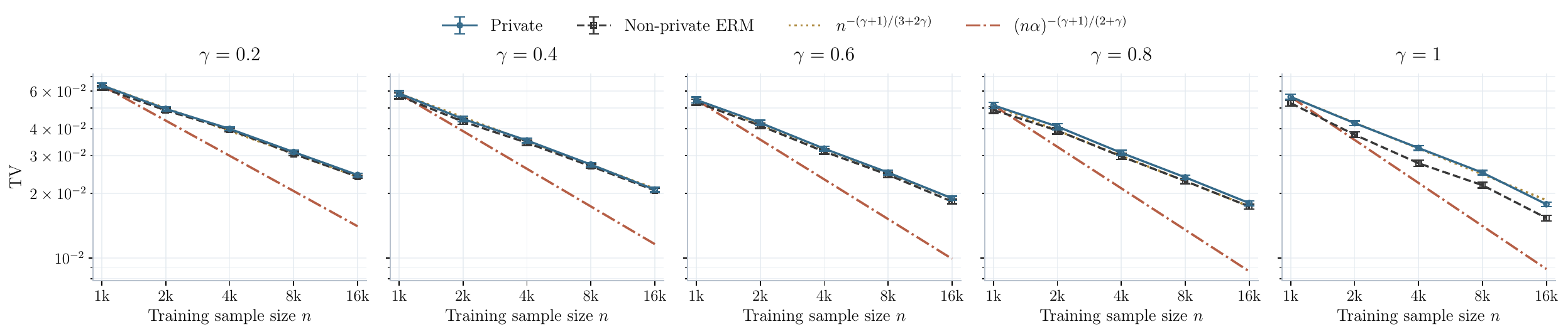}
	\caption{Mean TV error of the private diffusion model for $\alpha=2$, with each panel fixing a smoothness level $\gamma$. Error bars indicate pointwise 95\% confidence intervals. Dotted and dash-dot lines show the reference minimax rates, with their intercepts chosen separately to match the first private error point in each panel.}
	\label{fig:synthetic_DM2}
\end{figure}

In Figure \ref{fig:synthetic_DM2}, we fix $\alpha=2$ and examine the TV error across different smoothness levels $\gamma\in\{0.2,0.4,0.6,0.8,1\}$. Each panel reports the mean error over 100 independent repetitions with pointwise 95\% confidence intervals, together with the non-private benchmark. Both private and non-private errors decrease steadily as the training sample size increases. Across the panels, the observed decay of our algorithm aligns more closely with the non-private reference rate $n^{-(\gamma+1)/(3+2\gamma)}$ than with the steeper privacy-dependent rate $(n\alpha)^{-(\gamma+1)/(2+\gamma)}$. These results suggest a modest utility cost of privacy at the chosen privacy level.

\begin{figure}[hbt!]
\centering
	\includegraphics[width=0.5\textwidth]{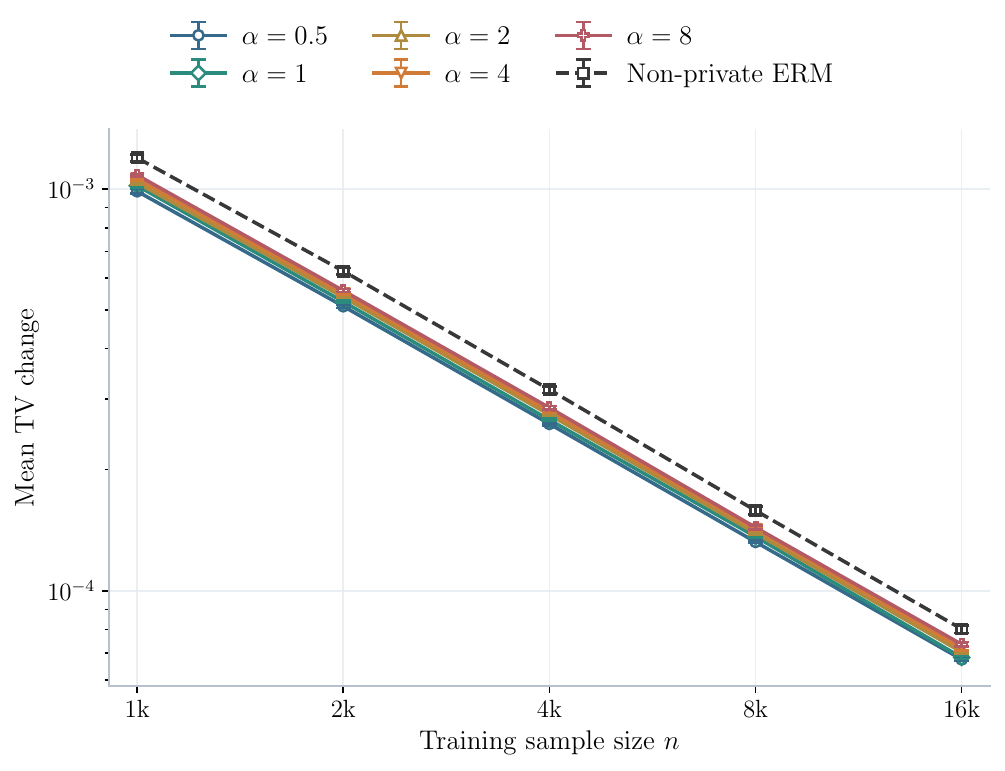}
	\caption{
	TV changes of both private and non-private DDPMs induced by replacing one observation with $3$. 
	}
	\label{fig:tv-change}
\end{figure}

We next examine how replacing the first observation in the dataset with $3$ affects the output distributions of private and non-private diffusion models. We quantify these changes using total variation distance. Figure \ref{fig:tv-change} reports the mean changes and pointwise 95\% confidence intervals for the private and non-private DDPMs at different privacy levels.
A comparison with Figure \ref{fig:coefficient-change} suggests that ensuring privacy for DDPM outputs may be less costly than ensuring privacy for clean-score estimates. We leave a systematic investigation of this possibility to future work. 

\subsection{Real data experiment on NHANES dataset}

We next evaluate our proposed methods on a real dataset. 
We use the NHANES 2017 -- March 2020 pre-pandemic dataset \citep{akinbami2022national}, a U.S.~survey that collects health and nutritional information through interviews and physical examinations. 
We analyze the joint distribution of body mass index (BMI) and waist circumference. Our analysis includes 7,486 nonpregnant adults aged 20-79 with both measurements observed. Before fitting, we apply the public log transformations
\[
  x_1=\log(\mathrm{BMI}/25)/0.35,
  \qquad
  x_2=\log(\mathrm{waist}/90)/0.30.
\]
In this experiment, we set $d = 2$, $\gamma = 0.5$ and $L = 4$. 
We conduct 100 repetitions for our experiment. 
In each repetition, we draw a fresh 75/25 split, yielding 5,614 training samples and 1,872 held-out samples. 
For different privacy levels we use the same split within each repetition. 
As before, we set $\beta = 0.01$ and $\alpha \in \{0.5, 1, 2, 4, 8\}$.   
Each private score fit uses 500,000 projected LMC iterations, and we generate 10,000 samples from each of the resulting diffusion models to assess the sampling error. 
We conduct experiments for both clean score estimation and diffusion model with corrected scores.

\begin{figure}[t]
  \centering
  \includegraphics[width=0.8\textwidth]{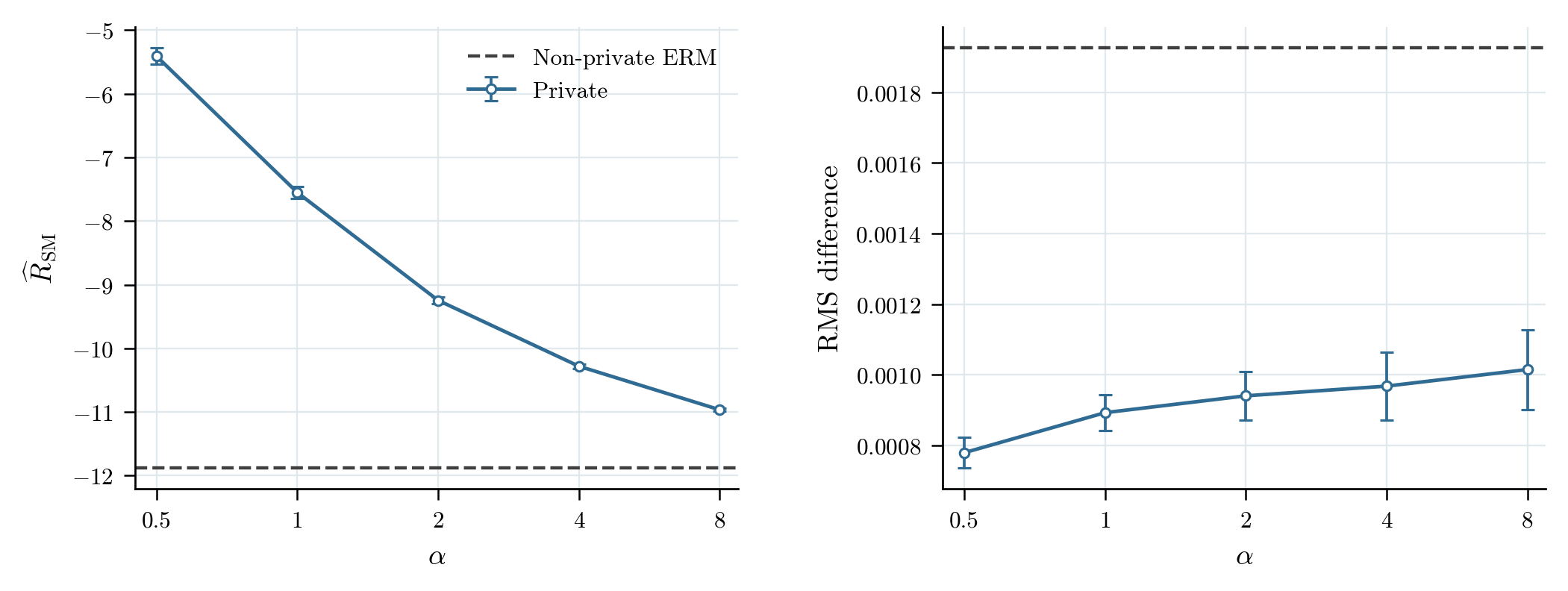}
  \caption{Clean score estimation results for the NHANES real-data experiment. The left panel reports the held-out empirical
  score-matching risk for the private score estimator and the non-private
  ERM benchmark. 
  The right panel reports the RMS difference when we perturb one training sample.
  Markers denote
  means and error bars are pointwise 95\% confidence intervals.
  }
  \label{fig:nhanes-score}
\end{figure}

\paragraph{Clean score estimation.} 
We first evaluate the performance of the clean score estimators using the empirical score matching risk on the held-out data, given by
\[
  \widehat R_{\mathrm{SM}}(s)
  =\frac{1}{n_{\mathrm{test}}}\sum_{i=1}^{n_{\mathrm{test}}}
  \left(\frac{1}{2}\lVert s(X_i)\rVert_2^2
  +\diver s(X_i)\right).
\]
By \cref{lemma:M-R}, the population score matching risk is an increasing affine function of the density-weighted squared risk, hence lower values of this risk suggest a more accurate score estimate.
The left panel of Figure~\ref{fig:nhanes-score} shows that the empirical risk on the held-out data decreases as the privacy budget increases, and approaches the non-private ERM risk.

We next consider a perturbation analysis that changes a single observation, as in Section \ref{subsec:1d-synthetic}. 
Specifically, in each repetition, we replace one randomly selected training sample by a randomly selected sample from the corresponding held-out dataset.  
The right panel of Figure~\ref{fig:nhanes-score} reports the root-mean-square (RMS) difference between the spline coefficients fitted to the original and perturbed training sets. 
From the figure, we see that the mean score perturbation averaged across repetitions increases as $\alpha$ increases, and is consistently lower than the non-private ERM baselines.

\begin{figure}[t]
  \centering
  \includegraphics[width=0.8\textwidth]{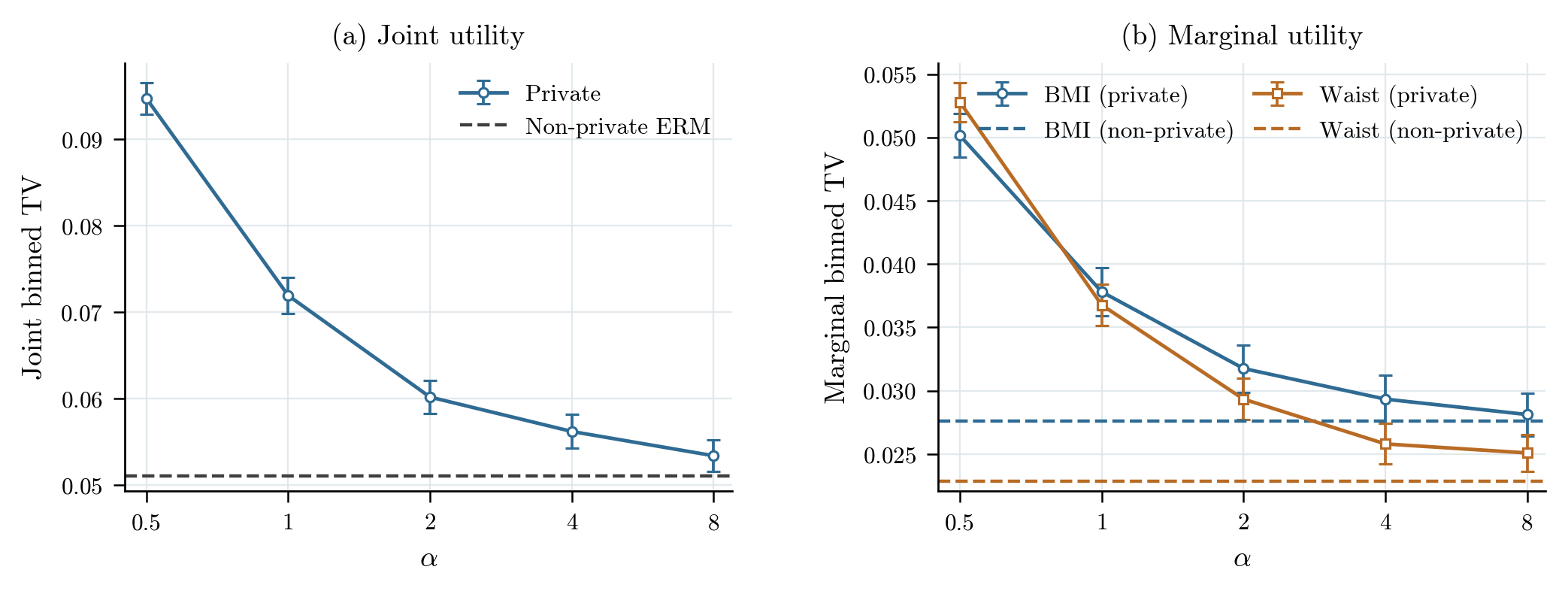}
  \caption{Binned TV distances between the generated and target distributions, estimated using held-out datasets.
  The left panel reports the joint binned TV distance, and the right panel reports the marginal binned TV distances. 
  Solid curves denote means for DP diffusion models across 100 repetitions with pointwise 95\% confidence intervals, and dashed horizontal lines denote non-private ERM means.
  }
  \label{fig:nhanes-real}
\end{figure}

\paragraph{Diffusion model application.} 
We next use our algorithm to construct a DP diffusion model. 
We assess the quality of the output distributions by the binned TV distances between generated and held-out samples. 
We use an $18 \times 18$ grid in the transformed BMI and waist circumference coordinates. 
Along each coordinate, we use 16 equal-width bins on $[-4,4]$ and one bin for each tail.
Figure~\ref{fig:nhanes-real} shows that as the privacy budget increases, the joint TV distance and the two marginal TV distances (for waist circumference and BMI) all decrease toward their non-private counterparts.

\begin{figure}[t]
  \centering
  \includegraphics[width=0.4\textwidth]{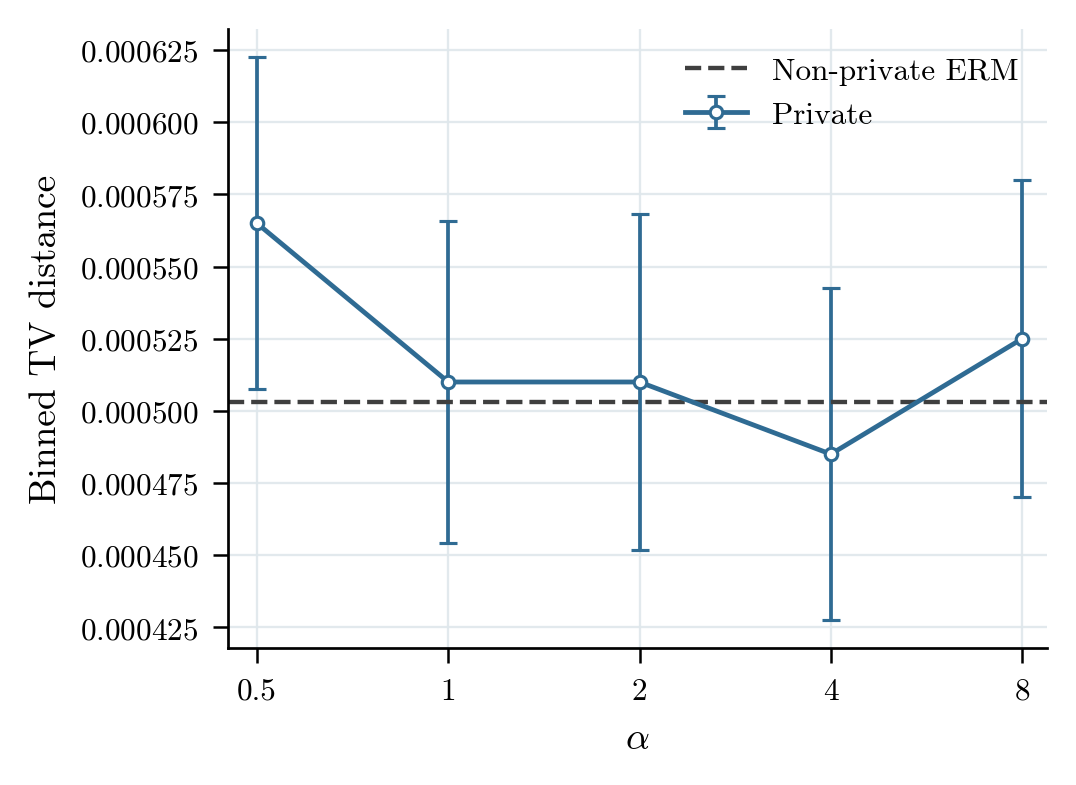}
  \caption{Binned TV distances between the output distributions of diffusion models trained on the original and perturbed datasets.  
  Markers denote means over 100 repetitions and error bars denote pointwise 95\%
  confidence intervals. 
  The dashed curve gives the non-private
  ERM benchmark.} 
  \label{fig:nhanes-perturbation}
\end{figure}

Finally, we assess how replacing a single training sample affects the generated distribution. 
\Cref{fig:nhanes-perturbation} reports the binned TV distance between the output distributions of diffusion models trained on the original and perturbed datasets.
We observe no clear trend as the privacy budget $\alpha$ increases.
One possible explanation is that the effect of replacing one observation may be small relative to the Monte Carlo variability of the binned TV estimates, making a trend difficult to detect.
Binned TV distance may also be too coarse to capture differences in sensitivity that are apparent at the coefficient level. 
In particular, the right panel of Figure \ref{fig:nhanes-score} shows that private and non-private methods exhibit clear differences in how their fitted clean-score coefficients change under the same perturbation.


\section{Discussion}
\label{sec:discussion}

In this paper, we establish minimax rates up to polylogarithmic factors, for DP score estimation and for sampling in total variation distance using a private diffusion model.
These rates hold under mild smoothness and tail conditions.  As in the non-private setting, score estimation suffers from the curse of dimensionality in the absence of additional structural assumptions \citep{wibisono2024optimal,liu2024kernel,oko2023diffusion,yu2026diffusion}.  
One possible explanation for the empirical success of score-based generative models on high-dimensional data such as images is that the data may have an intrinsic dimension $k$ much smaller than the ambient dimension $d$ \citep{pope2021the}.

A natural direction for future work is to develop private score estimators that adapt to the low-dimensional structure of the data, such as when the data are gathered around an unknown manifold of intrinsic dimension $k$ that is potentially much smaller than the ambient dimension $d$.  
Such adaptive settings raise the challenge of quantifying the additional error incurred when learning the manifold geometry under privacy constraints.  
Establishing matching upper and lower bounds when $k < d$ would clarify when a low intrinsic dimension improves the rates for private score estimation and whether these gains carry over to private diffusion models.

\section*{Acknowledgements}
Hung is supported by the Chancellors' Scholarship scheme and the Statistics Centre for Doctoral Training at the University of Warwick.
Yu is partially supported by the EPSRC programme grant DASS (EP/Z531327/1), UKRI-JST grant Statistical Safeguarding and Leverhulme Trust (Philip Leverhulme Prize).


\bibliographystyle{plainnat}
\bibliography{bib}

\appendix 

\section{Technical lemmas}

We collect in this section technical lemmas used for our analysis. 

\begin{lem}
\label{lem:subgaussian-concentration}
Suppose a $d$-dimensional random vector $X$ is $\xi$-sub-Gaussian, then for every $u\in\R^d$ with $\|u\|_2=1$ and every $t \geq 0$,  
\begin{align*}
    \P\big( |\langle X, u \rangle - \E[\langle X, u \rangle]| \geq t  \big) \leq 2 \exp \Big( -\frac{t^2}{2 \xi^2} \Big). 
\end{align*}
In addition, there exist numerical constants $c, C > 0$, such that for all $t \geq 0$
\begin{align*}
    \mathbb{P}\!\left(\|X - \mathbb{E}[X]\|_2 \ge C\xi(\sqrt{d} + t)\right) \le e^{-c t^2}.
\end{align*}
\end{lem}

\begin{lem}
\label{lemma:cP-non-empty}
Let $0<\gamma\leq 1$. If $\xi\geq L^{-1/2}$, then the distribution class $\mathcal P_{\xi,L,\gamma,M}$ is nonempty for every $M>0$.
On the other hand, if $\xi < L^{-1/2}$, then $\mathcal P_{\xi,L,\gamma,M}$ is empty for every $M>0$.

\end{lem}

\begin{proof}[Proof of \cref{lemma:cP-non-empty}]
Let $\rho_g$ be the density of $\mathsf N(0,L^{-1}I_d)$. Then
$\nabla\log\rho_g(x)=-Lx$. For $\|x-y\|_2\le1$,
\begin{align*}
\|\nabla\log\rho_g(x)-\nabla\log\rho_g(y)\|_2 = L\|x-y\|_2 \le L\|x-y\|_2^\gamma.
\end{align*}
Moreover, $\rho_g$ is $L^{-1/2}$-sub-Gaussian, and hence is
$\xi$-sub-Gaussian whenever $\xi\ge L^{-1/2}$. This proves $\mathcal P_{\xi,L,\gamma,M}$ is nonempty for every $M>0$.

We next prove $\mathcal P_{\xi,L,\gamma,M}$ is empty when $\xi < L^{-1/2}$.
Suppose there exists a distribution with density $\rho$ that belongs to $\mathcal P_{\xi,L,\gamma,M}$. 
For $x \in \R^d$, we divide the segment from $0$ to $x$ into unit-length pieces and a remaining piece of length less than $1$. By the local H\"older condition, we have $\|\nabla\log\rho(x)-\nabla\log\rho(0)\|_2\leq L(\|x\|_2+1)$. 
Consequently, 
\begin{align*}
\log\rho(x)-\log\rho(0) =\int_0^1\langle \nabla\log\rho(tx),x\rangle\,\dd t \geq-\frac{L}{2}\|x\|_2^2-\big( \|\nabla\log\rho(0)\|_2+L\big)\|x\|_2.
\end{align*}
For every $\varepsilon>0$, since $(\|\nabla\log\rho(0)\|_2+L)\|x\|_2 \leq \varepsilon \|x\|_2^2/2+(\|\nabla\log\rho(0)\|_2+L)^2/(2\varepsilon)$, we have 
\begin{align*}
\rho(x)\geq \rho(0)
\exp\!\left(-\frac{L+\varepsilon}{2}\|x\|_2^2 -\frac{(\|\nabla\log\rho(0)\|_2+L)^2}{2\varepsilon}\right).
\end{align*}
Thus, for any unit vector $u$ and $t>0$, 
\begin{align}
\label{eq:exp-lower}
\E_{\rho}\left[ \exp(t\langle u,X\rangle) \right]
\geq \rho(0)
\left(\frac{2\pi}{L+\varepsilon}\right)^{d/2}
\exp\!\left(\frac{t^2}{2(L+\varepsilon)} -\frac{(\|\nabla\log\rho(0)\|_2+L)^2}{2\varepsilon}\right).
\end{align}
On the other hand, \(\xi\)-sub-Gaussianity implies
\begin{align}
\label{eq:exp-upper}
\E_{\rho}\left[ \exp(t\langle u,X\rangle) \right] \leq \exp\!\left(t\langle u,\E_{\rho} [X]\rangle+\frac{\xi^2t^2}{2}\right).
\end{align}
Taking logarithms of the right hand sides of \cref{eq:exp-lower,eq:exp-upper}, dividing them by $t^2/2$, and letting $t\to\infty$ yields $\xi^2\geq(L+\varepsilon)^{-1}$. Sending $\varepsilon\downarrow0$ proves $\xi\geq L^{-1/2}$. 
Therefore, for $\mathcal P_{\xi,L,\gamma,M}$ to be nonempty, we must have $\xi\geq L^{-1/2}$.
\end{proof}

The next lemma provides a stability bound for conditional expectations in terms of the Hellinger distance, which we recall the definition of below.

\begin{defn}[Hellinger distance]

For distributions $P$ and $Q$ with density functions $p$ and $q$, we define their squared Hellinger distance as
\begin{align*}
 \mathsf H^2(P,Q)=\int\left(\sqrt{p(x)}-\sqrt{q(x)}\right)^2 \dd x = 2-2\int\sqrt{p(x)q(x)}\,\dd x. 
\end{align*}
Note that $0\leq\mathsf H^2(P,Q)\leq2$. 
\end{defn}

\begin{lem}
\label{lem:posterior-hellinger-stability}
Let $P$ and $Q$ be two joint probability measures on $(Y,Z)$ with probability density functions $p$ and $q$.  
Then for every vector-valued $f$ satisfying $\sup_y\|f(y)\|_2\leq B$,
\begin{align}
\label{eq:posterior-mean-hellinger}
 \int\left\|\E_{P}[f(Y)\mid z]-\E_{Q}[f(Y)\mid z]\right\|_2^2\,P(\dd z) \leq 40B^2 \mathsf{H}^2(P, Q).
\end{align}
In particular, if $p(y, z)=p_0( y)K(z\mid y)$ and $q(y, z)=q_0( y)K(z\mid y)$ for a transition density $K$ and probability density functions $p_0$ and $q_0$, 
then the right-hand side equals $40B^2\mathsf H^2(P_0,Q_0)$, where $P_0$ has probability density function $p_0$ and $Q_0$ has probability density function $q_0$.
\end{lem}

\begin{proof}[Proof of \cref{lem:posterior-hellinger-stability}]
Write $p_Z$ and $q_Z$ for densities of the two $z$-marginals $P_Z$ and $Q_Z$, and define $H_z^2=\mathsf H^2( P(\cdot\mid z),Q(\cdot\mid z))$.  The conditional Hellinger identity gives
\begin{align}
\label{eq:conditional-hellinger-identity}
 \mathsf H^2(P, Q) =\mathsf H^2(P_Z, Q_Z) +\int\sqrt{p_Z(z)q_Z(z)}\,H_z^2\,\dd z.
\end{align}
On the event $\{z: p_Z(z)\leq4q_Z(z)\}$, we have $p_Z(z)\leq2\sqrt{p_Z(z)q_Z(z)}$, while on $\{z: p_Z(z)>4q_Z(z)\}$, we have $H_z^2\leq2$ and
\begin{align*}
 \int_{\{z: p_Z(z)>4q_Z(z)\}}p_Z(z)\,\dd z
	 \leq 4\mathsf H^2(P_Z,Q_Z).
\end{align*}
Therefore, utilizing \cref{eq:conditional-hellinger-identity}
\begin{align*}
	\int p_Z(z) H_z^2 \dd z= &\int_{\{z: p_Z(z)\leq4q_Z(z)\}} p_Z(z) H_z^2 \dd z + \int_{\{z: p_Z(z)> 4q_Z(z)\}} p_Z(z) H_z^2 \dd z \\
	\leq &\, 2\int_{\{z: p_Z(z)\leq4q_Z(z)\}} \sqrt{p_Z(z)q_Z(z)} H_z^2 \dd z + 8\mathsf H^2(P_Z, Q_Z) \\
	\leq & \, 10\,  \mathsf H^2( P, Q).
\end{align*}
For each $z$, total variation is at most Hellinger distance, and hence
\begin{align*}
 \left\|\E_{P}[f(Y)\mid z]-\E_{Q}[f(Y)\mid z]\right\|_2^2
 \leq4B^2H_z^2.
\end{align*}
Taking the integral proves \cref{eq:posterior-mean-hellinger}.  
When the two probability distributions share the same kernel, direct calculation of Hellinger distance gives
$\mathsf H^2( P, Q)=\mathsf H^2(P_0,Q_0)$.
\end{proof}

\begin{lem}[Construction of $\chi_{R_n}$]
\label{lemma:chiR}
	The function that satisfies requirements \eqref{eq:chiR} exists. 
\end{lem}

\begin{proof}[Proof of \cref{lemma:chiR}]

Define the smooth mollifier
\begin{align*}
\varphi(z)=
\begin{cases}
c_d\exp\!\left(-\dfrac{1}{1-\|z\|_2^2}\right),
    & \|z\|_2<1,\\
0,  & \|z\|_2\geq1,
\end{cases}
\end{align*}
where $c_d>0$ is chosen so that $\int_{\mathbb{R}^d}\varphi(z)\,\mathrm{d}z=1$.
For $\varepsilon>0$, write
$\varphi_\varepsilon(y)=\varepsilon^{-d}\varphi(y/\varepsilon)$.
Set
\begin{align*}
g_{R_n}(x) = \min\left\{1,\max\left\{0,\,\frac{7}{2} -\frac{2\|x\|_\infty}{R_n}\right\}\right\}, \qquad
\chi_{R_n}(x) = \int_{\mathbb R^d}g_{R_n}(y)\varphi_{R_n/4}(x-y)\,\dd y.
\end{align*}
Then $\chi_{R_n}\in C_c^\infty(\mathbb{R}^d)$ is smooth and has a compact support, and $0\leq\chi_{R_n}\leq1$. Moreover,
\begin{align*}
\chi_{R_n}(x)=1 \quad \text{for }x\in\Omega_{R_n},
\qquad
\chi_{R_n}(x)=0 \quad \text{for }x\notin\Omega_{2R_n},
\qquad
\sup_{x\in\mathbb{R}^d}\|\nabla\chi_{R_n}(x)\|_2 \leq \frac{2}{R_n}.
\end{align*}
Indeed, $g_{R_n}(x)=1$ whenever $\|x\|_\infty \leq 5R_n/4$ and
$g_{R_n}(x)=0$ whenever $\|x\|_\infty \geq 7R_n/4$.
Observe that $\varphi_{R_n/4}$ is supported on the Euclidean ball centered at $\vec{0}_d$ of radius $R_n/4$.
Therefore, for all $y$ in its support satisfies $\|y\|_\infty\leq R_n/4$, 
if $x\in\Omega_{R_n}$, then $\|x-y\|_\infty\leq 5R_n/4$, which gives $\chi_{R_n}(x)=1$.
If $\|x\|_\infty\geq2R_n$, then $\|x-y\|_\infty\geq7R_n/4$,
which gives $\chi_{R_n}(x)=0$.
Finally, $g_{R_n}$ is $2/R_n$-Lipschitz with respect to the Euclidean norm.
Convolution with a nonnegative kernel of unit mass preserves this bound:
\[
|\chi_{R_n}(x)-\chi_{R_n}(z)|
\leq \int_{\mathbb{R}^d}
       |g_{R_n}(x-y)-g_{R_n}(z-y)|\varphi_{R_n/4}(y)\,\mathrm{d}y
\leq \frac{2}{R_n}\|x-z\|_2.
\]
Since $\chi_{R_n}$ is smooth, this implies the stated gradient bound.
Hence one may take $C_\chi=2$, which is indeed a positive numerical constant.
	
\end{proof}


%

\begin{lem}
\label{lem:sc-local-poincare}
Let $A$ be a rectangle with side lengths $b>0$, let $c$ be its
center, and let $v\in\R^d$. Define
\begin{align*}
 q_v(y)=\frac{\exp(\langle v,y-c\rangle)}
                  {\int_A\exp(\langle v,z-c\rangle)\,\dd z},\qquad y\in A.
\end{align*}
For a positive $C^1$ probability density function $\rho$ on $A$, 
\begin{align}
\label{eq:sc-local-transfer}
\begin{split}
& \mathsf H^2(\rho,q_v)\leq\frac{b^2}{2\pi^2}\int_A\|\nabla\log \rho(y)-v\|_2^2\,\rho(y)\dd y, \\
& \TV(\rho,q_v) \leq\frac{b}{2\pi}\sqrt{\int_A\|\nabla\log \rho(y)-v\|_2^2\,\rho(y)\dd y}.
\end{split}
\end{align}
\end{lem}

\begin{proof}[Proof of \cref{lem:sc-local-poincare}]

Observe that $q_v$ is a product of one-dimensional exponential densities on bounded intervals. 
By the one-dimensional weighted Poincaré inequality \citep{ferone2012remark} and tensorization of Poincaré inequalities, we obtain
\begin{align}
\label{eq:sc-weighted-poincare}
\begin{split}
\operatorname{Var}_{q_v}(\sqrt{\rho/q_v}) = & 1 -   \left(\int_A \sqrt{\rho(y) q_v(y)} \,\dd y \right)^2 \\
 \leq & \frac{b^2}{\pi^2}\int_A\|\nabla (\sqrt{\rho(y)/q_v(y)})\|_2^2 \,q_v(y)\,\dd y \\
 = & \frac{b^2}{4\pi^2} \int_A\|\nabla\log \rho(y)-v\|_2^2\,\rho(y)\dd y. 
 \end{split}
\end{align}
Since $\mathsf H^2(\rho,q_v)=2(1-\int_A \sqrt{\rho(y) q_v(y)} \,\dd y)\leq2(1-[\int_A \sqrt{\rho(y) q_v(y)} \,\dd y]^2)$, the first inequality follows.  
For the second inequality, the Cauchy-Schwarz inequality implies 
\begin{align*}
 \TV(\rho,q_v)^2 = & \frac{1}{4}\left(\int_A \big|\sqrt {\rho(y)} - \sqrt{q_v(y)}\big| \,\big(\sqrt {\rho(y)} + \sqrt{q_v(y)}\,\big)\,\dd y\right)^2 \\
 \leq & \frac{1}{4}\left(2-2\int_A \sqrt{\rho(y) q_v(y)} \,\dd y\right)\left(2 + 2\int_A \sqrt{\rho(y) q_v(y)} \,\dd y \right) \\
 = & 1 - \left( \int_A \sqrt{\rho(y) q_v(y)} \,\dd y \right)^2.
\end{align*}
Combining the above upper bound with \cref{eq:sc-weighted-poincare} proves the desired claim.
\end{proof}

\section{Properties of the score function}

We establish in this section several useful properties of the score function.

\begin{lem}
\label{lem:score-upper-bound}
Assume that $\rho \in \cP_{\xi, L, \gamma, M}$ with $\mu_\rho=\mathbb E_\rho[X]$, and let $s(x) = \nabla \log \rho(x)$. Then there exists a numerical constant $C_0>0$, such that
\begin{align*}
\|s(x)\|_2 \le C_0L\bigl(1+\xi\sqrt d+\|x - \mu_\rho\|_2\bigr), \qquad x\in \R^d.
\end{align*}
\end{lem}

\begin{proof}[Proof of \cref{lem:score-upper-bound}]
We first show that the local H\"older smoothness condition implies a global linear growth bound on the score differences. 

Fix $x,y\in\mathbb R^d$, and let
\begin{align*}
r=\|x-y\|_2, \qquad m=\max\{1,\lceil r\rceil\}.
\end{align*}
For $k=0,\ldots,m$, define
\begin{align*}
x_k=y+\frac{k}{m}(x-y).
\end{align*}
By definition, $\|x_k-x_{k-1}\|_2=r/m\le1$ for $k \in [m]$. Hence, by triangle inequality and the local H\"older smoothness condition,
\begin{align*}
\begin{aligned}
\|s(x)-s(y)\|_2 &\le \sum_{k=1}^m
\|s(x_k)-s(x_{k-1})\|_2 \le
mL\left(\frac{r}{m}\right)^\gamma
=
L m^{1-\gamma}r^\gamma.
\end{aligned}
\end{align*}
If $r\le1$, then $m=1$, so the last quantity is bounded by $L r^\gamma \le L$. If $r>1$, then $m\le r+1\le2r$, hence
\begin{align*}
Lm^{1-\gamma}r^\gamma \le L(2r)^{1-\gamma}r^\gamma = 2^{1-\gamma}Lr \le 2Lr.
\end{align*}
Combining both cases gives
\begin{align}
\label{eq:global-linear-score-growth}
\|s(x)-s(y)\|_2 \le 2L\,\bigl(1+\|x-y\|_2\bigr),
\qquad
x,y\in\mathbb R^d.
\end{align}
Next, we show that the score is centered under $\rho$, i.e., $\mathbb E_\rho[s(X)]=0$. 
Let $\eta\in C_c^\infty(\mathbb R^d)$ be a smooth cutoff function satisfying
\begin{align*}
0\le\eta(x)\le1, \qquad
\eta(x)=1 \text{ for }\|x\|_2\le1,
\qquad
\eta(x)=0 \text{ for }\|x\|_2\ge2,
\end{align*}
and $\|\nabla\eta\|_\infty\le C_0'$ for some constant $C_0'>0$. For $B>0$, set $\eta_B(x)=\eta(x/B)$. For every coordinate index $j\in[d]$, integration by parts yields
\begin{align*}
\begin{aligned}
\int_{\mathbb R^d}s_j(x)\eta_B(x)\rho(x)\,\dd x =
\int_{\mathbb R^d}\partial_j\rho(x)\eta_B(x)\,\dd x =
-\int_{\mathbb R^d}\rho(x)\partial_j\eta_B(x)\,\dd x.
\end{aligned}
\end{align*}
Since $\|\nabla\eta_B\|_\infty\le C_0'/B$, the right-hand side satisfies
\begin{align*}
\left|\int_{\mathbb R^d}\rho(x)\partial_j\eta_B(x)\,\dd x\right| \le \frac{C_0'}{B}\int_{\mathbb R^d}\rho(x)\,\dd x = \frac{C_0'}{B} \xrightarrow{B\to\infty} 0.
\end{align*}
Meanwhile, taking $y= \mu_\rho$ in \cref{eq:global-linear-score-growth} yields
\begin{align*}
|s_j(x)|\,\rho(x) \le \|s(x)\|_2\,\rho(x) \le \bigl(\|s(\mu_\rho)\|_2+2L(1+\|x - \mu_\rho\|_2)\bigr)\rho(x).
\end{align*}
Because $\rho$ is sub-Gaussian, $\mathbb E_\rho[\|X - \mu_\rho\|_2]<\infty$, so the majorant is integrable. By the dominated convergence theorem, taking $B\to\infty$ on the left-hand side yields
\begin{align*}
\mathbb E_\rho[s_j(X)]
=
\lim_{B\to\infty}\int_{\mathbb R^d}s_j(x)\eta_B(x)\rho(x)\,\dd x
=
0,
\qquad
j\in[d],
\end{align*}
which implies
\begin{align}
\label{eq:mean-score-zero}
\E_\rho[s_\rho(X)]=0.
\end{align}
We next bound $\|s(\mu_\rho)\|_2$. Using \eqref{eq:mean-score-zero} and Jensen's inequality, followed by \eqref{eq:global-linear-score-growth}, we obtain
\begin{align*}
\begin{aligned}
\|s(\mu_\rho)\|_2
&=
\left\|\mathbb E_\rho[s(\mu_\rho)-s(X)]\right\|_2 \le
\mathbb E_\rho\bigl[\|s(\mu_\rho)-s(X)\|_2\bigr] \le
2L\bigl(1+\mathbb E_\rho[\|X-\mu_\rho\|_2]\bigr).
\end{aligned}
\end{align*}
Since $X$ is $\xi$-sub-Gaussian, its second moment is upper bounded by
\begin{align*}
\mathbb E_\rho\|X - \mu_\rho\|_2^2 = \sum_{j=1}^d\mathbb E_\rho[(X_j - \mu_\rho)^2] \le d\xi^2.
\end{align*}
By Cauchy--Schwarz inequality, $\mathbb E_\rho[\|X- \mu_\rho\|_2] \le \sqrt{\mathbb E_\rho\|X-\mu_\rho\|_2^2} \le \xi\sqrt d$. Hence,
\begin{align*}
\|s(\mu_\rho)\|_2
\le
2L\bigl(1+\xi\sqrt d\bigr).
\end{align*}
Finally, applying the triangle inequality together with \eqref{eq:global-linear-score-growth} at $y=\mu_\rho$,
\begin{align*}
\begin{aligned}
\|s(x)\|_2
&\le
\|s(\mu_\rho)\|_2+\|s(x)-s(\mu_\rho)\|_2 \\
&\le
2L\bigl(1+\xi\sqrt d\bigr) + 2L\bigl(1+\|x -\mu_\rho\|_2\bigr) \\
&\le
C_0L\bigl(1+\xi\sqrt d+\|x -\mu_\rho\|_2\bigr)
\end{aligned}
\end{align*}
for a numerical constant $C_0>0$, completing the proof of the lemma.
\end{proof}
 
\begin{lem}
\label{lem:score-fourth-moment}
Under the assumptions of Lemma~\ref{lem:score-upper-bound}, there exists a numerical constant $C'>0$ such that
\begin{align*}
\mathbb E_\rho\bigl[\|s(X)\|_2^4\bigr]
\le
C' L^4\bigl(1+\xi^4d^2\bigr).
\end{align*}
\end{lem}

\begin{proof}[Proof of \cref{lem:score-fourth-moment}]
By Lemma~\ref{lem:score-upper-bound}, for all $x\in\mathbb R^d$, there exists a numerical constant $C_1 > 0$, such that 
\begin{align*}
\|s(x)\|_2
\le
C_1 L\bigl(1+\xi\sqrt d+\|x - \mu_\rho\|_2\bigr),
\end{align*}
for some numerical constant $C_1>0$. 
Raising both sides to the fourth power and using the inequality $(a+b+c)^4\le 27(a^4+b^4+c^4)$ for non-negative numbers $a,b,c$, we find
\begin{align*}
\|s(x)\|_2^4
\le
C_2 L^4\left(1 + \xi^4d^2 + \|x - \mu_\rho\|_2^4\right),
\end{align*}
where $C_2 = 27 C_1^4$. Taking expectations with respect to $X\sim\rho$ gives
\begin{equation}
\label{eq:score-fourth-moment-expansion}
\mathbb E_\rho\bigl[\|s(X)\|_2^4\bigr]
\le
C_2 L^4\left(1 + \xi^4d^2 + \mathbb E_\rho\bigl[\|X - \mu_\rho\|_2^4\bigr]\right).
\end{equation}
Denote $Y = X- \mu_\rho $, which is a centered sub-Gaussian random variable. To bound $\mathbb E_\rho[\|X - \mu_\rho\|_2^4] = E_\rho[\|Y|_2^4]$, observe that
\begin{align*}
\|Y\|_2^4
=
\left(\sum_{j=1}^d Y_j^2\right)^2
=
\sum_{j=1}^d Y_j^4 + 2\sum_{1\le j<k\le d} Y_j^2 Y_k^2.
\end{align*}
By the Cauchy--Schwarz inequality, $\mathbb E_\rho[Y_j^2 Y_k^2]\le \sqrt{\mathbb E_\rho[Y_j^4]\mathbb E_\rho[Y_k^4]}$. Because $Y$ is also $\xi$-sub-Gaussian, each coordinate $Y_j$ satisfies $\mathbb E_\rho[Y_j^4]\le C_3\xi^4$ for a universal constant $C_3>0$. Therefore,
\begin{align*}
\begin{aligned}
\mathbb E_\rho\bigl[\|Y\|_2^4\bigr]
&\le
\sum_{j=1}^d C_3\xi^4 + 2\sum_{1\le j<k\le d} C_3\xi^4 \\
&=
C_3\xi^4\left(d + 2\binom{d}{2}\right) \\
&=
C_3\xi^4 d^2.
\end{aligned}
\end{align*}
Substituting this upper bound into \eqref{eq:score-fourth-moment-expansion}, we obtain
\begin{align*}
\mathbb E_\rho\bigl[\|s(X)\|_2^4\bigr]
\le
C_2 L^4\bigl(1 + \xi^4d^2 + C_3\xi^4d^2\bigr)
\le
C' L^4\bigl(1+\xi^4d^2\bigr),
\end{align*}
with $C' = C_2\max\{1, 1+C_3\}$. This concludes the proof.
\end{proof}

\section{Proofs for differentially private score estimation}

\subsection{Proof sketch of \cref{thm:DP-score-upper-bound}}
\label{sec:proof-sketch}

The proof of \cref{thm:score-holder-lower} uses a differentially private version of Assouad's lemma \citep{acharya2021differentially}, and we give the details of this proof in Appendix~\ref{proof:thm:score-holder-lower}. 
In the remainder of this section, we sketch the proof of \cref{thm:DP-score-upper-bound}.

\subsubsection*{Score approximation}
We first show that, for a suitably chosen coefficient vector $\theta$, the spline estimator $f_{\theta}$ defined in \cref{eq:f-theta} closely approximates the target score function.
Throughout, we assume $n$ is large enough such that $b_n \leq 1 / (3 \sqrt{d})$ and $R_n > 3b_n$ (recall $R_n$ is defined in \cref{eq:cT-R} and $b_n$ is defined in \cref{thm:DP-score-upper-bound}). 

Recall $\rho_0$ is the target distribution. 
For notational simplicity, we let $s_0 (x) = \nabla \log \rho_0(x)$ and write $s_0 (x) = (s_{1, 0}(x), \ldots, s_{d, 0}(x))^{\top} \in \R^d$. 
For $(a,\nu)\in\mathcal J$, define
\begin{equation} \label{eq:theta-ast}
    \theta^*_{a,\nu} = s_{a, 0}(x_\nu), \qquad x_\nu=b_n\nu.
\end{equation}
By the definition of the effective index set $\mathcal J$, every
grid point $x_\nu$ associated with a nonzero basis function lies in
$[-2R_n - 3 b_n, \, 2 R_n + 3 b_n]^d$. Therefore, by
\cref{lem:score-upper-bound} from the appendix we see that $|\theta^*_{a,\nu}| = |s_{a, 0}(x_\nu)| \le \cR_n$ (recall $\cR_n$ is defined in \cref{eq:piD}). 
For every $q\in[d]$ and every pair
$(a,\nu),(a,\nu+e_q)\in\mathcal J$, the corresponding grid points satisfy $\|x_{\nu+e_q}-x_\nu\|_2=b_n \le 1$. 
The local $\gamma$-H\"older smooth condition therefore implies that
\begin{align*}
|\theta^*_{a,\nu+e_q}-\theta^*_{a,\nu}|
\le
\|s_0(x_{\nu+e_q})-s_0(x_\nu)\|_2
\le
Lb_n^\gamma.
\end{align*}
As a consequence, the vector $\theta^{\ast}$ defined in \cref{eq:theta-ast} satisfies $\theta^{\ast} \in \Theta_{n,\gamma}$. 
The next lemma shows that $f_{\theta^{\ast}}$ provides an accurate approximation to the target score $\nabla \log \rho_0$. 

\begin{lem}[Score approximation]
\label{lem:local-holder-approximation}
Assume $n$ is large enough such that $b_n \leq 1 / (3 \sqrt{d})$ and $R_n > 3b_n$.
Then there exists a constant $C_{\xi,L,d, M}>0$ that depends only on $(\xi, L, d, M)$, such that for all $\rho_0 \in \cP_{\xi, L, \gamma, M}$,

\begin{align*}
 R(f_{\theta^*},\rho_0) \le C_{\xi,L,d,M} \left( b_n^{2\gamma} + n^{-5} \right).
\end{align*}
\end{lem}

\begin{proof}[Proof of \cref{lem:local-holder-approximation}]

We prove \cref{lem:local-holder-approximation} in Appendix \ref{proof:lem:local-holder-approximation}. 
	
\end{proof}

\subsubsection*{Privacy guarantee for the distribution induced by the exponential mechanism}
We next show that drawing a sample from the distribution $\pi_{\cD}$ induced by the exponential mechanism satisfies $(\alpha,0)$-DP. 
We begin by analyzing the sensitivity of the score matching loss.
Recall that $Z_i \in \Omega_{R_n}$ for all $i \in [n]$, and $R_n$ is defined in \cref{eq:cT-R}. 
The following lemma upper bounds the score matching loss for all $\theta \in \Theta_{n, \gamma}$.

\begin{lem}
\label{lemma:sensitivity}
	For all $\theta \in \Theta_{n, \gamma}$ and $z \in \Omega_{R_n}$, it holds that 
	\begin{align*}
		\big| \ell_{\theta}(z) \big| \leq  d \cR_n^2 +  d L b_n^{\gamma-1}. 
	\end{align*}
\end{lem}

\begin{proof}[Proof of \cref{lemma:sensitivity}]
	We prove \cref{lemma:sensitivity} in Appendix \ref{proof:lemma:sensitivity}. 
\end{proof}

By \cref{lemma:sensitivity}, for any two neighboring datasets $\cD$ and $\cD'$ consisting of samples $X_1,\ldots,X_n$ and $X_1',\ldots,X_n'$, respectively, it holds that
\begin{align}
\label{eq:sensitivity}
	\Big| \sum_{i = 1}^n \ell_{\theta}(\cT_{R_n}(X_i )) - \sum_{i = 1}^n \ell_{\theta}(\cT_{R_n}(X_i')) \Big| \leq 2d \cR_n^2 + 2d Lb_n^{\gamma-1}. 
\end{align}
By \cref{lem:exponential-mechanism} below, the sensitivity bound in \cref{eq:sensitivity} ensures that drawing one sample from $\pi_{\cD}$ satisfies $(\alpha,0)$-DP.

\begin{lem}
\label{lem:exponential-mechanism}
	The mechanism $\cD \mapsto \theta \sim \pi_\cD$ is $(\alpha, 0)$-DP. 
\end{lem}

\begin{proof}[Proof of \cref{lem:exponential-mechanism}]
	We prove \cref{lem:exponential-mechanism} in Appendix \ref{proof:lem:exponential-mechanism}. 
\end{proof}

\subsubsection*{Impact of approximate sampling on privacy and accuracy}
In practice, exact sampling from $\pi_{\cD}$ is rarely feasible.
We next show that sampling from a distribution $\hat\pi_\cD$ that is close to $\pi_\cD$ in total variation distance ensures approximate DP. 

\begin{lem}
\label{lem:privacy-approximate-sampling}
	We assume that $\TV(\pi_\cD, \hat \pi_\cD) \leq \eta$ for every dataset $\cD$ and $\hat \pi_{\cD}(\Theta_{n, \gamma}) = 1$. 
	Then, for every $\rho \in \cP_{\xi, L, \gamma, M}$ the mechanism $\cD \mapsto \theta \sim \hat \pi_\cD$ is $(\alpha, (1 + e^{\alpha}) \eta)$-DP.  
\end{lem}

\begin{proof}[Proof of \cref{lem:privacy-approximate-sampling}]
	We prove \cref{lem:privacy-approximate-sampling} in Appendix \ref{proof:lem:privacy-approximate-sampling}
\end{proof}

We next show that the risk guarantee for the score estimator continues to hold under approximate sampling from $\pi_{\cD}$, up to a term controlled by the sampling error.   

\begin{lem}
\label{lemma:approximate-sampling-preserves-risk}
	We assume that $\TV(\pi_\cD, \hat \pi_\cD) \leq \eta$ for every dataset $\cD$ and $\hat \pi_{\cD}(\Theta_{n, \gamma}) = 1$. Then, for every $\rho \in \cP_{\xi, L, \gamma, M}$ 
	\begin{align*}
		 \E_{\cD, \theta \sim \hat\pi_\cD} \big[ R(f_{\theta}, \rho) \big] \leq \E_{\cD, \theta \sim \pi_\cD} \big[ R(f_{\theta}, \rho) \big] + 2\eta\, \big( d \cR_n^2 +  \sqrt{C'} L^2 (1 + \xi^2 d) \big), 
	\end{align*} 
	where the expectation is over i.i.d. samples from $\rho$ and $\theta$, and $C'$ is a positive numerical constant that appears in \cref{lem:score-fourth-moment}.
\end{lem}

\begin{proof}[Proof of \cref{lemma:approximate-sampling-preserves-risk}]
	We prove \cref{lemma:approximate-sampling-preserves-risk} in Appendix \ref{proof:lemma:approximate-sampling-preserves-risk}. 
\end{proof}

Together, \cref{lem:privacy-approximate-sampling} and \cref{lemma:approximate-sampling-preserves-risk} show that approximate sampling from $\pi_{\cD}$ satisfies a privacy guarantee and yields a score estimator whose risk exceeds that of the exact exponential mechanism by at most a term controlled by the sampling error. 
Next, we show that for the parameter choices in \cref{thm:DP-score-upper-bound}, we can upper bound the error incurred by \cref{alg:projected_LMC} in sampling from $\tilde{\pi}_{\cD}$. 
We also upper bound the total variation distance between $\tilde{\pi}_{\cD}$ and $\pi_{\cD}$ in \cref{lemma:ridge-regularized-distribution} below.  

\begin{lem}
\label{lemma:ridge-regularized-distribution}	

For $\pi_\cD$ and $\tilde \pi_\cD$ defined separately in \cref{eq:piD,eq:pi-tilde}, it holds that 
\begin{align*}
	\TV(\pi_\cD, \, \tilde \pi_\cD) \leq e^{\min (r_n, 2\cR_n^{-2}\log(1 + \Upsilon_{\alpha, \beta} / 2) ) \cR_n^2 / 2} - 1, 
\end{align*}
where we recall $r_n = \max( n^{-2\gamma / (d + 2\gamma +2)}, \, (n \alpha)^{-2\gamma / (d + \gamma +1)})$, $\Upsilon_{\alpha, \beta} = \beta / (1 + e^{\alpha})$ and $\cR_n$ is defined in \cref{eq:piD}. 
\end{lem}

\begin{proof}[Proof of \cref{lemma:ridge-regularized-distribution}]
	We prove \cref{lemma:ridge-regularized-distribution} in Appendix \ref{proof:lemma:ridge-regularized-distribution}. 
\end{proof}

We next show that, with suitable parameter choices, \cref{alg:projected_LMC} produces an approximate sample from $\tilde{\pi}_{\cD}$. 
The following lemma is an adaptation of \cite[Theorem 1]{bubeck2018sampling}. 

\begin{lem}
\label{thm:sampling}
	Let $\hat{\pi}_{\cD,\cL}$ denote the conditional distribution of the output $\theta_{\cL}$ of \cref{alg:projected_LMC} given the dataset $\cD$.
    Then for $\delta = \tilde \Theta\big(|\cJ|\cR_n^2 / \cL \big )$ and 
	\begin{align*}
		& \cL = \tilde \Omega \Big( \,\frac{|\cJ|^3 \cR_n^6 \max \big( |\cJ|, \,\bar \lambda\sqrt{|\cJ| } \cR_n L b_n^{\gamma}, \, \bar \lambda{|\cJ|} \cR_n^2 + \sqrt{|\cJ|} \cR_n n \alpha \sqrt{2d} Q_{n, \gamma}^{-1} b_n^{-1} \big)^{12}}{\varepsilon^{12} L^6 b_n^{6\gamma}}  \Big), \\
		& \bar \lambda = \frac{\min (r_n, 2\cR_n^{-2}\log(1 + \Upsilon_{\alpha, \beta} / 2) )}{|\cJ|} + \frac{nd\alpha}{4Q_{n, \gamma}},  
	\end{align*}
	it holds that $\TV(\tilde \pi_\cD, \,\hat \pi_{\cD, \cL}) \leq \varepsilon$. 
\end{lem}

\begin{proof}[Proof of \cref{thm:sampling}]
	We prove \cref{thm:sampling} in Appendix \ref{proof:thm:sampling}. 
\end{proof}

Together, \cref{lemma:ridge-regularized-distribution} and \cref{thm:sampling} imply that the distribution of $\theta_{\cL}$ returned by \cref{alg:projected_LMC} is close to $\pi_{\cD}$ in total variation under the specified parameter choices. 
By \cref{lem:privacy-approximate-sampling} and \cref{lemma:approximate-sampling-preserves-risk}, the resulting score estimator $f_{\hat\theta_{\mathrm{LMC}}}$ satisfies $(\alpha,\beta)$-DP, and its risk is at most that of $f_{\hat\theta_{\mathrm{EM}}}$ with $\hat\theta_{\mathrm{EM}}\sim\pi_{\cD}$ plus a term controlled by the total variation distance $\TV(\pi_\cD, \,\hat \pi_{\cD, \cL})$.
We then upper bound the risk of $f_{\hat\theta_{\mathrm{EM}}}$ to complete the proof. 

\subsubsection*{Upper bounding the risk of the estimator returned by the exponential mechanism}
Finally, we upper bound $R(f_{\hat\theta_{\mathrm{EM}}}, \rho_0)$ for $\hat\theta_{\mathrm{EM}} \sim \pi_{\cD}$. 
Recall that the score matching loss $\ell_{\theta}$ is defined in \cref{eq:score-matching-loss}.
For a distribution $\rho$ over $\R^d$ and $\theta \in \Theta_{n, \gamma}$, define $M_{\rho}(\theta) = \E_{\rho} [ \ell_{\theta}(X) ]$. 
For $\rho \in \mathcal{P}_{\xi, L, \gamma, M}$ and $\theta \in \Theta_{n, \gamma}$, define   
\begin{align}
\label{eq:R-rho-theta}
	R_{\rho}(\theta) = R(f_{\theta}, \rho) =  \int \|f_{\theta}(x) - \nabla \log \rho(x)\|_2^2 \, \rho(x) \mathrm{d} x. 
\end{align}
The following lemma relates $M_\rho$ and $R_\rho$. 
\begin{lem}
\label{lemma:M-R}
	For all $\theta \in \Theta_{n, \gamma}$ and $\rho \in \cP_{\xi, L, \gamma, M}$, it holds that 
	\begin{align*}
		R_{\rho}(\theta) = 2M_{\rho}(\theta) + \int \|\nabla \log \rho (x)\|_2^2 \, \rho(x) \mathrm{d} x. 
	\end{align*}
\end{lem}

\begin{proof}[Proof of \cref{lemma:M-R}]
	We prove \cref{lemma:M-R} in Appendix \ref{proof:lemma:M-R}. 
\end{proof}

For a dataset $\cD$ consisting of $n$ samples $X_1,\ldots,X_n$ and $\theta \in \Theta_{n, \gamma}$, we define the empirical score-matching loss as $\hat M_\cD(\theta)=n^{-1} \sum_{i=1}^n\ell_\theta(Z_i)$, where recall $Z_i=\cT_{R_n}(X_i)$.
We next show that, when $X_1,\ldots,X_n$ are i.i.d. samples from $\rho$ for $\rho\in\cP_{\xi,L,\gamma,M}$, the empirical loss $\hat M_\cD(\theta)$ uniformly approximates the population loss $M_\rho(\theta)$ for sufficiently large $n$.

\begin{lem}\label{lemma:ULLN}

Assume that $n$ is sufficiently large such that $R_n\cR_n \geq 2C_\chi/\sqrt d$, where we recall $C_{\chi}$ is the numerical constant from \cref{eq:chiR}.
For all $\rho \in \cP_{\xi, L, \gamma, M}$, $X_1, \ldots, X_n \sim_{i.i.d.} \rho$, and a dataset $\cD$ consisting of these $n$ samples, it holds that
\begin{align*}
	& \E_{\cD} \left[ \sup_{\theta \in \Theta_{n, \gamma}} \big|  M_{\rho}(\theta) - \hat M_\cD(\theta) \big| \right] \\
	& \leq 2 n^{-10} (d \cR_n^2 + d L b_n^{\gamma-1}) + 2C_u( d \cR_n^2 + d L b_n^{\gamma-1})\sqrt{\frac{|\cJ| }{n} \cdot \log \Big( 1 + 2\cR_n(d \cR_n + 2d b_n^{-1}) \Big) } \\
	&\,\,\,\,\, + 4C_u \sqrt{\frac{|\cJ| (d \cR_n^2 + d L b_n^{\gamma-1})}{n}}, 
\end{align*}
where $C_u$ is a positive numerical constant. 
\end{lem}

\begin{proof}[Proof of \cref{lemma:ULLN}]
	We prove \cref{lemma:ULLN} in Appendix \ref{proof:lemma:ULLN}. 
\end{proof}

To bound $R(f_{\hat\theta_{\mathrm{EM}}},\rho_0)$, we first apply \cref{lemma:M-R} to reduce the problem to bounding $M_{\rho_0}(\hat\theta_{\mathrm{EM}})$. By \cref{lemma:ULLN}, it then suffices to establish an upper bound on its empirical counterpart $\hat{M}_{\cD}(\hat\theta_{\mathrm{EM}})$, which we provide in \cref{lemma:exponential-mechanism-risk} below. 

\begin{lem}
\label{lemma:exponential-mechanism-risk}
	Recall that $\theta^{\ast} \in \Theta_{n,\gamma}$ is defined in \cref{eq:theta-ast}. 
    Then for every dataset $\cD$ and every large enough $n$ that satisfies $\cR_n > n^{-10}$, it holds that 
	\begin{align*}
		\E_{\hat\theta_{\mathrm{EM}} \sim \pi_\cD} [\hat M_\cD(\hat\theta_{\mathrm{EM}})] \leq \hat M_\cD(\theta^{\ast}) + \Big( d \cR_n + \frac{2d}{b_n} \Big) n^{-10} + \frac{4Q_{n, \gamma}|\cJ|}{n \alpha} \Big(  \log (2 \cR_n) +  10\log n \Big).
	\end{align*}
\end{lem}

\begin{proof}[Proof of \cref{lemma:exponential-mechanism-risk}]

We prove \cref{lemma:exponential-mechanism-risk} in Appendix \ref{proof:lemma:exponential-mechanism-risk}. 
	
\end{proof}

By \cref{lemma:ULLN}, the empirical objective $\hat M_\cD(\theta^{\ast})$ is close to its population counterpart $M_{\rho_0}(\theta^{\ast})$. We can then bound the latter using \cref{lem:local-holder-approximation} and \cref{lemma:M-R}.
Putting together the preceding results completes the proof of \cref{thm:DP-score-upper-bound}, see Appendix \ref{proof:thm:DP-score-upper-bound} for more details.

\subsection{Proof of \cref{lem:local-holder-approximation}}
\label{proof:lem:local-holder-approximation}

Recall that for $\theta^{\ast} \in \R^{\cJ}$, 
\begin{align*}
f_{\theta^*}(x)=\chi_{R_n}(x)g_{\theta^*}(x), \qquad g_{\theta^*,a}(x)=\sum_{\nu\in\mathbb Z^d}\theta^*_{a,\nu}\psi_\nu(x).
\end{align*}
We decompose the risk as
\[
\begin{aligned}
& R(f_{\theta^*},\rho_0) =
\int_{\Omega_{R_n}}
\|f_{\theta^*}(x)-s_0(x)\|_2^2\,\rho_0(x)\,\dd x+\int_{\Omega_{2R_n}\setminus\Omega_{R_n}}
\|f_{\theta^*}(x)-s_0(x)\|_2^2\,\rho_0(x)\,\dd x \\
& \qquad \qquad \qquad + \int_{\Omega_{2R_n}^c}
\|f_{\theta^*}(x)-s_0(x)\|_2^2\,\rho_0(x)\,\dd x.
\end{aligned}
\]
We bound the three terms separately.
For $x\in\Omega_{R_n}$, we have $\chi_{R_n}(x)=1$. Moreover, if
$\psi_\nu(x)\neq0$, then
\[
\|x-x_\nu\|_2\le3\sqrt d\,b_n\le1.
\]
Using the partition-of-unity property of the B-splines and the local
$\gamma$-H\"older smooth condition, we obtain
\[
\begin{aligned}
\|f_{\theta^*}(x)-s_0(x)\|_2
&=
\left\|
\sum_\nu
\bigl(s_0(x_\nu)-s_0(x)\bigr)\psi_\nu(x)
\right\|_2 \\
&\le
\sum_\nu
\|s_0(x_\nu)-s_0(x)\|_2\psi_\nu(x) \\
&\le
L(3\sqrt d\,b_n)^\gamma.
\end{aligned}
\]
Therefore, there exists a positive numerical constant $C$, such that
\[
\int_{\Omega_{R_n}}
\|f_{\theta^*}(x)-s_0(x)\|_2^2\,\rho_0(x)\,\dd x
\le
C L^2d^\gamma b_n^{2\gamma}.
\]
For $x\in\Omega_{2R_n}\setminus\Omega_{R_n}$, we have
\[
\|f_{\theta^*}(x)\|_2^2
\le d\cR_n^2.
\]
Hence,
\[
\begin{aligned}
&\int_{\Omega_{2R_n}\setminus\Omega_{R_n}}
\|f_{\theta^*}(x)-s_0(x)\|_2^2\rho_0(x)\,\dd x \le
2d\cR_n^2\mathbb P_{\rho_0}(X\notin\Omega_{R_n})
+
2\mathbb E_{\rho_0}
\left[
\|s_0(X)\|_2^2\mathbf 1\{X\notin\Omega_{R_n}\}
\right].
\end{aligned}
\]
By our choice of $R_n$, the sub-Gaussian tail bound guarantees $\mathbb P_{\rho_0}(X\notin\Omega_{R_n})\le n^{-10}$. Using Cauchy--Schwarz inequality and \cref{lem:score-fourth-moment}, the second term is upper bounded by
\[
2\mathbb E_{\rho_0}
\left[
\|s_0(X)\|_2^2\mathbbm 1\{X\notin\Omega_{R_n}\}
\right]
\le
2\mathbb P_{\rho_0}(X\notin\Omega_{R_n})^{1/2}
\left(
\mathbb E_{\rho_0} \left[\|s_0(X)\|_2^4 \right]
\right)^{1/2}
\le
C_{\xi,L,d}'n^{-5},
\]
where $C_{\xi,L,d}'$ depends only on $(\xi, L, d)$. 
As for the first term, observe that for all sufficiently large $n$, we have
\[
2d\,\cR_n^2\,\mathbb P_{\rho_0}(X\notin\Omega_{R_n})
\le
C_{\xi,L,d,M}''n^{-5},
\]
where $C_{\xi,L,d,M}''$ depends only on $(\xi, L, d, M)$. 
Consequently,
\[
\int_{\Omega_{2R_n}\setminus\Omega_{R_n}}
\|f_{\theta^*}(x)-s_0(x)\|_2^2\,\rho_0(x)\,\dd x
\le
(C_{\xi,L,d}' + C_{\xi,L,d,M}'')\,n^{-5}.
\]
Finally, observe that $f_{\theta^*}(x)=0$ for $x\notin\Omega_{2R_n}$. Therefore,
\[
\begin{aligned}
\int_{\Omega_{2R_n}^c}
\|f_{\theta^*}(x)-s_0(x)\|_2^2\,\rho_0(x)\,\dd x
&=
\mathbb E_{\rho_0}
\left[
\|s_0(X)\|_2^2\mathbbm 1\{X\notin\Omega_{2R_n}\}
\right] \le
C_{\xi,L,d,M}'''n^{-5},
\end{aligned}
\]
where $C_{\xi,L,d,M}''' > 0$ depends only on $(\xi, L, d, M)$. 
Combining the preceding bounds completes the proof.

\subsection{Proof of \cref{lemma:sensitivity}}
\label{proof:lemma:sensitivity}

By definition, for all $\theta \in \Theta_{n, \gamma}$ and $(a, \nu) \in \cJ$, we have $|\theta_{a, \nu}| \leq \cR_n$. As a consequence, $\|f_{\theta, a}\|_{\infty} \leq \cR_n$ for all $a \in [d]$, and 
\begin{align*}
	\|f_{\theta}(z)\|_2^2 \leq d \cR_n^2
\end{align*} 
for all $z \in \Omega_{R_n}$. 
As for the divergence term, note that for all $z \in \Omega_{R_n}$, 
\begin{align*}
	& \diver f_{\theta}(z) = \sum_{a = 1}^d \partial_a f_{\theta, a}(z) = \sum_{a = 1}^d \partial_a g_{\theta, a}(z), \\
	& g_{\theta, a}(z) = \sum_{\nu \in \Z^d} \theta_{a, \nu} \psi_{\nu}(z). 
\end{align*}
Observe that for all $z \in \Omega_{R_n}$, 
\begin{align*}
	\partial_a g_{\theta, a}(z) = \sum_{\nu \in \Z^d} \theta_{a, \nu} \psi'_{\nu_a}(z_a) \prod_{r \in [d] \backslash \{a\}} \psi_{\nu_r}(z_r). 
\end{align*}
For every fixed $(\nu_r)_{r \in [d] \backslash \{a\}}$ and $z \in \Omega_{R_n}$, summing over $\nu_a \in \Z$ and using \cref{eq:gcp-upper} yields  
\begin{align*}
	\Big| \sum_{\nu_a \in \Z} \theta_{a, \nu} \psi_{\nu_a}'(z_a) \Big| \leq \frac{1}{b_n} \sup_{(a, \nu), (a, \nu + e_a) \in \cJ} |\theta_{a, \nu} - \theta_{a, \nu + e_a}| \leq L b_n^{\gamma-1}. 
\end{align*}
As a consequence, for all $z \in \Omega_{R_n}$ 
\begin{align*}
	\big| \partial_a g_{\theta, a}(z) \big| \leq  L b_n^{\gamma - 1}\, \Big| \sum_{h \in [d] \backslash \{a\}  } \sum_{\nu_h \in \Z} \prod_{r \in [d] \backslash \{a\}} \psi_{\nu_r}(z_r) \Big| = Lb_n^{\gamma-1}. 
\end{align*}
Putting together the above inequalities, we can conclude that 
\begin{align}
\label{eq:bounded-ell}
	\big| \ell_{\theta}(z) \big| \leq \frac{1}{2} \|f_{\theta}(z)\|_2^2 + \big|\diver f_{\theta}(z) \big| \leq d \cR_n^2 + d Lb_n^{\gamma-1}. 
\end{align}

\subsection{Proof of \cref{lem:exponential-mechanism}}
\label{proof:lem:exponential-mechanism}

For a dataset $\cD$, define $S_\cD(\theta) = \sum_{i = 1}^n \ell_{\theta}(\cT_{R_n}(X_i))$. 
Let $\cD, \cD'$ be two neighboring datasets. By \cref{eq:sensitivity}, we know that $|S_\cD(\theta) - S_{\cD'}(\theta)| \leq 2Q_{n, \gamma}$.   
	We therefore have 
	\begin{align*}
		e^{-\alpha S_\cD(\theta) / (4Q_{n, \gamma})} \leq e^{-\alpha S_{\cD'}(\theta) / (4Q_{n, \gamma}) + \alpha  / 2}. 
	\end{align*}
	We write 
	\begin{align*}
		\pi_\cD(\theta) = \frac{1}{Z_\cD} \mathbbm{1}\{\theta \in \Theta_{n, \gamma}\} \exp \Big( -\frac{\alpha}{4Q_{n, \gamma}} \sum_{i = 1}^n \ell_{\theta}(\cT_{R_n}(X_i)) \Big), 
	\end{align*}
	then we have $Z_{\cD'} \leq e^{\alpha / 2} Z_{\cD}$. Combining these bounds, we see that for all $\theta \in \Theta_{n, \gamma}$, 
	\begin{align*}
		{\pi_\cD(\theta)} \leq e^{\alpha} \pi_{\cD'}(\theta). 
	\end{align*}
	Therefore, for every measurable set $\Theta'$, 
	\begin{align*}
		{\pi_\cD(\Theta')} \leq e^{\alpha} {\pi_{\cD'}(\Theta')}, 
	\end{align*}
	which proves $(\alpha, 0)$-differentially private. 
	
\subsection{Proof of \cref{lem:privacy-approximate-sampling}}
\label{proof:lem:privacy-approximate-sampling}

	Let $\cD, \cD'$ be two neighboring datasets and $S$ be any measurable set. By \cref{lem:exponential-mechanism}, we know that 
	\begin{align*}
		\hat \pi_\cD(S) \leq \pi_\cD(S) + \eta \leq e^{\alpha} \pi_{\cD'}(S) + \eta \leq e^{\alpha} \hat \pi_{\cD'}(S) + (1 + e^{\alpha}) \eta. 
	\end{align*}
	The above upper bound by the definition of approximate differential privacy already proves that $\cD \mapsto \theta \sim \hat \pi_\cD$ is $(\alpha, (1 + e^{\alpha}) \eta)$-differentially private. 
	
\subsection{Proof of \cref{lemma:approximate-sampling-preserves-risk}}
\label{proof:lemma:approximate-sampling-preserves-risk}

We next establish a uniform upper bound on the risk. 
Specifically, for every $\rho \in \cP_{\xi,L,\gamma,M}$ and $\theta \in \Theta_{n,\gamma}$, we show that 
\begin{align}
\label{eq:uniform-risk-upper}
	R_{\rho}(\theta) \leq 2d \cR_n^2 + 2 \sqrt{C'} L^2 (1 + \xi^2 d),  
\end{align}
where $C' > 0$ is a numerical constant that appears in \cref{lem:score-fourth-moment}.

	By triangle inequality, 
	\begin{align*}
		R_{\rho}(\theta) \leq 2 \int \|f_{\theta}(x)\|_2^2 \, \rho(x) \mathrm{d} x + 2 \int \|\nabla \log \rho(x)\|_2^2 \, \rho(x) \mathrm{d} x. 
	\end{align*}
	By \cref{lem:score-fourth-moment} and the Cauchy–Schwarz inequality, we have $\E_{\rho}[\|\nabla \log \rho(X)\|_2^2] \leq \sqrt{C'} L^2 (1 + \xi^2 d)$. Since $\|f_{\theta, a}\|_{\infty} \leq \cR_n$ for all $a \in [d]$ and $\theta \in \Theta_{n, \gamma}$, we know that $\int \|f_{\theta}(x)\|_2^2 \, \rho(x) \mathrm{d} x \leq d \cR_n^2$. 
	Putting together these upper bounds, we see that for all $\rho \in \cP_{\xi, L, \gamma, M}$ and $\theta \in \Theta_{n, \gamma}$, \cref{eq:uniform-risk-upper} holds. 

By \cref{eq:uniform-risk-upper}, we know that $R(f_{\theta}, \rho) \leq 2 d \cR_n^2 + 2 \sqrt{C'} L^2 (1 + \xi^2 d)$ for every $\rho \in \cP_{\xi, L, \gamma, M}$ and $\theta \in \Theta_{n, \gamma}$. 
	For every fixed dataset $\cD$ and $\rho \in \cP_{\xi, L, \gamma, M}$, 
	\begin{align*}
		\E_{\theta \sim \hat \pi_\cD} \big[ R(f_{\theta}, \rho) \big] \leq \E_{\theta \sim \pi_\cD} \big[ R(f_{\theta}, \rho) \big] +2 \eta \big( d\cR_n^2 + \sqrt{C'}L^2 (1 + \xi^2 d) \big). 
	\end{align*}
	Taking expectation over $\cD$ proves the lemma.

\subsection{Proof of \cref{lemma:ridge-regularized-distribution}}
\label{proof:lemma:ridge-regularized-distribution}

Let 
\begin{align*}
	& \pi_\cD(\theta) = \frac{1}{Z_\cD} \mathbbm{1}\{\theta \in \Theta_{n, \gamma}\} \exp \Big( -\frac{\alpha}{4Q_{n, \gamma}} \sum_{i = 1}^n \ell_{\theta}(Z_i) \Big), \\
	& \tilde \pi_\cD(\theta ) = \frac{1}{\tilde Z_\cD} \mathbbm{1}\{\theta \in \Theta_{n, \gamma}\} \exp \Big( -\frac{\alpha}{4Q_{n, \gamma}} \sum_{i = 1}^n \ell_{\theta}(Z_i) - \frac{\min (r_n, 2\cR_n^{-2}\log(1 + \Upsilon_{\alpha, \beta} / 2) )}{2|\cJ|} \|\theta\|_2^2 \Big), 
\end{align*}
where $Z_\cD$, $\tilde Z_\cD$ are normalizing constants. 
Note that for all $\theta \in \Theta_{n, \gamma}$ we have $\|\theta\|_2^2 \leq |\cJ| \cR_n^2$. 
Therefore, 
\begin{align*}
	 \tilde Z_\cD \leq Z_\cD \leq  e^{ \min (r_n, 2\cR_n^{-2}\log(1 + \Upsilon_{\alpha, \beta} / 2) ) \cR_n^2 / 2} \tilde Z_\cD. 
\end{align*}
Using the above inequalities, we see that for all $\theta \in \Theta_{n, \gamma}$, 
\begin{align*}
	\frac{\pi_\cD(\theta)}{\tilde \pi_\cD(\theta)} &= \frac{\tilde Z_\cD}{Z_\cD} \cdot \exp \Big(\, \frac{\min (r_n, 2\cR_n^{-2}\log(1 + \Upsilon_{\alpha, \beta} / 2) )}{2|\cJ|} \|\theta\|_2^2\, \Big) \\
    &\in \Big[e^{-\min (r_n, 2\cR_n^{-2}\log(1 + \Upsilon_{\alpha, \beta} / 2) ) \cR_n^2 / 2}, \,  e^{\min (r_n, 2\cR_n^{-2}\log(1 + \Upsilon_{\alpha, \beta} / 2) ) \cR_n^2 / 2} \Big],  
\end{align*}
which further implies that $\TV(\pi_\cD, \, \tilde \pi_\cD) \leq e^{\min (r_n, 2\cR_n^{-2}\log(1 + \Upsilon_{\alpha, \beta} / 2) ) \cR_n^2 / 2} - 1$. The proof is done. 

\subsection{Proof of \cref{thm:sampling}}
\label{proof:thm:sampling}

Recall $\psi_j$ is defined in Section \ref{sec:splines}. 
Using the spline identities $
\sum_{j \in \Z}\psi_j'(t)^2\le 2 b_n^{-2}$ and  $\sum_{j \in \Z}\psi_j(t)^2\le1$, we see that for all $z \in \Omega_{R_n}$, 
\begin{align}
\label{eq:h-norm}
	\|h(z)\|_2^2 \le\sum_{a=1}^d \Big(\sum_{j \in \Z}\psi_j'(z_a)^2\Big) \prod_{q \in [d] \backslash \{a\}}\Big(\sum_{j \in \Z}\psi_j(z_q)^2\Big) \le\frac{2d}{b_n^2}.
\end{align}
Observe that $\|\Phi(z)\|_F \leq \sqrt{d}$ and $\|h(z)\|_2 \leq \sqrt{2d} \,b_n^{-1}$ by \cref{eq:h-norm} for all $z \in \Omega_{R_n}$, hence 
\begin{align}
\label{eq:lambda-min-max} 
\begin{split}
	& \frac{\min (r_n, 2\cR_n^{-2}\log(1 + \Upsilon_{\alpha, \beta} / 2) )}{|\cJ|} \leq \lambda_{\min}(\Sigma_\cD) \leq \lambda_{\max}(\Sigma_\cD) \leq \frac{\min (r_n, 2\cR_n^{-2}\log(1 + \Upsilon_{\alpha, \beta} / 2) )}{|\cJ|} + \frac{nd\alpha}{4Q_{n, \gamma}}, \\
	& \|\omega_\cD\|_2 \leq \frac{n\alpha \sqrt{2d}}{4Q_{n, \gamma} b_n}. 
\end{split}
\end{align}
Our proof uses Theorem 1 of \cite{bubeck2018sampling}, which we present here a rescaled specialization for reference convenience. 

\begin{lem}[Theorem 1 of \cite{bubeck2018sampling}]
	Let $\Theta_{n, \gamma} \subseteq \R^{|\cJ|}$ be a convex body such that $\vec{0}_{|\cJ|} \in \Theta_{n, \gamma}$. 
	In addition, $\Theta_{n, \gamma}$ contains a Euclidean ball of radius $r$, and is contained in a Euclidean ball of radius $R$. 
	Then $\TV(\tilde \pi_\cD, \hat \pi_{\cD, \cL}) \leq \varepsilon$ provided that $\delta = \tilde\Theta(R^2  \cL^{-1})$ and that 
	\begin{align*}
		\cL = \tilde \Omega \Big( \frac{(R / r)^6 \max\big(|\cJ|, R (\lambda_{\max}(\Sigma_\cD) \sqrt{|\cJ|} \cR_n + \|\omega_D\|_2) , R r  \lambda_{\max}(\Sigma_\cD) \big)^{12}}{\varepsilon^{12} } \Big). 
	\end{align*}
\end{lem}

By definition, $\Theta_{n, \gamma}$ is contained in a ball with radius $\sqrt{|\cJ|} \cR_n$, and contains a ball with of radius $\min (\cR_n, Lb_n^{\gamma}/2)$. 
Since $b_n\leq1$ and the numerical constant in the definition
of $\cR_n$ may be chosen to satisfy $C\geq1$, we have
$\cR_n\geq L\geq Lb_n^\gamma/2$. Consequently,
$\Theta_{n,\gamma}$ contains the Euclidean ball centered
at the origin with radius $Lb_n^\gamma/2$. 
In addition, by \cref{eq:lambda-min-max} we have $\lambda_{\max}(\Sigma_\cD) \leq \bar \lambda$ and $\|\omega_\cD\|_2 \leq \frac{n\alpha \sqrt{2d}}{4Q_{n, \gamma} b_n}$. 
Substituting in these values completes the proof of the lemma.

\subsection{Proof of \cref{lemma:M-R}}
\label{proof:lemma:M-R}

Expanding the squared norm in the definition of \(R_\rho(\theta)\), we obtain
\begin{align*}
R_\rho(\theta)
&=
\int_{\mathbb{R}^d}
\left\|
f_\theta(x)-\nabla \log \rho(x)
\right\|_2^2
\rho(x)\dd x \\
&=
\int_{\mathbb{R}^d}
\|f_\theta(x)\|_2^2 \rho(x)\dd x
-
2\int_{\mathbb{R}^d}
\left\langle
f_\theta(x),\nabla \log \rho(x)
\right\rangle
\rho(x)\dd x
+
\int_{\mathbb{R}^d}
\|\nabla \log \rho(x)\|_2^2
\rho(x)\dd x .
\end{align*}
Since $\nabla \log \rho(x)\,\rho(x)=\nabla \rho(x)$, the cross term above can be written as
\begin{align*}
\int_{\mathbb{R}^d}\left\langle f_\theta(x),\nabla \log \rho(x) \right\rangle \rho(x)\dd x = \sum_{j=1}^d \int_{\mathbb{R}^d} f_{\theta,j}(x)\,\partial_j\rho(x)\,\dd x.
\end{align*}
The regularity and tail conditions imposed on $\rho\in\mathcal P_{\xi,L,\gamma, M}$
justify integration by parts, with the boundary terms vanishing. Hence,
for each $j\in[d]$,
$\int_{\mathbb{R}^d}f_{\theta,j}(x)\,\partial_j\rho(x)\dd x=-\int_{\mathbb{R}^d}\partial_j f_{\theta,j}(x)\,\rho(x)\dd x .$
Summing over $j$ gives
\begin{align*}
\int_{\mathbb{R}^d}\left\langle f_\theta(x),\nabla \log \rho(x) \right\rangle \rho(x)\dd x=-\int_{\mathbb{R}^d} \operatorname{div} f_\theta(x)\,\rho(x)\dd x .
\end{align*}
Substituting this identity into the expansion of \(R_\rho(\theta)\),
we obtain
\begin{align*}
R_\rho(\theta) &= \int_{\mathbb{R}^d} \left(\|f_\theta(x)\|_2^2 + 2\operatorname{div}f_\theta(x) \right) \rho(x)\,\dd x + \int_{\mathbb{R}^d} \|\nabla\log\rho(x)\|_2^2 \rho(x)\,\dd x \\
&= 2\int_{\mathbb{R}^d} \left(\frac{1}{2}\|f_\theta(x)\|_2^2 + \operatorname{div}f_\theta(x) \right) \rho(x)\,\dd x + \int_{\mathbb{R}^d} \|\nabla\log\rho(x)\|_2^2 \,\rho(x)\dd x .
\end{align*}
By the definition of $\ell_\theta$ and $M_\rho(\theta)$, the first
integral on the right-hand side equals \(2M_\rho(\theta)\). Therefore,
\begin{align*}
R_\rho(\theta)=2M_\rho(\theta)+\int_{\mathbb{R}^d}\|\nabla\log\rho(x)\|_2^2\,\rho(x)\dd x,
\end{align*}
as claimed.

\subsection{Proof of \cref{lemma:ULLN}}
\label{proof:lemma:ULLN}

For $\theta \in \Theta_{n, \gamma}$, define  
\begin{align*}
	\bar M_{\rho}(\theta) = \E_{\rho} \big[ \ell_{\theta}(\cT_{R_n}(X)) \big].
\end{align*}
We first establish a global bound for $\ell_\theta$.
Let $B_1(u)=u_+-2(u-1)_++(u-2)_+$ be the cardinal linear B-spline.
Recall that $g_{\theta}$ is from \cref{eq:f-theta-g-theta}. 
The identity $B_2'(u)=B_1(u)-B_1(u-1)$ yields
\begin{align*}
\partial_a g_{\theta,a}(z)
&=\frac{1}{b_n}\sum_{\nu\in\mathbb Z^d}
(\theta_{a,\nu+e_a}-\theta_{a,\nu})
B_1(z_a/b_n-\nu_a-1)
\prod_{j\ne a}\psi_{\nu_j}(z_j).
\end{align*}
The coefficient constraints imply that for all $(a,\nu),(a,\nu+e_a) \in \cJ$ it holds that $|\chi_{R_n}(z)\partial_a g_{\theta,a}(z)|\leq Lb_n^{\gamma-1}$.
Moreover, $\|g_\theta(z)\|_2\leq\sqrt d\,\cR_n$ for every $z \in \R^d$.
Using these upper bounds, we see that
\begin{align*}
|\diver f_\theta(z)|
&=\big|\chi_{R_n}(z)\diver g_\theta(z)
       +\langle\nabla\chi_{R_n}(z),g_\theta(z)\rangle\big|\\
&\leq dLb_n^{\gamma-1}
       +\frac{C_\chi\sqrt d\,\cR_n}{R_n}, 
\end{align*}
where we recall $C_{\chi}$ is a positive numerical constant satisfying \cref{eq:chiR}. 
Thus, under the additional condition $R_n\cR_n\geq2C_\chi/\sqrt d$, we have
\begin{align*}
\|\ell_\theta\|_\infty \leq \frac{d\cR_n^2}{2}+dLb_n^{\gamma-1}
       +\frac{C_\chi\sqrt d\,\cR_n}{R_n}\leq d\cR_n^2+dLb_n^{\gamma-1}=Q_{n,\gamma}.
\end{align*}
Since $\cT_{R_n}(X)=X$ on $\Omega_{R_n}$ and $\cT_{R_n}(X)=0$ otherwise,
\begin{align*}
M_\rho(\theta)-\bar M_\rho(\theta)
&=\E_\rho\!\left[
  \big(\ell_\theta(X)-\ell_\theta(0)\big)
  \mathbbm{1}\{X\notin\Omega_{R_n}\}\right].
\end{align*}
It follows that
\begin{align}
\label{eq:ULLN1}
\sup_{\theta\in\Theta_{n,\gamma}}
|M_\rho(\theta)-\bar M_\rho(\theta)|
&\leq2Q_{n,\gamma}\P_\rho(X\notin\Omega_{R_n})
\leq2Q_{n,\gamma}n^{-10}. 
\end{align}
Next, we upper bound $\sup_{\theta \in \Theta_{n, \gamma}} |\bar M_{\rho}(\theta) - \hat M_\cD(\theta)|$. 
For this purpose, define
\begin{align*}
    \overline\ell_\theta(x)=\ell_\theta(\cT_{R_n}(x)), \qquad \mathcal L_n =\left\{
        \overline\ell_\theta:
        \theta\in\Theta_{n, \gamma}
    \right\}.
\end{align*}
By \cref{eq:bounded-ell}, we see that $\sup_{\theta \in \Theta_{n, \gamma}} \|\bar \ell_{\theta}\|_{\infty} \leq d \cR_n^2 + d Lb_n^{\gamma-1}$. 
We next bound the covering numbers of $\mathcal L_n$. 	
For every $z\in \R^d$ and $a\in[d]$, observe that 
\begin{align*}
\left| f_{\theta,a}(z)-f_{\theta',a}(z) \right| = \Big| \chi_{R_n}(z) \sum_{\nu\in\mathbb Z^d}\bigl(\theta_{a,\nu}-\theta'_{a,\nu}\bigr)\psi_\nu(z) \Big| \le \|\theta - \theta'\|_{\infty} \sum_{\nu\in\mathbb Z^d}\psi_\nu(z) = \|\theta - \theta'\|_{\infty}.
\end{align*}
Consequently, for all $z \in \R^d$ and $\theta, \theta' \in \Theta_{n, \gamma}$,  
\begin{align*}
    \|f_\theta(z)-f_{\theta'}(z)\|_2
    \le \sqrt d\,\|\theta - \theta'\|_{\infty}.
\end{align*}
Moreover, for $\theta \in \Theta_{n, \gamma}$, $a \in [d]$ and $z \in \R^d$, $|f_{\theta, a}(z)| \leq  |\sum_{\nu \in \Z^d} \theta_{a, \nu} \psi_{\nu}(z)| \leq \cR_n \sum_{\nu \in \Z^d} \psi_{\nu}(z) = \cR_n$. Therefore, $\|f_\theta(z)\|_2\vee\|f_{\theta'}(z)\|_2\le\sqrt d\,\cR_n$, which further implies that 
\begin{align*}
\frac12 \left| \|f_\theta(z)\|_2^2 - \|f_{\theta'}(z)\|_2^2 \right| &\le \frac12 \left( \|f_\theta(z)\|_2+ \|f_{\theta'}(z)\|_2 \right)
\|f_\theta(z)-f_{\theta'}(z)\|_2 \le d \cR_n \|\theta - \theta'\|_{\infty}.
\end{align*}
We next consider the divergence term. For $z\in\Omega_{R_n}$, we have $\chi_{R_n}(z)=1$, and hence
\begin{align*}
    \operatorname{div}f_\theta(z) = \sum_{a=1}^d\partial_a g_{\theta,a}(z).
\end{align*}
Applying Eq.~\eqref{eq:gcp-upper}, we see that 
\begin{align*}
	&\left| \partial_a g_{\theta,a}(z) - \partial_a g_{\theta',a}(z) \right| \\
	& =  \Big| \sum_{\nu \in \Z^d} (\theta_{a, \nu} - \theta'_{a, \nu})\, \psi_{\nu_a}'(z_a) \prod_{j \in [d] \backslash\{a\}} \psi_{\nu_j}(z_j) \Big| \\
	& \leq \sum_{\nu_1 \in \Z} \sum_{\nu_2 \in \Z} \ldots \sum_{\nu_{a - 1} \in \Z} \sum_{\nu_{a + 1} \in \Z} \ldots \sum_{\nu_d \in \Z} \prod_{j \in [d] \backslash\{a\}} \psi_{\nu_j}(z_j) \Big| \sum_{\nu_a \in \Z} (\theta_{a, \nu} - \theta'_{a, \nu}) \psi_{\nu_a}'(z_a) \Big| \\
	& \leq \,\frac{1}{b_n} \sup_{\nu\in\mathbb Z^d} \left| \bigl(\theta_{a,\nu+e_a}-\theta'_{a,\nu+e_a}\bigr)-\bigl(\theta_{a,\nu}-\theta'_{a,\nu}\bigr) \right| \cdot \sum_{\nu_1 \in \Z} \sum_{\nu_2 \in \Z} \ldots \sum_{\nu_{a - 1} \in \Z} \sum_{\nu_{a + 1} \in \Z} \ldots \sum_{\nu_d \in \Z} \prod_{j \in [d] \backslash\{a\}} \psi_{\nu_j}(z_j) \\
		& = \,\frac{1}{b_n} \sup_{\nu\in\mathbb Z^d} \left| \bigl(\theta_{a,\nu+e_a}-\theta'_{a,\nu+e_a}\bigr)-\bigl(\theta_{a,\nu}-\theta'_{a,\nu}\bigr) \right|. 
\end{align*}
Summing over $a \in [d]$, we conclude that for all $z \in \Omega_{R_n}$, 
\begin{align*}
\left|\operatorname{div}f_\theta(z) - \operatorname{div}f_{\theta'}(z) \right| \le \frac{2d}{b_n} \|\theta - \theta'\|_{\infty}.
\end{align*}
Combining the quadratic and divergence terms, we see that for all $\theta, \theta' \in \Theta_{n, \gamma}$
\begin{align}
\label{eq:bar-ell-Lipschitz}
    \|\overline\ell_\theta
      -\overline\ell_{\theta'}\|_\infty
    \le
    \Big( d \cR_n + \frac{2d}{b_n} \Big)\cdot \|\theta-\theta'\|_\infty.
\end{align}
Since $\Theta_{n, \gamma}\subseteq[-\cR_n,\cR_n]^{|\cJ|}$, we then see that for every $\epsilon>0$,
\begin{align*}
    \cN\left(\Theta_{n, \gamma}, \|\cdot\|_\infty, \epsilon \right) \le \left(1+\frac{2\cR_n }{\epsilon} \right)^{|\cJ|}.
\end{align*}
For $X_1, X_2, \ldots, X_n \sim_{i.i.d.} \rho$, let $P_n$ denote the empirical distribution that assigns equal mass to each of the $n$ data points. 
Observe that for $\theta, \theta' \in \Theta_{n, \gamma}$, 
\begin{align*}
	\sqrt{\frac{1}{n} \sum_{i = 1}^n \big( \bar \ell_{\theta}(X_i) - \bar \ell_{\theta'}(X_i) \big)^2}  \leq \Big( d \cR_n + \frac{2d}{b_n} \Big)\cdot \|\theta-\theta'\|_\infty.
\end{align*}
As a consequence, 
\begin{align*}
	\mathcal{N}\left( \cL_n, L_2(P_n), \epsilon \right) \leq \cN \Big( \Theta_{n, \gamma},\, \|\cdot\|_{\infty},\, \big( d \cR_n + \frac{2d}{b_n} \big)^{-1} \epsilon  \Big) \le \left(1+\frac{2\cR_n (d \cR_n + 2d b_n^{-1}) }{\epsilon} \right)^{|\cJ|}. 
\end{align*}
Recall that by \cref{eq:bounded-ell}, we have $\sup_{\theta \in \Theta_{n, \gamma}} \|\bar \ell_{\theta}\|_{\infty} \leq d \cR_n^2 + d Lb_n^{\gamma-1}$.
Therefore, for all $\theta, \theta' \in \Theta_{n, \gamma}$, 
\begin{align*}
	\sqrt{\frac{1}{n} \sum_{i = 1}^n \Big(\bar \ell_{\theta}(X_i) - \bar\ell_{\theta'}(X_i) \Big)^2 } \leq 2 \big( d \cR_n^2 + d Lb_n^{\gamma-1} \big), 
\end{align*} 
which further implies that $\cN \big( \cL_n, L_2(P_n), \epsilon \big) = 1$ for all $\epsilon \geq 2( d \cR_n^2 + d L b_n^{\gamma-1})$. 
Invoking the maximal inequality (see Chapter 8 of \cite{vershynin2019high} for more details), we obtain the following upper bound: 
\begin{align*}
	\E_{\cD} \left[ \sup_{\theta \in \Theta_{n, \gamma}} \big| \bar M_{\rho}(\theta) - \hat M_\cD(\theta) \big| \right] \leq & \,\frac{C}{\sqrt{n}} \cdot \E \left[ \int_0^{2( d \cR_n^2 + d L b_n^{\gamma-1})} \sqrt{\log \cN \big( \cL_n, L_2(P_n), \epsilon \big)} \,\dd \epsilon \right] \\
	\leq & \, \frac{C}{\sqrt{n}} \int_0^{2( d \cR_n^2 + d L b_n^{\gamma-1})} \sqrt{|\cJ| \cdot \log \Big( 1 +  \frac{2\cR_n (d \cR_n + 2d b_n^{-1}) }{\epsilon}\Big)}\, \dd \epsilon, 
\end{align*}
%
where $C$ is a positive numerical constant. Observe that 
\begin{align*}
	\log \Big( 1 +  \frac{2\cR_n (d \cR_n + 2d b_n^{-1}) }{\epsilon}\Big) \leq \log \Big( 1 + 2\cR_n(d \cR_n + 2d b_n^{-1}) \Big) + \log \big( 1 + \epsilon^{-1} \big). 
\end{align*}
Substituting the above bound into the preceding inequality yields
\begin{align}
\label{eq:ULLN2}
\begin{split}
	& \E_{\cD} \left[ \sup_{\theta \in \Theta_{n, \gamma}} \big| \bar M_{\rho}(\theta) - \hat M_\cD(\theta) \big| \right] \\
	& \leq 2C( d \cR_n^2 + d L b_n^{\gamma-1})\sqrt{\frac{|\cJ| }{n} \cdot \log \Big( 1 + 2\cR_n(d \cR_n + 2d b_n^{-1}) \Big) } + 4C \sqrt{\frac{|\cJ| (d \cR_n^2 + d Lb_n^{\gamma-1})}{n}}.
\end{split} 
\end{align}
Combining \cref{eq:ULLN1,eq:ULLN2} completes the proof of this lemma. 

\subsection{Proof of \cref{lemma:exponential-mechanism-risk}}
\label{proof:lemma:exponential-mechanism-risk}

Denote by $\mu$ the uniform distribution over $\Theta_{n, \gamma}$. By the Gibbs variational formula, we see that 
\begin{align*}
	- \log \E_{\mu} \big[ e^{-\frac{n \alpha}{4Q_{n, \gamma}} \hat M_\cD(\theta) } \big] = \inf_{q \ll \mu} \left( \frac{n \alpha}{4Q_{n, \gamma}}\, \E_{q} \big[ \hat M_\cD(\theta) \big] + \KL\big( q \parallel \mu \big) \right), 
\end{align*} 
which further implies that for every probability measure $q \ll \mu$, 
\begin{align}
\label{eq:313-1}
	\E_{\pi_\cD} \big[ \hat M_\cD(\theta) \big] + \frac{4Q_{n, \gamma}}{n \alpha}\,\KL\big( \pi_\cD \parallel \mu \big) \leq   \E_q \big[ \hat M_\cD(\theta) \big] + \frac{4Q_{n, \gamma}}{n \alpha}\, \KL \big( q \parallel \mu \big). 
\end{align}
Fix $0<\zeta<\cR_n$, set $\lambda=\zeta/(2\cR_n)$, and define
\begin{align*}
K_\lambda=(1-\lambda)\theta^\ast+\lambda \Theta_{n, \gamma},
\qquad q=\operatorname{Unif}(K_\lambda).
\end{align*}
Because $\Theta_{n, \gamma}$ is convex, $\theta^\ast\in\Theta_{n, \gamma}$, and $0<\lambda<1$, we have $K_\lambda\subseteq\Theta_{n, \gamma}$, and hence $q\ll\mu$.
For every $\theta\in K_\lambda$, there is $u\in\Theta_{n, \gamma}$ such that
$\theta=(1-\lambda)\theta^\ast+\lambda u$, so
\begin{align*}
\|\theta-\theta^\ast\|_\infty
=\lambda\|u-\theta^\ast\|_\infty
\leq2\lambda\cR_n=\zeta.
\end{align*}
The Lipschitz bound in \cref{eq:bar-ell-Lipschitz} therefore yields
\begin{align*}
\E_q[\hat M_\cD(\theta)]
\leq\hat M_\cD(\theta^\ast)
 +\left(d\cR_n+\frac{2d}{b_n}\right)\zeta.
\end{align*}
Let $m$ denote Lebesgue measure on $\mathbb R^{|\cJ|}$.
Note that $m(K_\lambda)=\lambda^{|\cJ|} m(\Theta_{n, \gamma})$.
Since both $q$ and $\mu$ are uniform on their respective supports,
\begin{align*}
\KL(q\parallel\mu) =\log\frac{m(\Theta_{n, \gamma})}{m(K_\lambda)} =|\cJ|\log\frac1\lambda =|\cJ|\log\frac{2\cR_n}{\zeta}.
\end{align*}
Combining these inequalities and using
$\KL(\pi_\cD\parallel\mu)\geq0$, we obtain
\begin{align*}
\E_{\pi_\cD}[\hat M_\cD(\theta)] \leq \hat M_\cD(\theta^\ast)
 +\left(d\cR_n+\frac{2d}{b_n}\right)\zeta+\frac{4Q_{n,\gamma}|\cJ|}{n\alpha}
 \log\frac{2\cR_n}{\zeta}.
\end{align*}
Taking $\zeta=n^{-10}$ proves the desired statement.

\subsection{Proof of \cref{thm:DP-score-upper-bound}}
\label{proof:thm:DP-score-upper-bound}

We assume $n$ is large enough such that $b_n \leq (3\sqrt{d})^{-1}$. 
We denote by $\hat \pi_{\cD, \cL}$ the output distribution of \cref{alg:projected_LMC} under the above parameter choices.
Namely, conditional on $\cD$, we have $\hat\theta_{\mathrm{LMC}} \sim \hat \pi_{\cD, \cL}$. 
By \cref{thm:sampling}, we know that $\TV(\tilde \pi_\cD, \hat\pi_{\cD, \cL}) \leq \min(r_n, \Upsilon_{\alpha, \beta} / 2)$, which, together with \cref{lemma:ridge-regularized-distribution}, implies that 
\begin{align*}
	\TV(\pi_\cD, \hat \pi_{\cD, \cL}) \leq \TV(\pi_\cD, \tilde\pi_\cD) + \TV(\tilde \pi_\cD, \hat\pi_{\cD, \cL}) \leq \min(r_n, \Upsilon_{\alpha, \beta} / 2) + e^{\min (r_n, 2\cR_n^{-2}\log(1 + \Upsilon_{\alpha, \beta} / 2) ) \cR_n^2 / 2} - 1. 
\end{align*} 
The above upper bound yields two separate upper bounds for $\TV(\pi_\cD, \hat \pi_{\cD, \cL})$: (1) $\TV(\pi_\cD, \hat \pi_{\cD, \cL}) \leq \Upsilon_{\alpha, \beta}$; (2) $\TV(\pi_\cD, \hat \pi_{\cD, \cL}) \leq r_n + e^{r_n \cR_n^2 / 2} - 1$. 
By \cref{lem:privacy-approximate-sampling}, the mechanism $\cD \mapsto \hat\theta_{\mathrm{LMC}} \sim \hat \pi_{\cD, \cL}$ is $(\alpha, (1 + e^{\alpha}) \Upsilon_{\alpha, \beta})$-differentially private, or equivalently $(\alpha, \beta)$-differentially private. 
 
By \cref{lemma:approximate-sampling-preserves-risk}, we have  
\begin{align*}
	\E\big[ R\big(f_{\hat\theta_{\mathrm{LMC}}},\, \rho_0 \big) \big] = & \E_{\cD} \big[\, \E_{\hat \theta \sim \hat\pi_{\cD, \cL}}  \big[ R\big(f_{\hat\theta}, \, \rho_0\big) \big] \,\big] \\
	\leq & \E_{\cD} \big[\, \E_{\hat \theta \sim \pi_\cD}  \big[ R\big(f_{\hat\theta}, \, \rho_0\big) \big] \,\big] +  2\big( r_n + e^{r_n \cR_n^2 / 2} - 1 \big)\, \big( d \cR_n^2 +  C^{1/2} L^2 (1+ \xi^2 d) \big). 
\end{align*}
Recall that $\theta^{\ast} \in \R^{|\cJ|}$ is defined in \cref{eq:theta-ast}. 
Note that by \cref{lemma:M-R}, 
\begin{align*}
	& \E_{\cD} \big[\, \E_{\hat \theta \sim \pi_\cD}  \big[ R\big(f_{\hat\theta}, \, \rho_0\big) \big] \,\big] \\
	& = \E_{\cD} \big[\, \E_{\hat \theta \sim \pi_\cD}  \big[ R_{\rho_0}(\hat\theta \,) - R_{\rho_0}(\theta^{\ast}) \big] \,\big] + R_{\rho_0}(\theta^{\ast}) \\
	& = 2\E_{\cD} \big[\, \E_{\hat \theta \sim \pi_\cD}  \big[ M_{\rho_0}(\hat\theta \,) - M_{\rho_0}(\theta^{\ast}) \big] \,\big] + R_{\rho_0}(\theta^{\ast}) \\
	& \leq  \underbrace{4\E_{\cD, \, \hat\theta \sim \pi_\cD} \big[\, \sup_{\theta \in \Theta_{n, \gamma} } \big|\hat M_\cD(\theta) - M_{\rho_0}(\theta) \big| \big]}_{\mathrm{I}} + \underbrace{ 2\E_{\cD, \, \hat\theta \sim \pi_\cD} \big[\, \hat M_\cD(\hat\theta) - \hat M_\cD(\theta^{\ast})  \big]}_{\mathrm{II}} + R_{\rho_0}(\theta^{\ast}). 
\end{align*}
By \cref{lemma:ULLN}, we see that term $\mathrm{I}$ is bounded by four times the upper bound established therein.
By \cref{lemma:exponential-mechanism-risk}, we know that term $\mathrm{II}$ is upper bounded by two times the upper bound from the lemma. The last term $R_{\rho_0}(\theta^{\ast})$ is upper bounded in \cref{lem:local-holder-approximation}. 
Combining these bounds yields the desired result.

\subsection{Proof of \cref{thm:score-holder-lower}}
\label{proof:thm:score-holder-lower}

Our proof of the private rate utilizes a differentially private version of Assouad's Lemma \citep{acharya2021differentially}. 
Specifically, we first construct a family of distributions indexed by a hypercube. 
We then show that distributions with close indices are hard to distinguish, while the corresponding score estimation error remains lower bounded. 
We establish the private and non-private lower bounds separately, then combine them to complete the proof. 

\paragraph{Distribution construction.}
We set the base distribution as $\normal (0, \sigma^2 I_d)$, 
where $\sigma^2$ is a positive number that we will specify later. 
The probability density function and the score of this base distribution are listed below: 
\[
    \rho_\sigma(x)
    =
    \frac{1}{(2\pi\sigma^2)^{d/2}}
    \exp\left(-\frac{\|x\|_2^2}{2\sigma^2}\right),
    \qquad
    s_\sigma(x)
    =
    -\frac{x}{\sigma^2}.
\]
We then construct perturbations of this base distribution to form a distribution class. Specifically, define a bump function
\[
    \psi(t)
    =
    \begin{cases}
        \displaystyle
        \exp\left(
            1-\frac{1}{1-t^2}
        \right),
        & |t|<1,\\[1ex]
        0,
        & |t|\ge 1.
    \end{cases}
\]
Then $\psi\in C_c^\infty(\mathbb{R})$, $0\le\psi\le1$,
$\psi(0)=1$ and $\psi(t)=0$ for $|t|\geq 1$. Define
\begin{equation*}
    g(u)=\prod_{k=1}^d
    \psi(4u_k-2),
    \qquad
    u=(u_1,\ldots,u_d)\in\mathbb R^d. 
\end{equation*}
Note that $0\leq g(x) \leq 1$ for all $x \in \R^d$, $g(1/2, \dots, 1/2) =1$ and $g(x) =0$ whenever $x \notin [1/4, 3/4]^d$.

Let $m\ge2$ be an integer to be specified later, and set $h = m^{-1}$. 
We define an index set $\cI_m$ and denote its size by $K_m$: 
\[
    \mathcal I_m=\{0,\ldots,m-1\}^d,
    \qquad
    K_m=|\mathcal I_m|=m^d.
\]
For $ i=(i_1,\ldots,i_d)\in\mathcal I_m$, set $x_{i} = {i} / m$,  and for $\vartheta=(\vartheta_{i}:i\in\mathcal \cI_m)\in\{0,1\}^{K_m}$,
define
\begin{equation*}
    V_\vartheta(x)
    =
    \omega
    \sum_{i\in\mathcal I_m}\vartheta_i g\left(\frac{x-x_i}{h}\right), 
\end{equation*}
where $\omega>0$ is a sufficiently small constant that we will also specify later. 
For $\vartheta \in \{0, 1\}^{K_m}$, we define an associated probability distribution $\rho_{\vartheta}$: 
\begin{equation}
\label{eq:density-construction}
    \rho_\vartheta(x)
    =
    \frac{\rho_\sigma(x)e^{V_\vartheta(x)}}{Z_\vartheta},
    \qquad
    Z_\vartheta
    =
    \int_{\mathbb R^d}\rho_\sigma (x)e^{V_\vartheta(x)}\,\dd x.
\end{equation}
Observe that $\|V_\vartheta\|_\infty \leq \omega$, hence
\begin{equation}
\label{eq:likelihood-ratio-uniform}
    e^{-2\omega}
    \le
    \frac{\rho_\vartheta(x)}{\rho_\sigma(x)}
    \le
    e^{2\omega}
    \qquad
    \text{for all }x\in\mathbb R^d.
\end{equation}
As a consequence, $\rho_{\vartheta}$ is $e^{\omega}\sigma$-sub-Gaussian for all $\vartheta \in \{0, 1\}^{K_m}$. 
Therefore, to ensure $\rho_{\vartheta}$ is $\xi$-sub-Gaussian, it suffices to ensure 
\begin{align}
\label{eq:condition1}
	e^{\omega} \sigma \leq \xi. 
\end{align}
We next show that with a sufficiently small $\omega$, $\rho_{\vartheta}$ has a local $(L, \gamma)$-H\"older smooth score function for all $\vartheta \in \{0, 1\}^{K_m}$.
Observe that for all ${\vartheta} \in \{0, 1\}^{K_m}$, the score of $\rho_{\vartheta}$ can be written as 
\begin{align*}
    s_\vartheta(x)
    &=
    \nabla\log\rho_\vartheta(x) =
    s_\sigma(x)
    +
    \frac{\omega}{h}
    \sum_{i\in\mathcal I_m}
    \vartheta_i
    \nabla g\left(\frac{x-x_i}{h}\right). 
\end{align*}
Observe that the functions $x\mapsto \nabla g(h^{-1}(x-x_i))$ have disjoint supports for different $i \in \cI_m$. 
Therefore, there exists a positive numerical constant $C_0$, such that for all $x, y \in \R^d$ with $\|x - y\|_2 \leq 1$, 
\begin{align*}
	& \|s_{\vartheta}(x) - s_{\vartheta}(y)\|_2 \leq \left( \frac{1}{\sigma^2} + \frac{C_0 \omega d}{h^2} \right) \|x - y\|_2 \leq \left( \frac{1}{\sigma^2} + d^{\frac{1 - \gamma}{2}} h^{1 - \gamma} \cdot \frac{C_0 \omega d}{h^2} \right) \|x - y\|_2^{\gamma},  & \mbox{if }\|x - y\|_2 \leq \sqrt{d}h, \\
	& \|s_{\vartheta}(x) - s_{\vartheta}(y)\|_2 \leq \frac{\|x - y\|_2}{\sigma^2} + \frac{C_0 \omega \sqrt{d}}{h} \leq \left( \frac{1}{\sigma^2} + \frac{C_0 \omega d^{\frac{1 - \gamma }{2}}}{h^{1 + \gamma}} \right) \cdot \|x - y\|_2^{\gamma }, & \mbox{if }\|x - y\|_2 > \sqrt{d}h. 
\end{align*}
Hence, to ensure $s_{\vartheta}$ is local $(L, \gamma)$-H\"older  for all $\vartheta \in \{0, 1\}^{K_m}$, it suffices to choose $(\sigma^2, \omega)$ such that 
\begin{align}
\label{eq:condition2}
	\frac{1}{\sigma^2} + \frac{C_0 \omega d^{\frac{3 - \gamma }{2}}}{h^{1 + \gamma}} \leq L. 
\end{align}
To satisfy requirements \eqref{eq:condition1} and \eqref{eq:condition2}, we set 
\begin{align}
\label{eq:sigma-omega}
	\sigma = \sqrt{2 L^{-1}}, \qquad \omega = \min \left\{ \frac{1}{2} \log \frac{\xi^2L}{2},\,\, \, \frac{Lh^{1  + \gamma }}{2C_0 d^{\frac{3 - \gamma}{2}}}, \, \frac{1}{4} \right\}.  
\end{align}
When $\xi^2 L > 2$ we know that $\omega > 0$. 
In addition, conditions \eqref{eq:condition1} and \eqref{eq:condition2} are both satisfied with the above choice of $(\sigma, \omega)$. 

Finally, we show that $\rho_{\vartheta} \in \cP_{\xi, L, \gamma, M}$. 
To this end, it suffices to investigate the mean vector of $\rho_{\vartheta}$. 
Invoking the Cauchy-Schwarz inequality and \cref{eq:likelihood-ratio-uniform}, we see that 
\begin{align*}
	\left\| \int x \rho_{\vartheta}(x) \dd x \right\|_2 \leq \left( \int \|x\|_2^2 \rho_{\vartheta}(x) \dd x \right)^{1/2} \leq e^{\omega }\left( \int x^2 \rho_{\sigma}(x) \dd x \right)^{1/2} = \sqrt{d}e^{\omega} \sigma \leq \sqrt{d} \xi. 
\end{align*}
Therefore, $\rho_{\vartheta} \in \cP_{\xi, L, \gamma, M}$ whenever $M \geq \sqrt{d} \xi$.

\paragraph{Non-private lower bound.} 
We first establish a non-private lower bound that applies to all score estimators, without imposing differential privacy.
%

By the Gilbert-Varshamov bound \citep[Lemma 2.9]{tsybakov2009introduction}, there exists $\cW \subseteq \{0, 1\}^{K_m}$ and a positive numerical constant $c_0$, such that $|\cW| \geq \exp(c_0 K_m)$, and for all $\vartheta, \vartheta' \in \cW$ with $\vartheta \neq \vartheta'$, we have $\Ham(\vartheta, \vartheta') \geq c_0 K_m$. 
Therefore, for all $\vartheta, \vartheta' \in \cW$ with $\vartheta \neq \vartheta'$,
\begin{align*}
	\int \|s_\vartheta(x) - s_{\vartheta'}(x)\|_2^2\, \rho_{\vartheta}(x) \,\dd x \geq c_0\,  C_{L, \xi, d, \gamma}'h^{2 \gamma }. 
\end{align*}
We next upper bound the KL-divergence between $\rho_{\vartheta}$ and $\rho_\sigma$ for $\vartheta \in \cW$. 
Specifically, for $\vartheta \in \cW$, by \cref{eq:likelihood-ratio-uniform} 
\begin{align*}
    \KL(\rho_{\vartheta} \parallel \rho_{\sigma}) \leq  \chi^2 (\rho_{\vartheta} \parallel \rho_{\sigma}) \leq (e^{2\omega} - 1 )^2 \leq 16 \omega^2 \leq \frac{4 L^2 h^{2 + 2 \gamma}}{C_0^2 d^{3 - \gamma}}. 
\end{align*}
As a consequence, for all $\vartheta \in \cW$, 
\begin{equation}
\KL\left(\rho_\vartheta^{\otimes n} \parallel   \rho_{\sigma}^{\otimes n} \right) = n\,\KL(\rho_\vartheta \parallel \rho_{\sigma}) \leq \frac{4 n L^2 h^{2 + 2 \gamma}}{C_0^2 d^{3 - \gamma}},   
\label{eq:n-sample-kl}
\end{equation}
which further implies that 
\begin{equation*}
    \frac{1}{|\cW|}
    \sum_{\vartheta \in \cW}
    \KL 
    \left(
        \rho_\vartheta^{\otimes n}
        \parallel 
        \rho_\sigma^{\otimes n}
    \right) \leq  \frac{4 n L^2 h^{2 + 2 \gamma}}{C_0^2 d^{3 - \gamma}}. 
\end{equation*}
For an arbitrary estimator $\hat s$, define
\begin{align*}
\hat\vartheta_{\cW}
\in\operatorname*{arg\,min}_{\vartheta\in\cW}\int_{[0, 1]^d}\|\hat s(x)-s_\vartheta(x)\|_2^2\, \dd x,
\end{align*}
with a fixed and deterministic rule for breaking ties.
For every $\vartheta\in\cW$, the minimizing property and the triangle inequality give
\begin{align*}
\int_{[0, 1]^d} \|s_{\hat\vartheta_{\cW}}(x)-s_\vartheta(x)\|_2^2 \,\dd x
&\leq 2\int_{[0, 1]^d} \|s_{\hat\vartheta_{\cW}}(x)-\hat s(x)\|_2^2 \, \dd x+ 2\int_{[0, 1]^d} \|\hat s(x)-s_\vartheta(x)\|_2^2 \, \dd x \\
&\leq 4 \int_{[0, 1]^d} \|\hat s(x)-s_\vartheta(x)\|_2^2 \, \dd x.
\end{align*}
Set
\begin{align*}
	c_{\ast} = e^{-1/4}(2\pi\sigma^2)^{-d/2}
\exp\left(-\frac{d}{2\sigma^2}\right). 
\end{align*}
Since $0\leq V_\vartheta\leq\omega\leq1/4$ and $1\leq Z_\vartheta\leq e^\omega$, we have $\rho_\vartheta(x)\geq c_*$ for every $x\in[0,1]^d$.
Therefore,
\begin{align*}
R(\hat s,\rho_\vartheta)\geq c_*\int_{[0,1]^d}\|\hat s(x)-s_\vartheta(x)\|_2^2\,\dd x\geq \frac{c_*}{4}\int_{[0,1]^d}
\|s_{\hat\vartheta_{\cW}}(x)-s_\vartheta(x)\|_2^2\,\dd x.
\end{align*}
For sufficiently small $h$ (equivalently, sufficiently large $m$), the chosen amplitude satisfies $\omega=Lh^{1+\gamma}/(2C_0d^{(3-\gamma)/2})$.
Using the disjoint supports of the bump gradients and the packing separation, we obtain
\begin{align}
\label{eq:R-separation}
	R(\hat s,\rho_\vartheta) \geq \frac{c_*}{4}\omega^2h^{d-2} \|\nabla g\|_{L^2(\mathbb R^d)}^2 \Ham(\hat\vartheta_{\cW},\vartheta) \geq \frac{c_0c_{\ast} L^2 h^{2\gamma}\|\nabla g\|_{L^2(\mathbb R^d)}^2}{16 C_0^2 d^{3 - \gamma}} \mathbbm{1}\{\hat\vartheta_{\cW}\neq\vartheta\}. 
\end{align}
Taking expectations and averaging over the packing yields
\begin{align*}
\sup_{\rho\in\cP_{\xi,L,\gamma,M}}\E_{\rho^{\otimes n}}[R(\hat s,\rho)] \geq  \frac{c_0c_{\ast} L^2 h^{2\gamma}\|\nabla g\|_{L^2(\mathbb R^d)}^2}{16|\cW| C_0^2 d^{3 - \gamma}}\sum_{\vartheta\in\cW}\P_{\rho_\vartheta^{\otimes n}} \bigl(\hat\vartheta_{\cW}\neq\vartheta\bigr).
\end{align*}
Fano's inequality applies to the average error probability on the right.
Together with the preceding KL upper bound and $\log|\cW|\geq c_0K_m$, it gives the following risk lower bound after taking the infimum over $\hat s \in \cM_{\alpha, \beta}$: 
\begin{align*}
	\inf_{\widehat s \in \cM_{\alpha, \beta}} \sup_{\rho\in\mathcal P_{\xi, L, \gamma, M}} \mathbb E_{\rho^{\otimes n}} \left[ R(\hat s, \rho)\right] \geq & \,\inf_{\widehat s}   \sup_{\rho\in\mathcal P_{\xi, L, \gamma, M}} \mathbb E_{ \rho^{\otimes n}} \left[R(\hat s, \rho)\right] \\
	\geq & \,   C_{L, \xi, d, \gamma}'h^{2 \gamma }  \left( 1 - \frac{1}{c_0 K_m} \cdot \Big( \frac{4 n L^2 h^{2 + 2 \gamma}}{C_0^2 d^{3 - \gamma}} + \log 2  \Big) \right), 
\end{align*}
where $C_{L, \xi, d, \gamma}' > 0$ depends only on $(L, \xi, d, \gamma)$.
Setting $h_0 = n^{-1/(2\gamma + d + 2)} (8L^2 C_0^{-2} c_0^{-1} d^{-3 + \gamma})^{-1 / (2 + 2 \gamma + d)}$, and $h = 1/\lceil h_0^{-1} \rceil$, we see that for a sufficiently large $n$,  
\begin{align*}
    \inf_{\widehat s \in \cM_{\alpha, \beta}}\sup_{\rho\in\mathcal P_{\xi, L, \gamma, M}}\mathbb E_{D \sim \rho^{\otimes n}}\left[ R(\hat s, \rho)\right] \geq &\, C_{L, \xi, d, \gamma} n^{-2 \gamma / (2\gamma + d + 2)}, 
\end{align*}
where $ C_{L, \xi, d, \gamma} > 0$ depends only on $(L, \xi, d, \gamma)$. This proves the non-private lower bound.

\paragraph{Private lower bound.} For $i \in [K_m]$, we define the following mixture distributions: 
\begin{align*}
\label{eq:assouad-mixture}
    P_{+i} =  \frac{1}{2^{K_m-1}} \sum_{\vartheta \in \{0, 1\}^{K_m}: \vartheta_i = 1} \rho_\vartheta^{\otimes n}, \qquad  \quad P_{-i} =  \frac{1}{2^{K_m-1}} \sum_{\vartheta \in \{0, 1\}^{K_m}: \vartheta_i = 0} \rho_\vartheta^{\otimes n}.
\end{align*}
Consider $\vartheta, \vartheta' \in \{0, 1\}^{K_m}$ that differ only in the
$i$-th coordinate. Without loss of generality, suppose that
$\vartheta_i=1$ and $\vartheta_i'=0$. Then
\begin{equation*}
    V_\vartheta(x)-V_{\vartheta'}(x)
    =
    \omega g_i(x), \quad g_i(x) = 
    g\left(\frac{x-x_i}{h}\right), \qquad \frac{Z_\vartheta}{Z_{\vartheta'}}
    =
    \mathbb E_{\rho_{\vartheta'}}
    \left[e^{\omega  g_i(X)}\right].
\end{equation*}
Observe that
\begin{align*}
    \TV(\rho_\vartheta,\rho_{\vartheta'})= \frac{1}{2}  \int\rho_{\vartheta'}(x) \left| \frac{e^{\omega g_i(x)}}{\mathbb E_{\rho_{\vartheta'}} [e^{\omega  g_i(X)}]}-1\right| \dd  x  \leq (e^\omega-1)\P_{\rho_{\vartheta'}}(X\in\operatorname{supp}g_i) \leq C_{L, d} \,\omega  h^d,  
\end{align*}
where $C_{L, d} > 0$ depends only on $(L, d)$. 
Define $\nu$ to be the $n$-fold product of a maximal coupling between $\rho_\vartheta$ and $\rho_{\vartheta'}$, so that for $(X, Y)$ drawn from $\nu$, we have
\begin{equation*}
\mathbb{E}_\nu[\Ham (X, Y)] \leq D = n C_{L, d} \,\omega  h^d, 
\end{equation*}
which further implies by averaging over $\vartheta$ and $\vartheta'$ that there exists a coupling $(X, Y)$ between $P_{+i}$ and $P_{-i}$ with $\E[\Ham (X, Y)] \leq D $. 

Given a score estimator $\hat s$, define
\begin{equation*}
\widehat\vartheta
\in \operatorname*{arg\,min}_{\vartheta\in\mathcal \{0, 1\}^{K_m}}
\int_{[0,1]^d}\|\widehat s(x)-s_\vartheta(x)\|_2^2\,\dd x,
\end{equation*}
For every $\vartheta\in \{0, 1\}^{K_m}$, the definition of $\widehat\vartheta$ gives $\int_{[0, 1]^d} \|s_{\widehat\vartheta}(x)-\widehat s(x)\|_2^2 \dd x \le \int_{[0, 1]^d}\|s_\vartheta(x) - \widehat s(x)\|_{2}^2 \dd x$.
The triangle inequality therefore yields
$\int_{[0, 1]^d}\|s_{\widehat\vartheta}(x)-s_\vartheta(x)\|_{2}^2 \dd x
\le 4 \int_{[0, 1]^d}\|\widehat s(x)-s_\vartheta(x)\|_{2}^2 \dd x$.
Since $\rho_\vartheta(x)\ge c_*$ on $[0,1]^d$, we see that $R(\widehat s,\rho_\vartheta)
\ge c_* \int_{[0, 1]^d} \|\widehat s(x)-s_\vartheta(x)\|_{2}^2 \dd x
\ge \frac{c_*}{4} \int_{[0, 1]^d}\|s_{\widehat\vartheta}(x) -s_\vartheta(x)\|_{2}^2 \dd x$. 
To evaluate the last norm, observe that
\begin{equation*}
s_{\widehat\vartheta}(x)-s_\vartheta(x)
=\frac{\omega}{h}\sum_{i\in \mathcal I_m}
(\widehat\vartheta_i-\vartheta_i)
\nabla g\!\left(\frac{x-x_i}{h}\right).
\end{equation*}
Similar to the derivation of \cref{eq:R-separation}, we know that 
\begin{align*}
R(\widehat s,\rho_\vartheta)
\ge \frac{c_*}{4}\,\omega^2h^{d-2}
\|\nabla g\|_{L^2(\mathbb R^d)}^2
\operatorname{Ham}(\widehat\vartheta,\vartheta).
\end{align*}
For sufficiently small $h$, \cref{eq:sigma-omega} gives \(\omega=Lh^{1+\gamma}/(2C_0d^{(3-\gamma)/2})\). Substituting this into the above lower bound yields
\begin{align*}
R(\hat s,\rho_\vartheta)
\geq C_{L, \xi, d, \gamma}'' h^{d+2\gamma}\Ham(\hat\vartheta,\vartheta), 
\end{align*}
where $C_{L, \xi, d, \gamma}'' > 0$ depends only on $(L, \xi, d, \gamma)$. 
Apply Theorem 3 of \cite{acharya2021differentially} to the Hamming loss and the mixture distributions defined above.
Since $K_m=h^{-d}$, the coupling bound yields
\begin{align*}
\inf_{\hat s\in\cM_{\alpha,\beta}}
\sup_{\rho\in\cP_{\xi,L,\gamma,M}}
\E_{\rho^{\otimes n}}[R(\hat s,\rho)]\geq\frac{C_{L, \xi, d, \gamma}'' h^{2\gamma}}{4}
\bigl(0.9e^{-10\alpha D}-10\beta D\bigr),
\qquad D=C_{L,d}n\omega h^d.
\end{align*}
Setting $h = c(n \alpha)^{-1 / (\gamma + d + 1)}$ for a sufficiently small $c$ that also ensures $1 / h$ is an integer, we see that there exists $C_{L, \xi, d, \gamma} > 0$ that depends only on $(L, \xi, d, \gamma)$, such that 
\begin{align*}
	\inf_{\widehat s \in \cM_{\alpha, \beta}} \sup_{\rho\in\mathcal P_{\xi, L, \gamma, M}} \mathbb E_{ \rho^{\otimes n}}\left[  R(\hat s, \rho)\right]  \geq C_{L, \xi, d, \gamma} (n \alpha)^{-2 \gamma / (\gamma + d + 1)}. 
\end{align*}
The proof is done.

\section{Proofs for differentially private diffusion model}

\subsection{Proof of \cref{lem:private-histogram}}
\label{proof:lem:private-histogram}

	For neighboring datasets $\cD$ and $\cD'$, the empirical cell-probability
	vectors satisfy $\|\bar p_\cD^\hist-\bar p_{\cD'}^\hist\|_1\leq2/n$.  Therefore, adding
	independent Laplace noise with scale $4/(n\alpha)$ to every coordinate is
	$(\alpha/2,0)$-differentially private.  Projection onto $\cS_{\cH}$ and
	the subsequent construction of $\hat\rho_D^\hist$ are post-processing operations that preserves privacy.

\subsection{Proof of Theorem \ref{thm:ddpm-upper}}
\label{proof:thm:ddpm-upper}

To prove \cref{thm:ddpm-upper}, we prove a more general theorem (\cref{thm:private-ddpm-tv}) that applies to all discretization schemes that satisfy the assumptions of \cite{benton2024nearly}, which is stated as Assumption \ref{ass:time-grid} below.
 
\begin{ass}
\label{ass:time-grid}
	There exists $\kappa \in (0, 1)$, such that 
\begin{align*}
\Delta_k \leq \kappa\min\{1,T-t_{k+1}\},\qquad k=0,\ldots,N-1.
\end{align*}
\end{ass}

The following lemma establishes the existence of a discretization scheme satisfying Assumption \ref{ass:time-grid}.

\begin{lem}
\label{lem:discretize-scheme}
There exists a discretization scheme $0=t_0<t_1<\ldots<t_N = T - \tau < T$ satisfying Assumption \ref{ass:time-grid}, such that 
\begin{align*}
N\leq C_{\mathrm{dis}}\,\kappa^{-1}\big(T+\log(1/\tau)+1\big)
\end{align*}
for a positive numerical constant $C_{\mathrm{dis}}$.
\end{lem}

\begin{proof}[Proof of \cref{lem:discretize-scheme}] 
Starting from $u_0=\tau$, recursively set
\begin{align*}
 u_{j+1}=\begin{cases}
     \min\{1,(1+\kappa)u_j\},&u_j<1,\\
     \min\{T,u_j+\kappa\},&u_j\geq1,
 \end{cases}
\end{align*}
and stop at the first $N$ for which $u_N=T$. 
For $k = 0, 1, \ldots, N$, set $t_k=T-u_{N-k}$.
Then $\Delta_k=u_{N-k}-u_{N-k-1}
\leq\kappa\min\{1,u_{N-k-1}\}$, hence satisfying the condition in Assumption \ref{ass:time-grid}.
Observe that there are at most
$\lceil\log(1/\tau)/\log(1+\kappa)\rceil$ steps below one and
$\lceil(T-1)/\kappa\rceil$ afterward. 
The bound follows from the inequality $\log(1+\kappa)\geq\kappa/10$ for all $\kappa \in (0, 1)$.
\end{proof}

The following theorem bounds the total variation distance between $\hat\rho_1$ and $\rho_0$ under Assumption \ref{ass:time-grid}. 

\begin{thm}
\label{thm:private-ddpm-tv}
Let $\rho_0\in\cP_{\xi,L,\gamma,M}$ and suppose that the discretization scheme satisfies Assumption~\ref{ass:time-grid} for some $T \geq 1$. 
Then 
\begin{align}
\label{eq:private-ddpm-tv}
\begin{split}
\E\left[\TV(\rho_0,\hat\rho_1)\right]&\le n^{-10}+\frac{1}{2}\sqrt{\frac{|\cH|}{n}}+\frac{2\sqrt2|\cH|}{n\alpha}+\frac{\bar b_n}{2\pi}\sqrt{\cE_n}+\frac{d^{3/2}L}{4\pi}\bar b_n^{\gamma+1}\notag\\
&\quad
  +C_{L,\xi,d,\gamma,M}\sqrt{\tau}
  +C_2\sqrt{d\left(
      \kappa^2N+\kappa(T+R_n^2)
      +(1+R_n^2)e^{-2T}
  \right)},
 \end{split}
\end{align}
where
\begin{equation}
\label{eq:cEn}
\cE_n =\E\left[
  \int
  \left\|
    f_{\hat\theta_{\mathrm{LMC}}}(x)-\nabla\log\rho_0(x)
  \right\|_2^2
  \rho_0(x)\,\dd x
\right].
\end{equation}
Here, $C_2>0$ is a numerical constant,
$C_{L,\xi,d,\gamma,M}>0$ depends only on $(L,\xi,d,\gamma,M)$,
and the expectations are taken over the data and the randomness
of the private mechanisms.
Moreover, the resulting diffusion model is $(\alpha,\beta)$-DP.
\end{thm}

\begin{proof}[Proof of \cref{thm:private-ddpm-tv}]
	We prove \cref{thm:private-ddpm-tv} in Appendix \ref{proof:thm:private-ddpm-tv}. 
\end{proof}

We next prove \Cref{thm:ddpm-upper} by applying \Cref{thm:private-ddpm-tv} to bound $\E[\TV(\rho_0,\hat\rho_1)]$ under the discretization scheme from \cref{lem:discretize-scheme}. 
By \cref{thm:DP-score-upper-bound}, we know that $\cE_n\leq C_{L, \xi, d, \gamma, M} \bar b_n^{2\gamma}(\log n)^{2-\gamma}$, where $C_{L, \xi, d, \gamma, M} > 0$ is a constant that depends only on $(L, \xi, d, \gamma, M)$. 
Since
$R_n \leq C_{L,\xi,d,\gamma,M}'\sqrt{\log n}$ for a constant $C_{L,\xi,d,\gamma,M}' > 0$,
the upper bound on $|\cH|$ then gives
$|\cH|\leq C_{L, \xi, d, \gamma, M}''(\bar b_n/\sqrt{\log n})^{-d}$ where $C_{L,\xi,d,\gamma,M}'$ and $C_{L,\xi,d,\gamma,M}''$ are positive constants that depend only on $(L,\xi,d,\gamma,M)$.
The definition of $\bar b_n$ then implies that 
\begin{align*}
 n\left(\frac{\bar b_n}{\sqrt{\log n}}\right)^{d+2\gamma+2}\geq1,
 \qquad
 n\alpha\left(\frac{\bar b_n}{\sqrt{\log n}}\right)^{d+\gamma+1}\geq2.
\end{align*}
Therefore, 
\begin{align*}
 &\sqrt{\frac{|\cH|}{n}}+\frac{|\cH|}{n\alpha}
 \leq K_{L,\xi,d,\gamma,M}\left(\frac{\bar b_n}{\sqrt{\log n}}\right)^{\gamma+1}
 \leq K_{L,\xi,d,\gamma,M}'\sqrt{\tau},\\
  &\bar b_n\sqrt{\cE_n}+\bar b_n^{\gamma+1}
\leq K_{L,\xi,d,\gamma,M}'' \bar b_n^{\gamma+1}(\log n)^{1-\gamma/2}
 =K_{L,\xi,d,\gamma,M}''\sqrt{\tau}, 
\end{align*}
where $K_{L,\xi,d,\gamma,M}, K_{L,\xi,d,\gamma,M}'$ and $K_{L,\xi,d,\gamma,M}''$ are positive constants that depend only on $(L,\xi,d,\gamma,M)$. 
Therefore, by \cref{lem:corrected-density-tv}, we can bound the density error by
$K_{L,\xi,d,\gamma,M}'''\sqrt{\tau}+n^{-10}$, where $K_{L,\xi,d,\gamma,M}''' > 0$ depends only on $(L,\xi,d,\gamma,M)$.
The choices of $T$ and $\tau$ give
$d(1+R_n^2)e^{-2T}\leq \tau$. The choice of $\kappa$ and
\cref{lem:discretize-scheme} give
\begin{align*}
 &d\kappa(T+R_n^2)\leq\tau,\\
 &d\kappa^2N \leq C_{{\mathrm{dis}}}d\kappa\big(T+\log(1/\tau)+1\big)
 \leq C_{{\mathrm{dis}}}\tau.
\end{align*}
Applying \cref{thm:private-ddpm-tv} proves the rate. Substituting
$\kappa$ into \cref{lem:discretize-scheme} gives the step count.

\subsection{Proof of \cref{thm:private-ddpm-tv}}
\label{proof:thm:private-ddpm-tv}
We first establish an error bound for the corrected private density estimate \eqref{eq:sc-density}.
Specifically, the next lemma upper bounds $\E[\TV(\rho_0, \hat\rho_\cD)]$. 

\begin{lem}
\label{lem:corrected-density-tv}
With $\cE_n$ defined in \cref{eq:cEn}, 
\begin{align}
\label{eq:corrected-density-tv}
 \E\left[\TV(\rho_0,\hat\rho_\cD) \right] \leq \, n^{-10}+\frac{1}{2}\sqrt{\frac{|\cH|}{n}}
       +\frac{2\sqrt{2} |\cH|}{n\alpha}
       +\frac{\bar b_n}{2\pi}\sqrt{\cE_n}
       +\frac{d^{3/2}L\bar b_n^{\gamma+1}}{4\pi}.
\end{align}
\end{lem}

\begin{proof}[Proof of \cref{lem:corrected-density-tv}]
	We prove \cref{lem:corrected-density-tv} in Appendix \ref{proof:lem:corrected-density-tv}. 
\end{proof}

We next derive an upper bound on $\TV(\rho_0, \rho_1)$ to quantify the effect of early stopping. 

\begin{lem}
\label{lem:TV-rho0-rho1}
For $\rho_0 \in \cP_{\xi, L, \gamma, M}$, there exists a positive constant $C_{L, \xi, d, \gamma, M}$ that depends only on $(L, \xi, d, \gamma, M)$, such that  
\begin{align*}
	\TV(\rho_0, \rho_1) \leq C_{L, \xi, d, \gamma, M} \sqrt{\tau}. 
\end{align*}	
\end{lem}

\begin{proof}[Proof of \cref{lem:TV-rho0-rho1}]
	We prove \cref{lem:TV-rho0-rho1} in Appendix \ref{proof:lem:TV-rho0-rho1}. 
\end{proof}

For a probability density $q$ and $t > 0$, we write $\sP_tq$ for the law of
$e^{-t}Y+\sqrt{1-e^{-2t}}\,G$, where
$(Y,G)\sim q\otimes\normal(0,I_d)$.
By \cref{eq:sk}, we see that for all $k \in \{0\} \cup [N - 1]$, 
\begin{equation}
 \hat s_k=\nabla\log(\sP_{T-t_k}\hat\rho_\cD).
 \label{eq:exact-fitted-score}
\end{equation}
Note that the second moment of $\hat\rho_\cD$ is at most $dR_n^2$, since it is supported on $\Omega_{R_n}$.
By \cref{eq:exact-fitted-score}, the scores used in DDPM are exact from the forward process started from $\hat\rho_\cD$. 
Applying \cref{thm:benton}, we obtain
\begin{align*}
 \KL(\sP_\tau\hat\rho_\cD\|\hat\rho_1) \leq C_0 d\left(\kappa^2N+\kappa(T+R_n^2)
                          +(1+R_n^2)e^{-2T}\right).
\end{align*}
By the triangle inequality for total variation distance and Pinsker’s inequality, we obtain 
\begin{align*}
 \TV(\rho_0,\hat\rho_1)
 &\leq \TV(\rho_0, \sP_\tau\rho_0)
       +\TV(\sP_\tau\rho_0, \sP_\tau\hat\rho_\cD)
       +\TV(\sP_\tau\hat\rho_\cD,\hat\rho_1)\\
 &\leq \TV(\rho_0, \sP_\tau\rho_0)+\TV(\rho_0,\hat\rho_\cD)
          +\sqrt{\tfrac{1}{2}\KL(\sP_\tau\hat\rho_\cD\|\hat\rho_1)}.
\end{align*}
Taking expectations and applying Lemmas \ref{lem:corrected-density-tv} and \ref{lem:TV-rho0-rho1} completes the proof of the total variation distance upper bound.
In addition, the scores in \cref{eq:DM-score}, and hence the diffusion model, are postprocessing of the jointly $(\alpha,\beta)$-DP release $(f_{\hat\theta_{\mathrm{LMC}}},\hat\rho_\cD^{\hist})$.
Consequently, the resulting diffusion model is $(\alpha,\beta)$-DP.

\subsection{Proof of \cref{thm:tv-holder-lower}}
\label{proof:thm:tv-holder-lower}

We use the family $\{\rho_\vartheta:\vartheta\in\{0,1\}^{K_m}\}$ defined in \cref{eq:density-construction}, with the parameter choices from the proof of \cref{thm:score-holder-lower}.
Recall that $h=m^{-1}$, $K_m=m^d$, and $\sigma^2=2/L$.
Also recall $\omega$ is defined in \cref{eq:sigma-omega}, and for sufficiently large $m$, the perturbation amplitude satisfies
\begin{align*}
\omega=\frac{Lh^{\gamma+1}}{2C_0d^{(3-\gamma)/2}},
\qquad 0<\omega\leq\frac14,
\end{align*}
where $C_0$ is the numerical constant used in that construction.
The same proof establishes that each $\rho_\vartheta$ is $\xi$-sub-Gaussian and has a locally $(L,\gamma)$-H\"older-continuous score.
Moreover, \cref{eq:likelihood-ratio-uniform} gives
\begin{align*}
\|\E_{\rho_\vartheta}[X]\|_2
\leq\big(\E_{\rho_\vartheta}[\|X\|_2^2]\big)^{1/2}
\leq e^\omega\sigma\sqrt d
\leq\xi\sqrt d\leq M.
\end{align*}
Thus the entire family $\{\rho_\vartheta:\vartheta\in\{0,1\}^{K_m}\}$ belongs to $\cP_{\xi,L,\gamma,M}$.
Throughout this proof, $c,C>0$ denote constants that may change from line to line and depend only on $(L,\xi,d,\gamma)$. In particular, these constants are independent of $M$.

\paragraph{Total variation separation.}
For $\vartheta,\vartheta'\in\{0,1\}^{K_m}$, define
\begin{align*}
&F_{\vartheta,\vartheta'}(x)=\sum_{i\in\cI_m}(\vartheta_i-\vartheta_i')
g\left(\frac{x-x_i}{h}\right),\\
&c_{\vartheta,\vartheta'} =\frac1\omega\log\frac{Z_\vartheta}{Z_{\vartheta'}}
=\frac1\omega\log\E_{\rho_{\vartheta'}}
\left[e^{\omega F_{\vartheta,\vartheta'}(X)}\right].
\end{align*}
The disjoint supports of the bump functions imply $|F_{\vartheta,\vartheta'}|\leq1$, and hence $|c_{\vartheta,\vartheta'}|\leq1$.
Since $\omega\leq1/4$, the inequality $|e^t-1|\geq e^{-1/2}|t|$ for $|t|\leq1/2$ yields
\begin{align*}
\TV(\rho_\vartheta,\rho_{\vartheta'})
&=\frac12\int\rho_{\vartheta'}(x)
\left|e^{\omega(F_{\vartheta,\vartheta'}(x)-c_{\vartheta,\vartheta'})}-1\right|\,\dd x\\
&\geq\frac{e^{-1/2}\omega}{2}
\int\rho_{\vartheta'}(x)
|F_{\vartheta,\vartheta'}(x)-c_{\vartheta,\vartheta'}|\,\dd x.
\end{align*}
By \cref{eq:likelihood-ratio-uniform}, the densities $\rho_{\vartheta'}$ are uniformly bounded below on $[0,1]^d$ by a positive constant depending only on $(L,d)$.
Also, $g$ is non-constant on $[0,1]^d$, so
\begin{align*}
a_g=\inf_{a\in\R}\int_{[0,1]^d}|g(u)-a|\,\dd u>0.
\end{align*}
For each $i$ with $\vartheta_i\neq\vartheta_i'$, let $B_i=x_i+h[0,1]^d$.
On $B_i$, $F_{\vartheta,\vartheta'}(x)=(\vartheta_i-\vartheta_i')g((x-x_i)/h)$, and therefore
\begin{align*}
\int_{B_i}|F_{\vartheta,\vartheta'}(x)-c_{\vartheta,\vartheta'}|\,\dd x
\geq h^d a_g.
\end{align*}
Summing over these cells, whose interiors are disjoint, gives
\begin{equation}
\label{eq:tv-hamming-sep}
\TV(\rho_\vartheta,\rho_{\vartheta'})
\geq c\omega h^d\Ham(\vartheta,\vartheta')
\geq c h^{d+\gamma+1}\Ham(\vartheta,\vartheta').
\end{equation}

\paragraph{Non-private lower bound.}
By the Gilbert--Varshamov bound, choose $\cW\subseteq\{0,1\}^{K_m}$ such that $\log|\cW|\geq cK_m$ and $\Ham(\vartheta,\vartheta')\geq cK_m$ for distinct $\vartheta,\vartheta'\in\cW$.
Since $K_mh^d=1$, \cref{eq:tv-hamming-sep} implies
\begin{equation}
\label{eq:fano-tv-sep}
\TV(\rho_\vartheta,\rho_{\vartheta'})\geq c h^{\gamma+1},
\qquad \vartheta\neq\vartheta',\quad \vartheta,\vartheta'\in\cW.
\end{equation}
For an arbitrary distribution estimator $\widehat\rho$, let
\begin{align*}
\hat\vartheta_{\cW}\in
\operatorname*{arg\,min}_{\vartheta\in\cW}\TV(\widehat\rho,\rho_\vartheta),
\end{align*}
with a fixed deterministic rule for breaking ties.
The minimizing property and the triangle inequality imply
\begin{align*}
\TV(\widehat\rho,\rho_\vartheta)
\geq\frac12\TV(\rho_{\hat\vartheta_{\cW}},\rho_\vartheta)
\geq c h^{\gamma+1}\mathbbm1\{\hat\vartheta_{\cW}\neq\vartheta\},
\qquad\vartheta\in\cW.
\end{align*}
On the other hand, \cref{eq:n-sample-kl} gives
\begin{align*}
\KL(\rho_\vartheta^{\otimes n}\|\rho_\sigma^{\otimes n})
\leq16n\omega^2\leq Cnh^{2\gamma+2}.
\end{align*}
Fano's inequality consequently yields
\begin{align*}
\inf_{\widehat\rho}\sup_{\rho\in\cP_{\xi,L,\gamma,M}}
\E_{\rho^{\otimes n}}[\TV(\widehat\rho,\rho)]
\geq c h^{\gamma+1}
\left(1-Cnh^{d+2\gamma+2}-Ch^d\right).
\end{align*}
Take $m=\lceil A n^{1/(d+2\gamma+2)}\rceil$ and $h=m^{-1}$, where $A\geq1$ is a sufficiently large constant depending only on $(L,\xi,d,\gamma)$.
For sufficiently large $n$, the expression in parentheses is at least $1/2$, and
$h\geq(2A)^{-1}n^{-1/(d+2\gamma+2)}$.
It follows that
\begin{equation}
\label{eq:tv-nonprivate-lower}
\inf_{\widehat\rho}\sup_{\rho\in\cP_{\xi,L,\gamma,M}}
\E_{\rho^{\otimes n}}[\TV(\widehat\rho,\rho)]
\geq c n^{-\frac{\gamma+1}{d+2\gamma+2}}.
\end{equation}
This lower bound also applies when the infimum is restricted to private estimators.

\paragraph{Private lower bound.}
For $i\in\cI_m$, define the mixtures of dataset distributions
\begin{align*}
P_{+i}&=\frac1{2^{K_m-1}}
\sum_{\vartheta:\,\vartheta_i=1}\rho_\vartheta^{\otimes n},
&P_{-i}&=\frac1{2^{K_m-1}}
\sum_{\vartheta:\,\vartheta_i=0}\rho_\vartheta^{\otimes n}.
\end{align*}
For neighboring vertices $\vartheta,\vartheta'$ that differ only in coordinate $i$, the argument in the proof of \cref{thm:score-holder-lower} gives
\begin{align*}
\TV(\rho_\vartheta,\rho_{\vartheta'})\leq C\omega h^d.
\end{align*}
Couple each pair of observations maximally and independently across the $n$ samples.
The resulting coupling of $\rho_\vartheta^{\otimes n}$ and $\rho_{\vartheta'}^{\otimes n}$ has expected Hamming distance at most $Cn\omega h^d$.
Averaging these couplings over the common coordinates $\vartheta_{-i}$ gives a coupling of $P_{+i}$ and $P_{-i}$ with the same bound, which we denote by $D=Cn\omega h^d$.

Applying the private Assouad inequality \citep[Theorem 3]{acharya2021differentially} with the TV separation in \cref{eq:tv-hamming-sep}, and using $K_mh^d=1$, yields
\begin{align*}
&\inf_{\widehat\rho\in\mathcal M_{\alpha,\beta}}
\sup_{\rho\in\cP_{\xi,L,\gamma,M}}
\E_{\rho^{\otimes n}}[\TV(\widehat\rho,\rho)]\\
&\qquad\geq c\omega
\left(0.9e^{-10\alpha D}-10\beta D\right),
\qquad D=Cn\omega h^d.
\end{align*}
Now take $m=\lceil A(n\alpha)^{1/(d+\gamma+1)}\rceil$ and $h=m^{-1}$, with $A\geq1$ sufficiently large depending only on $(L,\xi,d,\gamma)$.
Since $\omega=a_0h^{\gamma+1}$, this choice ensures $\alpha D\leq1/100$.
The assumption $\beta\leq\alpha$ then implies
\begin{align*}
0.9e^{-10\alpha D}-10\beta D
\geq0.9e^{-1/10}-\frac1{10}>\frac12.
\end{align*}
For sufficiently large $n$, we also have
$h\geq(2A)^{-1}(n\alpha)^{-1/(d+\gamma+1)}$.
Consequently,
\begin{align*}
\inf_{\widehat\rho\in\mathcal M_{\alpha,\beta}}
\sup_{\rho\in\cP_{\xi,L,\gamma,M}}
\E_{\rho^{\otimes n}}[\TV(\widehat\rho,\rho)]
\geq c(n\alpha)^{-\frac{\gamma+1}{d+\gamma+1}}.
\end{align*}
Combining this with \cref{eq:tv-nonprivate-lower} proves the theorem.

\subsection{Proof of \cref{lem:corrected-density-tv}}
\label{proof:lem:corrected-density-tv}

For $\nu \in \cH$, 
let $p_\nu'=\int_{A_\nu}\rho_0(y)\,\dd y$,
$p_\nu=\P_{\rho_0}(\cT_{R_n}(X)\in A_\nu)$, and
$\rho_{0,\nu}(y) = \rho_0(y)\mathbbm1_{A_\nu}/p_\nu'$. Then
$p_\nu\geq p_\nu'$ and
$\sum_{\nu \in \cH}(p_\nu-p_\nu')=\rho_0(\Omega_{R_n}^c)\leq n^{-10}$.
With these definitions, we obtain the following upper bound for the total variation distance:
\begin{align*}
 \TV(\rho_0,\hat\rho_\cD)
 &\leq n^{-10}+\frac12\|\hat p_\cD^{\hist}-p\|_1
          +\sum_{\nu \in \cH} p_\nu'\TV(\rho_{0,\nu},q_{\cD,\nu})\\
 &\leq n^{-10}+\frac12\|\hat p_\cD^{\hist}-p\|_1
   +\frac{\bar b_n}{2\pi}\left\{
     \sum_{\nu \in \cH}\int_{A_\nu}
       \|\nabla\log\rho_0(y)-v_{\cD,\nu}\|_2^2\rho_0(y)\,\dd y
                         \right\}^{1/2},
\end{align*}
where the second step above uses \cref{lem:sc-local-poincare} and the Cauchy-Schwarz inequality. 

The coefficient constraints and \cref{eq:gcp-upper} imply
\begin{align*}
 \sup_{y\in A_\nu}\big\|f_{\hat\theta_{\mathrm{LMC}}}(y) - v_{\cD,\nu}\big\|_2 \leq \frac{d^{3/2}L \bar b_n^\gamma}{2},
\end{align*}
By triangle inequality, it holds that 
\begin{align*}
	\left\{
     \sum_{\nu \in \cH}\int_{A_\nu}
       \|\nabla\log\rho_0(y)-v_{\cD,\nu}\|_2^2\rho_0(y)\,\dd y
                         \right\}^{1/2} \leq \left\{\int \|\nabla \log \rho_0(y) - \hat f_{\hat\theta_{\mathrm{LMC}}}(y)\|_2^2 \rho_0(y) \dd y\right\}^{1/2} + \frac{d^{3/2}L\bar b_n^\gamma}{2}. 
\end{align*}
The multinomial variance bound yields
\begin{align*}
 \E[\|\hat p_\cD^{\hist}-p\|_1]
 &\leq \E[\|\bar p_\cD^{\hist}-p\|_1]
          +\sqrt{|\cH|}\,\E[\|\hat p_\cD^{\hist}-\bar p_\cD^{\hist}\|_2]\\
 &\leq \sqrt{\frac{|\cH|}{n}} + \sqrt{|\cH|}\,(\E[\|\zeta\|_2^2])^{1/2}
 =\sqrt{\frac{|\cH|}{n}}+\frac{4\sqrt{2}|\cH|}{n\alpha}.
\end{align*}
Taking expectations completes the proof of the lemma.

\subsection{Proof of \cref{lem:TV-rho0-rho1}}
\label{proof:lem:TV-rho0-rho1}

Observe that 
\begin{align*}
	\TV(\rho_0, \rho_1) = &\, \frac{1}{2}\int |\rho_0(x) - \rho_1(x)|\, \dd x \\
	\leq &\, \frac{1}{2}\int |\rho_0(x) - e^{d\tau} \rho_0(e^{\tau} x)|\, \dd x + \frac{1}{2}\int |e^{d\tau} \rho_0(e^{\tau} x) -\rho_1(x) |\, \dd x \\
	= & \, \frac{1}{2}\int |\rho_0(x) - e^{d\tau} \rho_0(e^{\tau} x)|\, \dd x + \frac{1}{2}\int |\rho_0(x) - \rho_0 \ast \varphi_{e^{2\tau} - 1}(x)|\, \dd x. 
\end{align*}
Since $\rho_0 \ast \varphi_{e^{2\tau} - 1}(x) = \E_G[\rho_0(x - \sqrt{e^{2\tau} - 1}\, G)]$, where the expectation is over $G \sim \normal(0, I_d)$. 
Jensen's inequality implies that  
\begin{align*}
	\frac{1}{2}\int |\rho_0(x) - \rho_0 \ast \varphi_{e^{2\tau} - 1}(x)|\, \dd x \leq &\, \frac{1}{2} \,\E_G \left[ \int \left|\rho_0(x) - \rho_0\left(x - \sqrt{e^{2\tau} - 1} G\right)\right| \dd x\right] \\
	\leq & \,\frac{\sqrt{e^{2\tau} - 1}}{2} \, \E_G \left[ \|G\|_2 \int \|\nabla \rho_0(x)\|_2 \, \dd x \right] \\
	\leq & \, \frac{\sqrt{d(e^{2\tau} - 1)}}{2} \E_{\rho_0}\left[ \|\nabla \log \rho_0(X)\|_2 \right] \\
	\leq & \, C^{1/4} L (1 + \xi \sqrt{d})\frac{\sqrt{d(e^{2\tau} - 1)}}{2}, 
\end{align*}
where the last inequality follows from \cref{lem:score-fourth-moment}, and $C > 0$ is the numerical constant therein. 

For $z \in [0, \tau]$, define $\kappa_z(x) = e^{dz} \rho_0(e^z x)$. Differentiating with respect to $z$ yields 
\begin{align*}
	\partial_z \kappa_z(x) = de^{dz} \rho_0(e^z x) + e^{(d + 1)z} \langle x,  \nabla \rho_0(e^z x) \rangle = \kappa_z(x) \left(d + \langle x, \nabla \log \kappa_z(x) \rangle \right). 
\end{align*}
As a consequence, 
\begin{align*}
	\frac{1}{2}\int |\rho_0(x) - e^{d\tau} \rho_0(e^{\tau} x)|\, \dd x \leq & \,\frac{1}{2} \int_0^{\tau} \int \kappa_z(x) \left|d + \langle x, \nabla \log \kappa_z(x) \rangle \right| \dd x \, \dd z \\
	\leq & \, \frac{\tau}{2} \left( d + \int \rho_0(y) \left| \langle y, \nabla \log \rho_0(y) \rangle \right| \dd y \right) \\
	\leq & \,  \frac{\tau}{2} \left( d +  \frac{1}{2} \int \|y\|^2_2 \,\rho_0(y)  \dd y + \frac{1}{2} \int \|\nabla \log \rho_0(y)\|^2_2 \,\rho_0(y)  \dd y \right) \\
	\leq & \,\tau \left( d + \frac{1}{4}C^{1/2} L^2 (1 + \xi^2 d) + \frac{1}{4} (M^2 + d \xi^2) \right). 
\end{align*}
Combining the two bounds, we see that 
\begin{align*}
	\TV(\rho_0, \rho_1) \leq C_{L, \xi, d, \gamma, M} \sqrt{\tau}, 
\end{align*}
where $ C_{L, \xi, d, \gamma, M} > 0$ depends only on $(L, \xi, d, \gamma, M)$. The proof is done.

\section{Additional numerical experiments}

\subsection{LMC diagnostics}
\label{sec:diagnostics}

\begin{figure}[htbp]
    \centering
    \includegraphics[width=1\linewidth]{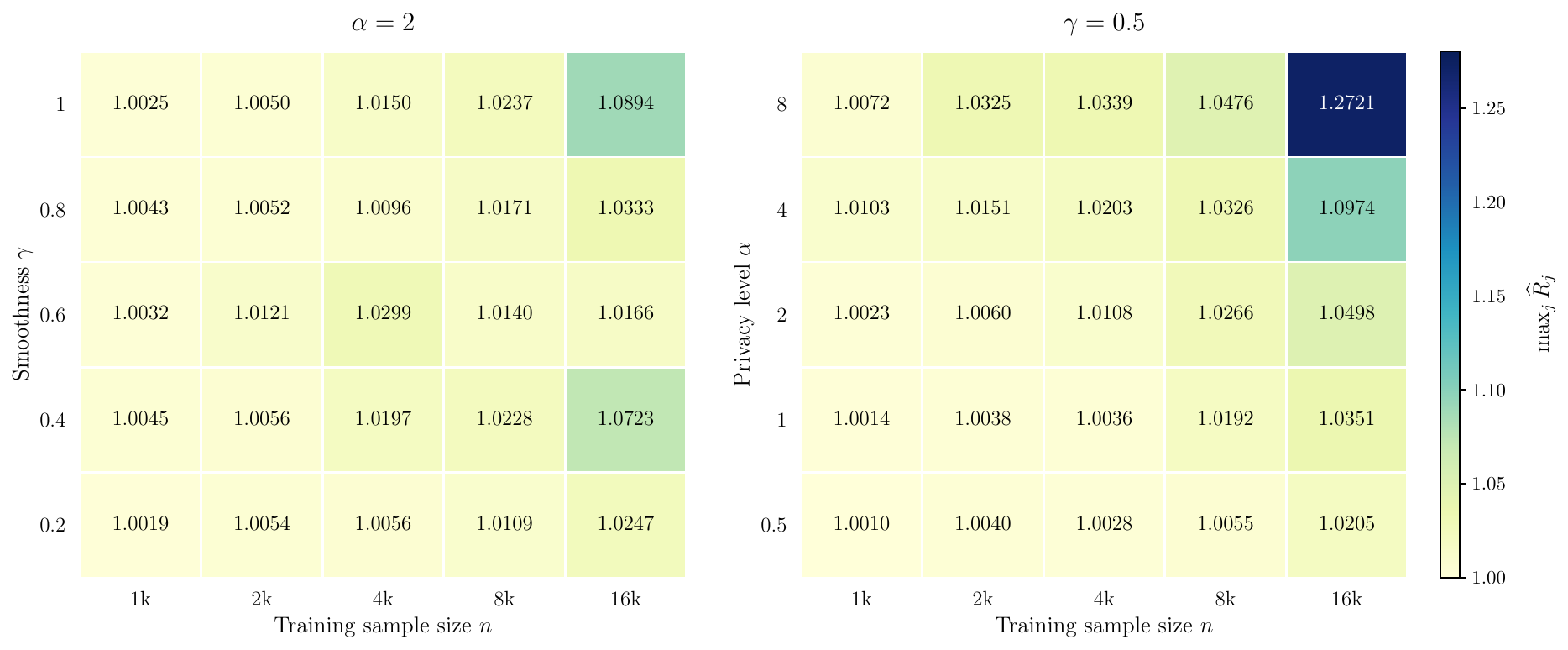}
    \caption{Empirical convergence diagnostics for projected LMC. Each cell reports the maximum $\widehat R$ across all spline coefficients, computed from four independent chains. The left panel fixes $\alpha=2$ and varies the smoothness parameter $\gamma$, while the right panel fixes $\gamma=0.5$ and varies $\alpha$. }
    \label{fig:r-hat-heatmap}
\end{figure}

We assess the convergence of projected LMC empirically using four independently simulated chains initialized from zero. 
For each coefficient, we compute the maximum of the rank-normalized split $\widehat{R}$ and its folded counterpart, following \citet{vehtari2021rank}, and interpret a value at or below $1.01$ as evidence of adequate mixing. 
The heatmaps in Figure \ref{fig:r-hat-heatmap} report the largest value across all coefficients for each configuration. 
All chains are run using 500,000 projected LMC iterations. 
From these figures, we see that for smaller sample sizes, the maximum $\widehat R$ values are generally close to one, indicating relatively adequate mixing. 
However, larger values occur at larger sample sizes, particularly under weaker privacy constraints, where increased concentration of the sampling distribution may make mixing more challenging under a fixed iteration budget. Since the diagnostic takes the maximum over all coefficients, such value does not indicate poor mixing throughout the coefficient vector. Taken together, these results suggest generally favorable mixing behavior across the experimental settings, while highlighting a small number of more demanding configurations.

\subsection{Two-dimensional synthetic experiments}
\label{sec:two-dim-synthetic}
\paragraph{Multidimensional simulation setting.}
We extend the synthetic experiments in Section~\ref{subsec:1d-synthetic} to a correlated two-dimensional target distribution. Let $Z = (Z_1, Z_2)^{\top}$, where $Z_1$ and $Z_2$ are independent and have the same univariate density
\begin{align}
    q_{\gamma}(z)\propto \exp\left(-\frac{z^2}{2} -\frac14\sum_{k=0}^{\infty}2^{-k(1+\gamma)}\cos(2^kz)\right). 
\end{align}
Here, we similarly truncate to the first 40 terms when $\gamma = 1$. 
We define $X=BZ$, where
\[
B\approx
\begin{pmatrix}
0.876463 & 0.234848\\
0.234848 & 0.876463
\end{pmatrix}.
\]
The matrix $B$ is kept fixed across all experiments and is chosen so that each component of $X$ has unit variance and the correlation between its first and second components is $0.5$. 
Writing $s_{\gamma} (z) = \partial \log q_{\gamma}(z)$, we obtain
\begin{align*}
    \nabla\log\rho_0(x)
    = (B^{-1})^{\top}
    \begin{pmatrix}
        s_{\gamma}((B^{-1}x)_1) \\
        s_{\gamma}((B^{-1}x)_2)
    \end{pmatrix}.
\end{align*}
We use the same sample sizes $n \in \{1000,2000,4000,8000,16000\}$, privacy levels $\alpha \in \{0.5, 1, 2, 4,8\}$ and smoothness levels $\{0.2, 0.4, 0.6, 0.8, 1\}$ as in Section~\ref{subsec:1d-synthetic}. 
For the projected LMC, we run $500,000$ iterations with a fixed step scaling factor $\delta = 0.02 / \lambda_{\max}(\Sigma_{\cD})$. Unless otherwise specified, we follow the evaluation and repetition protocols described in the one-dimensional experiments.

\begin{figure}[!htbp]
    \centering
    \includegraphics[width=\textwidth]{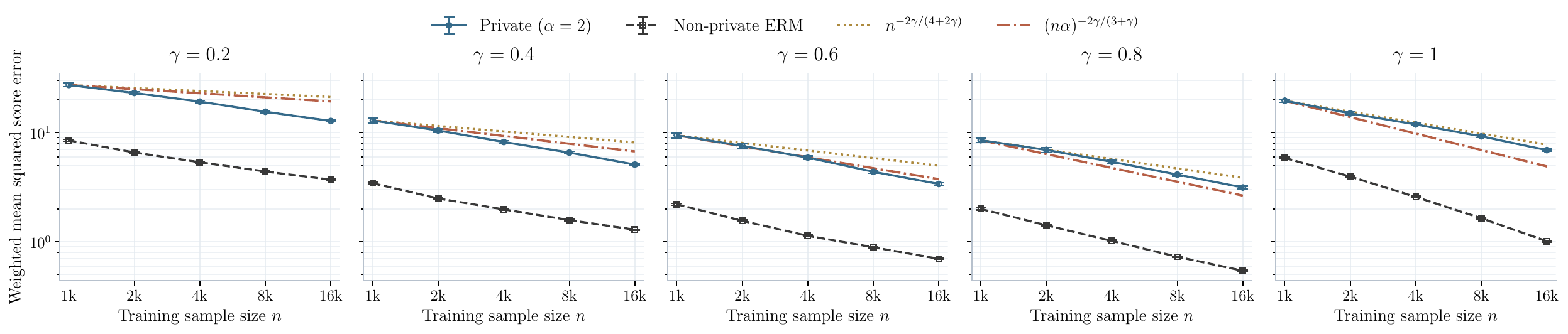}
    \caption{Two-dimensional clean-score estimation risk with $\alpha=2$. Each panel fixes a smoothness parameter $\gamma$ and plots the density-weighted mean squared score error against the training sample size $n$ on logarithmic scales. The non-private ERM benchmark is included for comparison. The dotted and dash-dotted lines represent $n^{-2\gamma/(4+2\gamma)}$ and $(n\alpha)^{-2\gamma/(3+\gamma)}$, respectively, with intercepts calibrated to the first private error point in each panel. The lines illustrate theoretical slopes.}
    \label{fig:two-dimensional-gamma-score-risk}
\end{figure}

\paragraph{Clean score estimation.}
We first evaluate the density-weighted mean squared score error of the private score estimator and compare it with the non-private ERM estimator.
Figure~\ref{fig:two-dimensional-gamma-score-risk} fixes $\alpha =2$ and compares the density-weighted mean squared score error at different smoothness levels. For every $\gamma$ considered, we observe that the private risk decreases as the training sample size increases, and the non-private ERM benchmark remains lower. The observed decay rates are steeper than the reference lines, particularly at smaller $\gamma$. This aligns with the theoretical results, as the reference rates characterize the worst-case rates over the considered class of distributions.

\begin{figure}[!htbp]
    \centering
    \includegraphics[width=\textwidth]{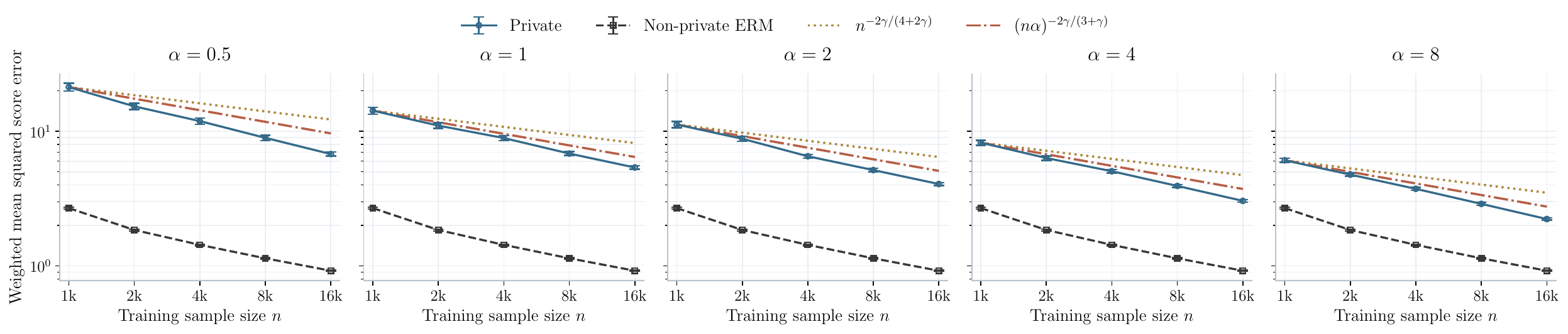}
    \caption{Two-dimensional clean-score estimation risk with $\gamma=0.5$. Each panel fixes a privacy level $\alpha$. }
    \label{fig:two-dimensional-alpha-score-risk}
\end{figure}

Figure~\ref{fig:two-dimensional-alpha-score-risk} fixes $\gamma = 0.5$ and varies the privacy budget. 
We observe that the private risk decreases with $n$ for every value of $\alpha$. Moreover, increasing $\alpha$ consistently improves the accuracy of the private estimator and reduces its gap relative to the non-private ERM. The private curves also exhibit a faster decline over the sample size choices than the two displayed reference slopes.

\begin{figure}[!htbp]
    \centering
    \includegraphics[width=\textwidth]{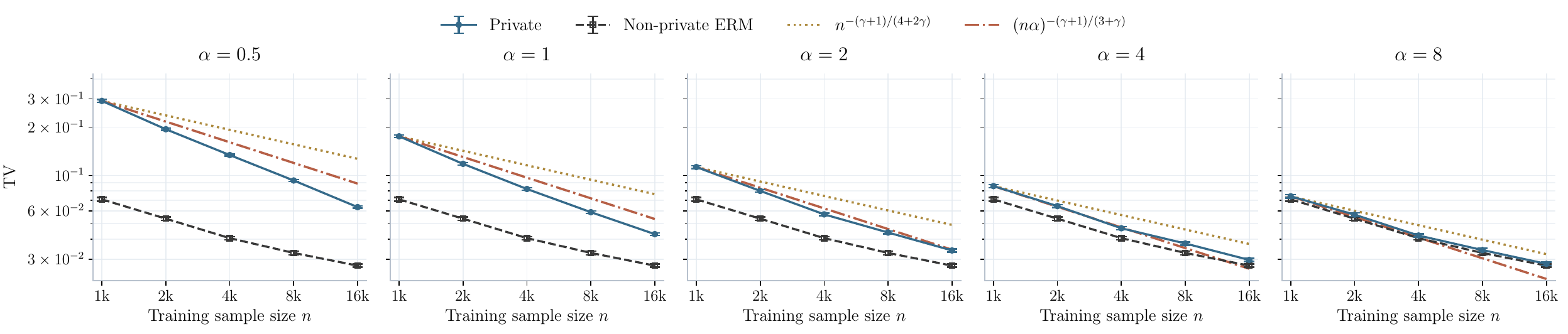}
    \caption{Two-dimensional diffusion-model TV error with $\gamma=0.5$. Each panel fixes a privacy level $\alpha$ and shows the mean TV error against the training sample size $n$, together with the non-private benchmark. The dotted and dash-dotted lines display the reference rates $n^{-(\gamma+1)/(4+2\gamma)}$ and $(n\alpha)^{-(\gamma+1)/(3+\gamma)}$, respectively, and their intercepts are matched to the first private error point in each panel.}
    \label{fig:two-dimensional-alpha-tv}
\end{figure}

\paragraph{Diffusion model application.}
We next use the fitted private clean score estimator together with a private histogram to construct a private two-dimensional diffusion model, also following the procedure in Section~\ref{subsec:1d-synthetic}. The privacy budget for clean score estimation remains unchanged, while allocating same budget for private histogram. We compare the estimated total variation error between the resulting output distribution and the target with the corresponding non-private diffusion benchmark. For $d=2$, the TV minimax rate is the maximum of $n^{-(\gamma+1)/(2\gamma +4)}$ and $(n\alpha)^{-(\gamma+1)/(\gamma +3)}$. 
Figure~\ref{fig:two-dimensional-alpha-tv} fixes $\gamma = 0.5$ and examines the effect of varying the privacy parameter. The TV error decreases steadily with $n$ for both the private and non-private models. As $\alpha$ increases, the gap between the two models narrows, and their curves are close at $\alpha = 8$. 

\begin{figure}[!htbp]
    \centering
    \includegraphics[width=\textwidth]{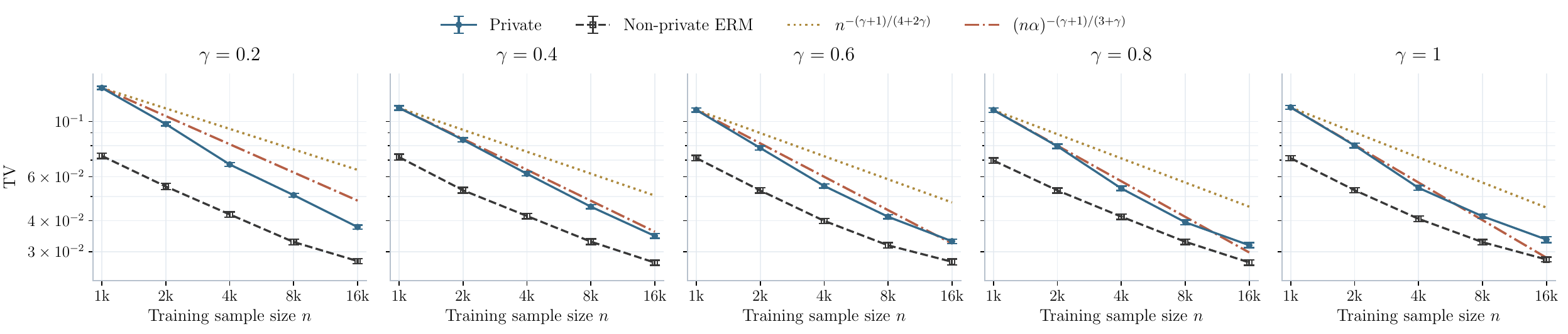}
    \caption{Two-dimensional diffusion-model TV error with $\alpha=2$. Each panel fixes a smoothness parameter $\gamma$. The non-private benchmark and the reference rates are plotted using the same conventions as in Figure~\ref{fig:two-dimensional-alpha-tv}.}
    \label{fig:two-dimensional-gamma-tv}
\end{figure}

Figure~\ref{fig:two-dimensional-gamma-tv} fixes $\alpha =2$ and changes $\gamma$. Both private and non-private TV errors decrease across all the sample sizes considered. The private curves generally falls near the private reference rate and are steeper than the non-private reference rate. 

In summary, the two-dimensional results show that the qualitative privacy–accuracy trade-off observed in the one-dimensional synthetic experiments persists in the two-dimensional setting when the coordinates are dependent.

\begin{figure}[!htbp]
    \centering
    \includegraphics[width=0.5\textwidth]{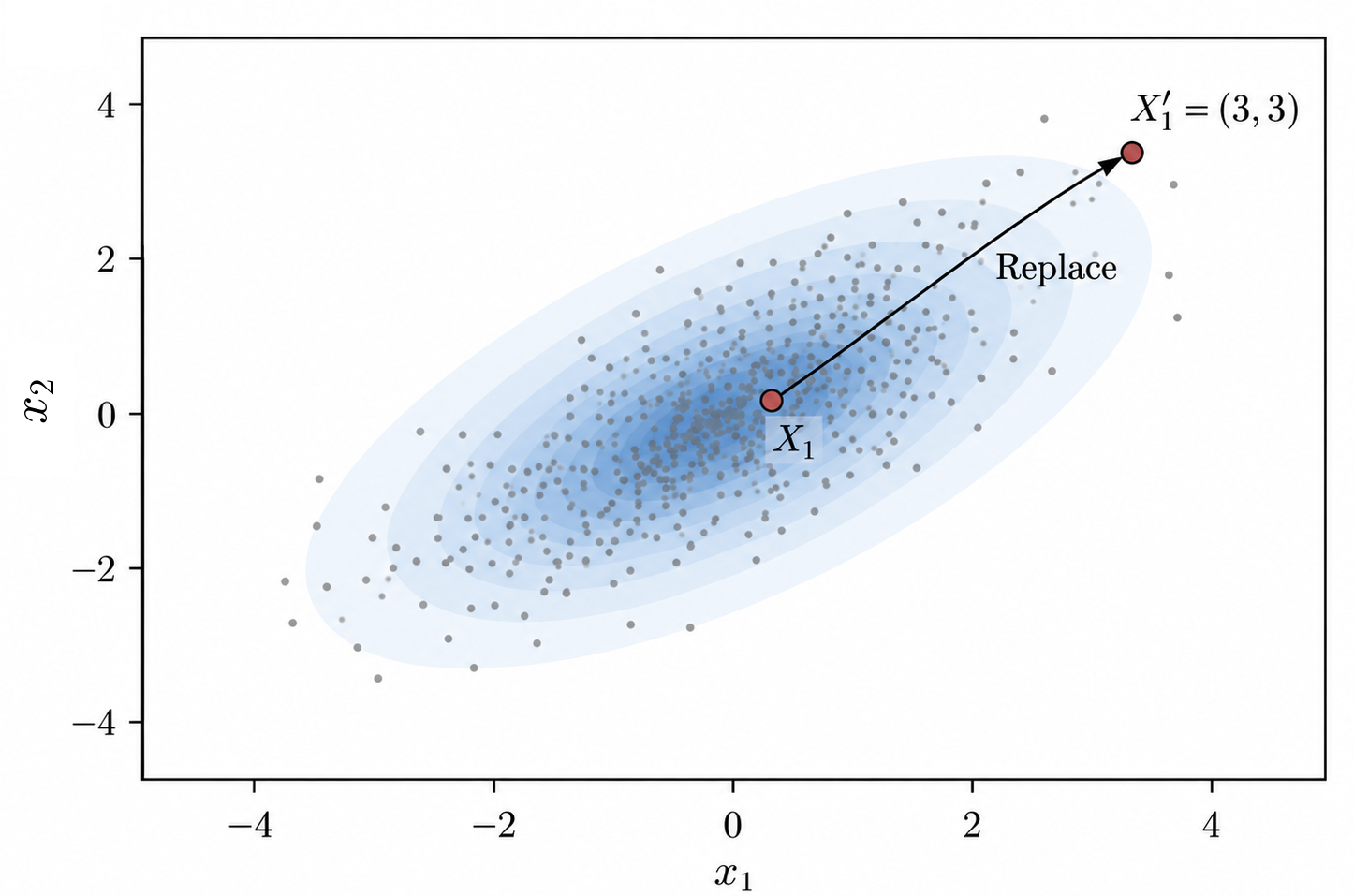}
    \caption{The target distribution and sampled observations for the 2-dimensional setting, illustrating the replacement of a single observation by $(3, 3)$.}
    \label{fig:two-dimensional-perturbation}
\end{figure}

\paragraph{Perturbation analysis.} We also assess privacy by examining the empirical stability of our score estimator under a single observation perturbation. Similar to the one-dimensional scenario, for each repetition, we construct two neighboring dataset $\cD$ and $\cD'$, where $\cD'$ is obtained by replacing the first observation in $D$ by $(3,3)$. The other sample remains unchanged. We refer to Figure~\ref{fig:two-dimensional-perturbation} for an illustration of this perturbation. 

Let $\hat\theta_{\cD}$ and $\hat\theta_{\cD'}$ denote the spline coefficient estimates obtained from the two neighboring datasets. We quantify their difference using the root mean squared (RMS) coefficient change, and compare the results with those of the non-private ERM. We fix $\gamma=0.5$ and consider different sample sizes and privacy levels.

\begin{figure}[!htbp]
    \centering
    \begin{subfigure}[b]{0.48\textwidth}
        \centering
        \includegraphics[width=\linewidth]{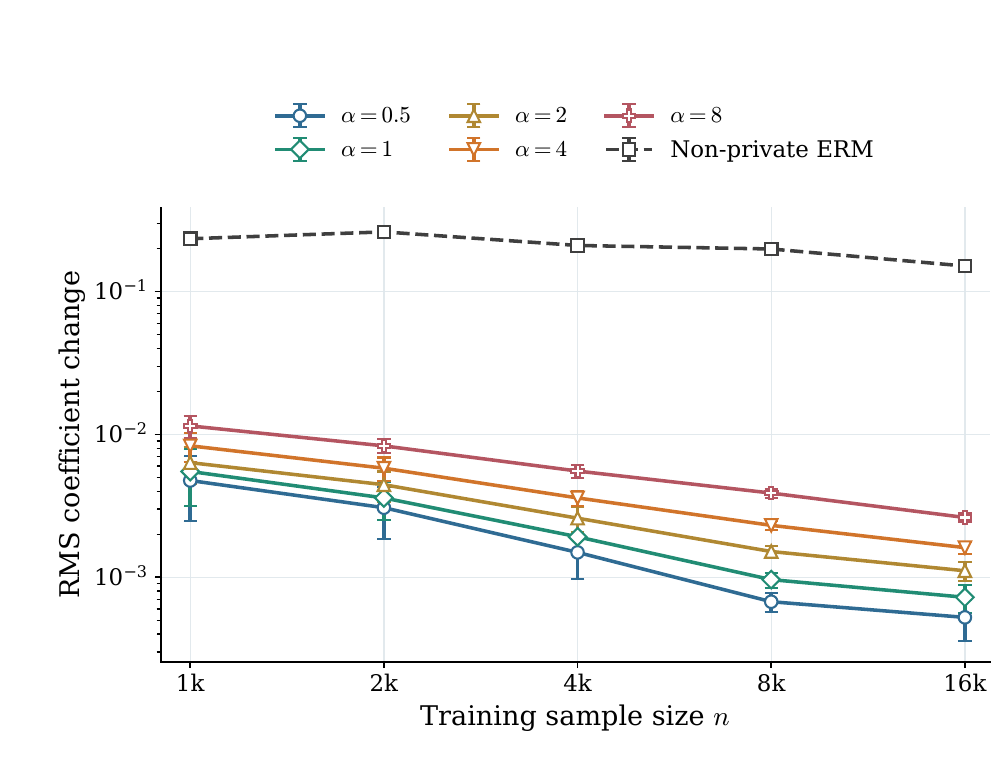}
        \caption{RMS coefficient changes induced by replacing one observation with $(3,3)$.}
    \end{subfigure}
    \hfill
    \begin{subfigure}[b]{0.48\textwidth}
        \centering
        \includegraphics[width=\linewidth]{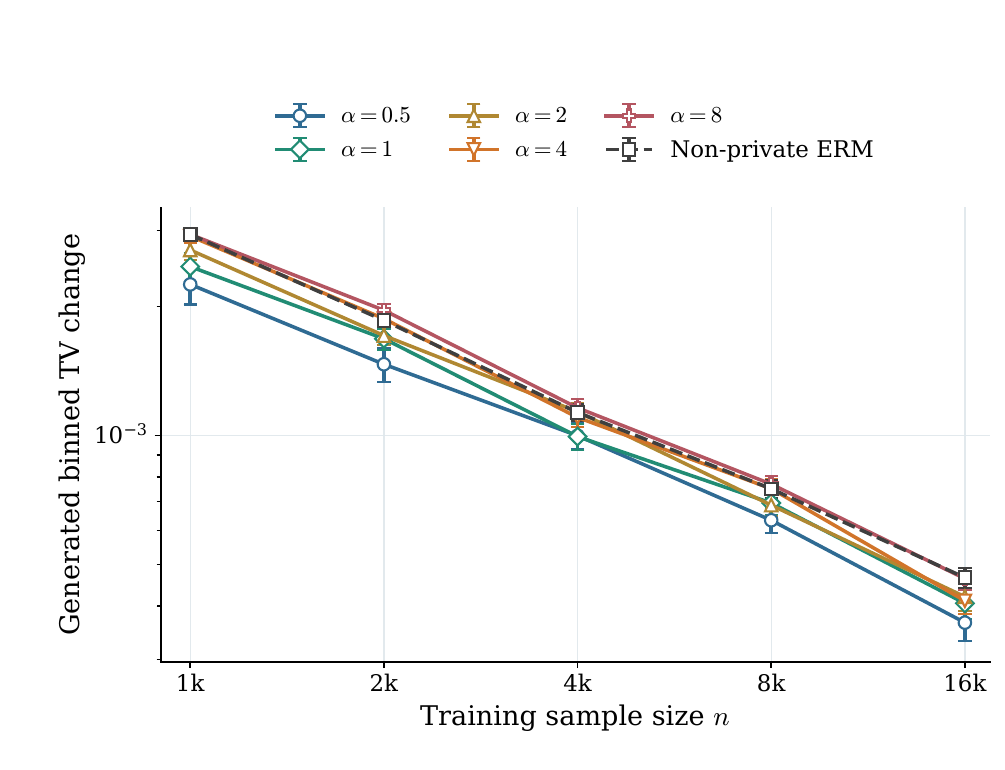}
        \caption{Binned TV changes of the generated distributions under the same replacement.}
    \end{subfigure}
    \caption{Empirical stability under single-observation perturbations in the two-dimensional setting with $\gamma=0.5$. Results are shown for different privacy levels and training sample sizes, with non-private ERM as a benchmark.}
    \label{fig:two-dimensional-perturbation-results}
\end{figure}

Panel (a) in Figure \ref{fig:two-dimensional-perturbation-results} shows that our private estimator exhibits substantially smaller coefficient changes than non-private ERM across all settings considered. More specifically, for the private estimator, the coefficient changes decrease as the sample size increases and increase with the privacy parameter $\alpha$. In contrast, the coefficient changes of non-private ERM remain substantially larger and vary relatively little across the sample sizes considered. These observations are consistent with the stability induced by stronger privacy constraints.

We next examine the effect of the same single observation replacement on the generated distributions, measuring the changes using binned total variation distance. 
Panel (b) in Figure~\ref{fig:two-dimensional-perturbation-results} shows that the empirical TV changes decrease steadily when we increase the sample sizes for both private and non-private diffusion models. 
Private diffusion models generally exhibit smaller TV changes, although the differences are much less pronounced than those observed at the coefficient level. Similarly to the one-dimensional results, these findings suggest that the output distributions of diffusion models can be relatively insensitive to single-observation replacements, even when their underlying score estimators exhibit substantially different coefficient-level sensitivities.

\end{document}